\pdfoutput=1

\documentclass[11pt]{article}
\usepackage[utf8]{inputenc}
\usepackage{times}
\usepackage{algorithm}
\usepackage[noend]{algorithmic}
\usepackage{nicefrac}
\usepackage{booktabs}
\usepackage{footmisc}
\usepackage[shortlabels]{enumitem}
\usepackage[margin=1in]{geometry}
\usepackage{comment}
\usepackage{booktabs,tabularx}
\usepackage{multirow}
\usepackage{textcomp}
\usepackage[numbers]{natbib}
\usepackage{graphicx}
\usepackage{color}
\usepackage[hidelinks]{hyperref}
\usepackage{wrapfig}
\usepackage{amssymb,amsmath,amsthm}
\usepackage[capitalise,noabbrev]{cleveref}
\usepackage{array}
\usepackage{url}
\usepackage[strict]{changepage}
\usepackage{makecell}
\usepackage{titlesec}
\usepackage{setspace}
\usepackage{bm}
\usepackage{bbm}
\usepackage{threeparttable}
\usepackage{subfigure}
\newtheorem{Def}{Definition}
\usepackage[T1]{fontenc}
\newcommand{\myparagraph}[1]{\smallskip \indent{\it {#1}}}
\usepackage{titletoc}
\usepackage{longtable}

\titleformat*{\subparagraph}{\itshape}

\newsavebox\actorsfigure

\title{A Comprehensive Survey on Pretrained Foundation Models: A History from BERT to ChatGPT}
\author{
Ce Zhou$^{1}$\footnote{The authors contributed equally to this research. Correspondence to Ce Zhou(\url{zhouce@msu.edu}) and  Qian Li (\url{liqian@act.buaa.edu.cn}).} \and
Qian Li$^{2*}$ \and
Chen Li$^{2*}$ \and
Jun Yu$^{3*}$ \and
Yixin Liu$^{3*}$ \and
Guangjing Wang$^{1}$ \and
Kai Zhang$^{3}$ \and
Cheng Ji$^{2}$ \and
Qiben Yan$^{1}$ \and
Lifang He$^{3}$ \and
Hao Peng$^{2}$ \and
Jianxin Li$^{2}$ \and
Jia Wu$^{4}$ \and
Ziwei Liu$^{5}$ \and
Pengtao Xie$^{6}$ \and
Caiming Xiong$^{7}$ \and
Jian Pei$^{8}$ \and
Philip S. Yu$^{9}$ \and
Lichao Sun$^{3}$ \and\\
\small{$^{1}$Michigan State University, $^{2}$Beihang University, $^{3}$Lehigh University,}\\\small{$^{4}$Macquarie University, $^{5}$Nanyang Technological University, $^{6}$University of California San Diego,}\\\small$^{7}$Salesforce AI Research,{$^{8}$Duke University, $^{9}$University of Illinois at Chicago}
}

\date{}

\begin{document}

\begin{spacing}{1.1}
\maketitle
\end{spacing}

\begin{abstract}

Pretrained Foundation Models (PFMs) are regarded as the foundation for various downstream
tasks with different data modalities. A PFM (e.g., BERT, ChatGPT, and GPT-4) is trained on large-scale data which provides a reasonable parameter initialization for a wide range of downstream applications. In contrast to earlier approaches that utilize convolution and recurrent modules to extract features, BERT learns bidirectional encoder representations from Transformers, which are trained on large datasets as contextual language models. Similarly, the Generative Pretrained Transformer (GPT) method employs Transformers as the feature extractor and is trained using an autoregressive paradigm on large datasets. Recently, ChatGPT shows promising success on large language models, which applies an autoregressive language model with zero shot or few shot prompting.  
The remarkable achievements of PFM have brought significant breakthroughs to various fields of AI in recent years. Numerous studies have proposed different methods, datasets, and evaluation metrics, raising the demand for an updated survey.

This study provides a comprehensive review of recent research advancements, challenges, and opportunities for PFMs in text, image, graph, as well as other data modalities. The review covers the basic components and existing pretraining methods used in natural language processing, computer vision, and graph learning. Additionally, it explores advanced PFMs used for different data modalities and unified PFMs that consider data quality and quantity. The review also discusses research related to the fundamentals of PFMs, such as model efficiency and compression, security, and privacy. Finally, the study provides key implications, future research directions, challenges, and open problems in the field of PFMs. Overall, this survey aims to shed light on the research of the PFMs on scalability, security, logical reasoning ability, cross-domain learning ability, and the user-friendly interactive ability for artificial general intelligence.

\end{abstract}


\pagebreak

\begin{small}
\tableofcontents
\end{small}

\setlength{\parskip}{0.5em}

\pagebreak

\section{Introduction}\label{Section 1}





Pretrained Foundation Models (PFMs) are regarded as essential and significant components of Artificial Intelligence (AI) in the era of big data. The foundation model is first named in~\cite{bommasani2021opportunities}, which means a broader class of models and their functions. PFMs are extensively studied in the three major AI fields: natural language processing (NLP)~\cite{chowdhury2003natural}, computer vision (CV)~\cite{forsyth2011computer} and graph learning (GL)~\cite{bondy1976graph}. PFMs are powerful general models that are effective in various fields or across fields. They have demonstrated great potential in learning feature representations in various learning tasks, such as text classification~\cite{qiu2020pre}, text generation~\cite{li2021pretrained}, image classification~\cite{han2020survey}, object detection~\cite{sanchez2020review}, and graph classification~\cite{hu2019pre}. 
PFMs show superior performance for training on multiple tasks with large-scale corpus and fine-tuning it to similar small-scale tasks, making it possible to initiate rapid data processing. 


\subsection{PFMs and Pretraining}

\begin{figure*}[!t]
    \centering
    \includegraphics[width=1\linewidth]{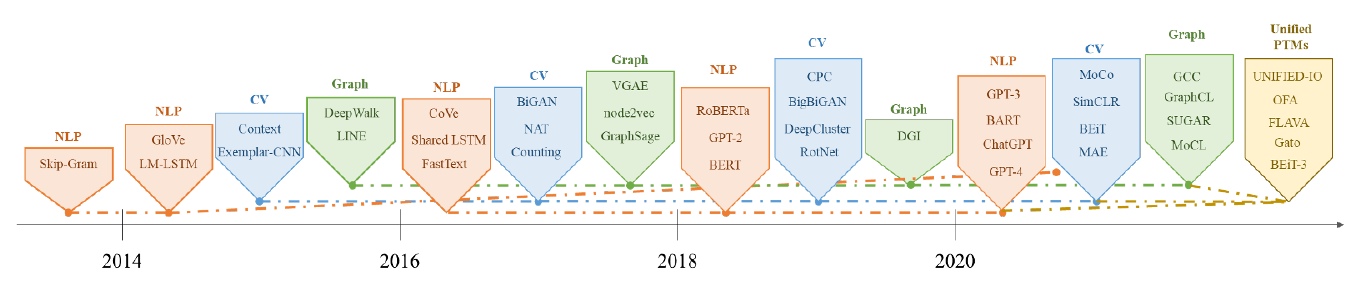}
    \caption{The history and evolution of PFMs.}
    \label{history_evolution}
\end{figure*}

PFMs are built upon the pretraining technique, which aims to train a general model using large amounts of data and tasks that can be fine-tuned easily in different downstream applications.
The idea of pretraining originates from transfer learning~\cite{zhuang2020comprehensive} in CV tasks. Recognizing the effectiveness of pretraining in the field of CV, people have begun to use pretraining technology to enhance model performance in other areas. When pretraining techniques are applied to the NLP domain, well-trained language models (LMs) can capture rich knowledge beneficial for downstream tasks, such as long-term dependencies, hierarchical relationships, etc. In addition, the significant advantage of pretraining in the NLP field is that training data can be derived from any unlabeled text corpus, that is, there is an unlimited amount of training data in the pretraining process.
Early pretraining is a static technique, such as NNLM~\cite{DBLP:journals/jmlr/BengioDVJ03} and Word2vec~\cite{DBLP:journals/corr/abs-1301-3781}, but static methods were difficult to adapt to different semantic environments. Therefore, dynamic pretraining techniques are proposed, such as BERT~\cite{DBLP:conf/naacl/DevlinCLT19}, XLNet~\cite{DBLP:conf/nips/YangDYCSL19}, etc.
Fig.~\ref{history_evolution} depicts the history and evolution of PFMs in the NLP, CV, and GL domains. The PFMs based on the pretraining technique use large corpora to learn generic semantic representations. With the introduction of these pioneering works, various PFMs have emerged and been applied to downstream tasks and applications.


A great example of PFM application is ChatGPT\footnote{\url{https://openai.com/blog/chatgpt/}}. ChatGPT is fine-tuned from the generative pretrained transformer GPT-3.5, which was trained on a blend of text and code~\cite{chen2021evaluating, neelakantan2022text}. ChatGPT applies reinforcement learning from human feedback (RLHF) \cite{christiano2017deep, stiennon2020learning}, which has become a promising way to align large language models (LLMs) with a human's intent~\cite{ouyang2022training}. The surprisingly superior performance of ChatGPT may lead to a tipping point for a shift of training paradigm for each type of PFMs -- applying \textit{instruction aligning} techniques, e.g., reinforcement learning (RL), prompt tuning \cite{brown2020language, lester2021power, schick2021exploiting}, and chain-of-thought (COT) \cite{zhang2022automatic, weichain}, to move towards artificial general intelligence. 



We focus on reviewing PFMs for text, image, and graph, which is a relatively mature research taxonomy.
For text, it is a multi-purpose LM to predict the next word or character in a sequence. For example, PFMs can be used for machine translation, question-answering systems, topic modeling, sentiment analysis, etc.
For image, it is similar to PFMs on text, which uses huge datasets to train a big model suitable for many CV tasks. For graphs, a similar pretraining idea is also applied to obtain PFMs, which are used for many downstream tasks. Apart from the PFMs for a specific data domain, we also review and state some other advanced PFMs, such as the PFMs for speech, video, and cross-domain data, and multimodal PFMs. An exemplary illustration is the GPT-4 model, as described by OpenAI~\cite{openai2023gpt4}, which is a massive multimodal language model that can process both text and image inputs and generate text outputs. GPT-4 has demonstrated human-level performance on various professional and academic evaluation tasks.  
Moreover, there is a growing trend in PFMs that deals with multimodal data, known as unified PFMs. This term refers to models that can handle different types of data such as text, images, and audio. In this regard, we provide a definition of unified PFMs and a review of the current state-of-the-art models in recent research. Notable examples include OFA~\cite{wang2022unifying}, UNIFIED-IO~\cite{lu2022unified}, FLAVA~\cite{singh2022flava}, BEiT-3~\cite{wang2022image}, and others.


According to the features of existing PFMs, we conclude that the PFMs have the following two major advantages. First, minor fine-tuning is required to enhance the model performance on downstream tasks. Second, the PFMs have already been vetted on the quality aspect. Instead of building a model from scratch to solve a similar problem, we can apply PFMs to task-related datasets. 
The great promise of PFMs has inspired a wealth of related work to focus on the model efficiency~\cite{DBLP:conf/iclr/ClarkLLM20}, security~\cite{nlp52019arxiv, nlp62020acl, nlp32019acl, wang2023glint} and compression~\cite{DBLP:conf/rep4nlp/GordonDA20, DBLP:conf/iclr/LanCGGSS20}.



\subsection{Contribution and Organization}

There are several survey studies~\cite{han2021pre, sanchez2020review, qiu2020pre, li2021pretrained, han2020survey, bommasani2021opportunities} that have reviewed the pretrained models for some specific areas such as text generation~\cite{li2021pretrained}, visual transformer~\cite{han2020survey}, objection detection~\cite{sanchez2020review}.

Bommasani et.al.~\cite{bommasani2021opportunities} summarize the opportunities and risks of the foundation model. However, existing works did not achieve a comprehensive review of PFMs in different areas (e.g., CV, NLP, GL, Speech, Video) and different aspects such as pretraining tasks, efficiency, efficacy, and privacy.
In this survey, we specifically track the evolution of PFMs in the NLP domain, as well as how pretraining  is transferred to and adopted by CV and GL.
Compared with other surveys, there is no comprehensive introduction and analysis of existing PFMs from all three fields.
Unlike reviews of previous pretrained models, we summarize existing models ranging from traditional models to PFMs with recent works in the three domains.
Traditional models emphasize static feature learning. 
Dynamic PFMs give an introduction to structures, which is the mainstream research.
We further present some other research for PFMs, including other advanced and unified PFMs, model efficiency and compression, security, and privacy. Finally, we summarize future research challenges and open problems in different domains. We also comprehensively present the related evaluation metrics and datasets \textbf{in Appendix~\ref{Evaluation_Metrics} and~\ref{datasets}}.
In summary, the main contributions are as follows:

\begin{itemize}
    \item We present a solid and up-to-date review of the development of PFM in NLP, CV, and GL. Over the review, we discuss and provide insights about the generalized PFM design and pretraining methodology among the three major application domains.
    
    \item We summarize the development of PFMs in other multimedia areas such as speech and video. Besides, we discuss advanced topics about PFMs, including unified PFMs, model efficiency and compression, and security and privacy.

    \item Through the review of PFMs in various modalities for different tasks, we discuss the main challenges and opportunities for future research of very large models in the big data era, which guides a new generation of collaborative and interactive intelligence based on PFMs.
\end{itemize}



The rest of the survey is organized as follows. 
Section~\ref{Section 2} introduces the basic components.
Sections~\ref{Section 3},~\ref{Section 4} and~\ref{Section 5} summarize the existing PFMs in NLP, CV and GL, respectively.
Sections~\ref{Section 6},~\ref{Section 7} introduce other advanced research for PFMs, including advanced and unified PFMs, model efficiency and compression, as well as security and privacy, respectively. 
Furthermore, we summarize the main challenges for PFMs in Section~\ref{Section 11} before concluding the survey in Section~\ref{Section 12}.
\section{Basic Components} 
\label{Section 2}
\begin{figure*}
    \centering
    \includegraphics[width=1\linewidth]{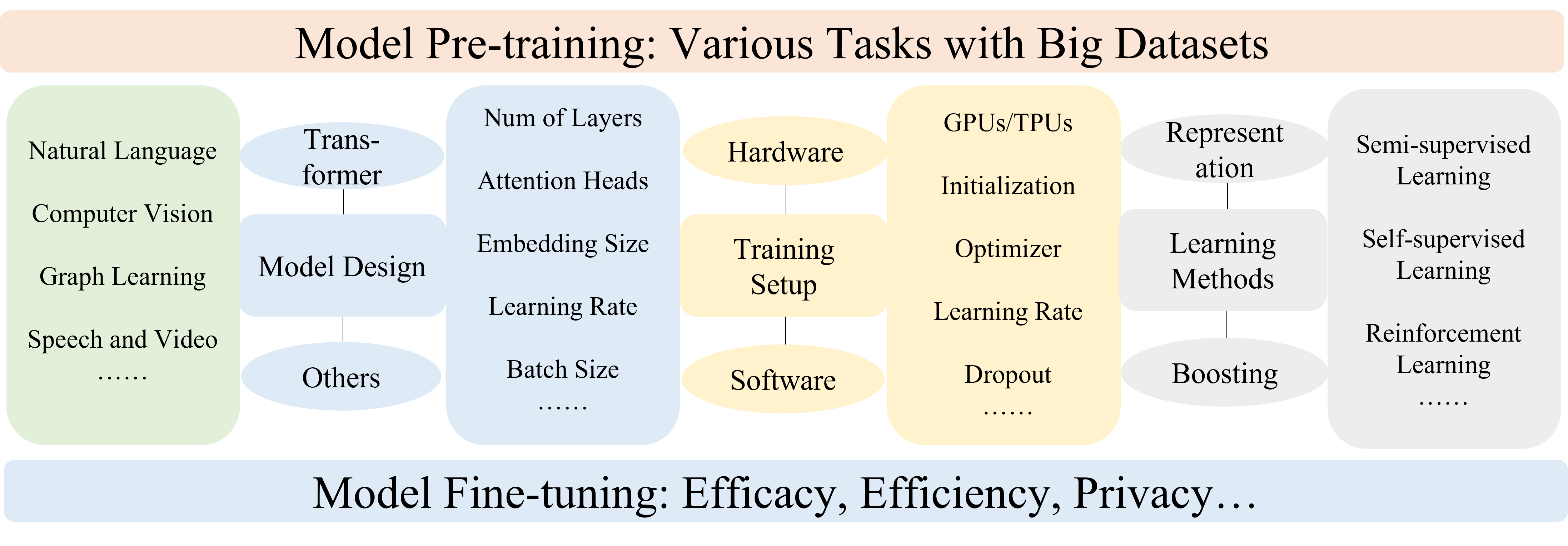}
    \caption{The general conceptual architecture of PFMs: data, model, and system.}
    \label{fig:ptm_concepts}
\end{figure*}

The general conceptual architecture of PFMs is shown in Fig.~\ref{fig:ptm_concepts}. The PFMs are huge neural network models, which are all about neural information processing. The specific designs of PFMs vary according to the data modality and task requirements in different areas. Transformer is a mainstream model architecture design for PFMs in many areas such as NLP and CV. Training large models need to have various datasets for model pretraining. After training the PFMs, the model should be fine-tuned to satisfy downstream requirements such as efficacy, efficiency, and privacy. In this section, we introduce the basic model architectures, concepts, and settings of PFMs in NLP, CV, and GL domains. For the introduction of a more detailed component, please refer to \textbf{Appendix~\ref{Components}}.

\subsection{Transformer for PFMs}
The Transformer~\cite{vaswani2017attention} is an innovative architecture that facilitates the transfer of weighted representation knowledge between various neural units. It relies solely on attention mechanisms and doesn't use recurrent or convolutional architectures. The attention mechanism is a crucial component of the Transformer as it assigns weights to all the encoded input representations and learns the most important part of the input data. The output of the attention is obtained by taking the weighted sum of the values, and the weights are calculated using the compatibility function of the query with the corresponding key~\cite{vaswani2017attention}. Numerous attention mechanisms~\cite{guo2022attention} have been developed in large models. For instance, in natural language processing, self-attention is created to connect various positions in a single sequence for generating a representation of the same sequence. Transformer leverages a mask matrix to provide an attention mechanism based on self-attention, in which the mask matrix specifies which words can ``see''  each other.

Transformer is an important structure for PFMs in NLP, CV, and GL areas. For NLP, the Transformer can help solve the long-range dependency issues when processing sequential input data. For example, the GPT-3~\cite{brown2020language} is a generative model based on the transformer. For CV, the Vision Transformer (ViT)~\cite{dosovitskiy2020image} is proposed to represent an image to a series of image patches, which is similar to a series of word embeddings. For GL, the Graph Transformer Networks (GTN)~\cite{yun2019graph} are employed to learn new graph structures and powerful node representations without domain knowledge. Transformers become scalable enough to drive ground-breaking capabilities for PFMs thanks to the transformer structures to achieve higher parallelization. The ViT-22B model~\cite{vitransform22b}, for instance, has about 22B parameters, and the largest language models can have upwards of 100B parameters (e.g., GPT-3 has 175B and PaLM~\cite{chowdhery2022palm} has 540B parameters).


\subsection{Learning Mechanisms for PFMs}

Deep learning models in CV have been shown a large margin to outperform traditional learning models in most tasks, including the common classification, recognition, detection, and segmentation tasks and the specific matching, tracking, and sequence prediction. These learning methods are not only available in CV, but also in NLP and GL.
\paragraph{Supervised Learning} Suppose we are given a training dataset $\bm X$ containing $\{(\bm{x}_i, y_i)\}_{i=1}^{n}$ to represent the original data in training dataset, where $\bm{x}_i$ denotes the $i$-th training sample, and $y_i$ denotes the corresponding label. The complete network is to learn a function $f(\bm{x};\bm \theta)$ by minimizing the objective function as follows.
\begin{equation}
    \mathop{\arg\min}_{\bm \theta} \ \ \frac{1}{n}\sum\nolimits_{i=1}^{n}{\mathcal{L}(f(\bm{x}_i;\bm \theta),y_i)}
    +\lambda\Omega(\bm\theta),
\end{equation}
where $\mathcal{L}$ and $\Omega$ represent the predefined loss function and a regularization term, respectively. The function $f$ has a nested form like
\begin{equation}
\begin{aligned}
    \bm h_{1}(\bm{x}_i)&=g(\bm{x}_i^\top\bm{\omega}_1+b_1), \\ \bm h_{l+1}(\bm{x}_i)&=g(\bm h_l(\bm{x}_i)^\top\bm{\omega}_l+b_l), 
    l = 1, 2, \cdots, N
\end{aligned}
\end{equation}
where $l$ is the index of layer in deep learning model and $N$ is the number of layers, which means that $\bm\theta = \{\bm\omega_l,b_l, l = 1, 2, \cdots, N\}$.

\paragraph{Semi-Supervised Learning} Assume we are given another unlabelled dataset $\bm Z = \{\bm z_i\}_{i=1}^{m}$ in addition to the previous dataset with human labels. If we want to utilize both datasets to learn an ideal network, the learning process can be formulated as
\begin{equation}
    \mathop{\arg\min}_{\bm\theta} \ \ \frac{1}{n}\sum\nolimits_{i=1}^n{\mathcal{L}(f(\bm x_i;\bm\theta),y_i) + \\ 
    \frac{1}{m}\sum\nolimits_{i=1}^m{\mathcal{L}^{\prime}(f^{\prime}(\bm{z}_i;\bm\theta^{\prime}),R(\bm z_i, \bm X))}} + \lambda\Omega(\bm\theta),
\end{equation}
where $R$ is a relation function defining the targets for unlabelled data, and then these pseudo-labels are integrated into the end-to-end training process. $f^{\prime}$ is an encoder to learn a new representation for the original data in the dataset $\bm Z$. Specifically, if there is no label to any data in the training process, we can learn from the properties inside the data itself via the internal distance or the designed pretext tasks, which are known as unsupervised learning and self-supervised learning(SSL), respectively. The latter is our main focus discussed in detail in Section~\ref{sec:Learning by Generation}.
\paragraph{Weakly-Supervised Learning}
The weakly-supervised method is the balance between fully-supervised learning and SSL according to the dependence on human labels. The SSL designs special pretext tasks 
to serve as the supervised learning, but the fully supervised learning
utilizes existing labels attached to the data. However, both of them can learn good visual features and perform well on specific downstream tasks. Suppose there are inaccurate $K$ labels for the dataset, and any label can be attached to a data sample. Thus, we denote the true label of image ${\bm x}_i$ as ${\bm y}_i\in\{0,1\}^K, i=1,2,\cdots,n$, and any entry of ${\bm y}_i$ could be $0$ or $1$. Here we need to minimize the total $nK$ loss terms
, which are formulated as follows.
\begin{equation}
    \mathop{\arg\min}_{\bm\theta} \ \ \frac{1}{nK}\sum\nolimits_{i=1}^n\sum\nolimits_{k=1}^K{\mathcal{L}(f(\bm x_i;\bm\theta),y_i^k)} + \lambda\Omega(\bm\theta),
\end{equation}
where $\left[y_i^1, y_i^2, \cdots, y_i^K\right] = {\bm y}_i$, and $\mathcal{L}$ could be a loss function suitable for binomial classification problem. For any entry in ${\bm y}_i$, computing the loss function of the one-versus-all binomial classification is needed.

\paragraph{Self-Supervised Learning} SSL utilizes the information in the data itself to learn essential feature representations for different tasks. By applying the self-defined pseudo labels, it can avoid the cost of manually labeling large datasets for PFMs. In NLP, the language models can be trained by predicting masked characters, words, or sentences. Variational autoencoder (VAE) and generative adversarial network (GAN) are two types of generative SSL methods, which are to reconstruct the data itself. Besides, contrastive learning, as a type of discriminative SSL method, is widely applied in CV, NLP, and GL. The main idea of contrastive learning is to learn the prior knowledge distribution of the data itself with the aid of various methods such as data augmentation. In this way, contrastive learning can learn a model that makes similar instances closer in the projected space, and dissimilar instances farther apart in the projected space. Here we show a simple version of contrastive loss: 
\begin{equation}
    \mathcal{L}_\text{c}(\mathbf{x}_i, \mathbf{x}_j, \theta) = m \| f_\theta(\mathbf{x}_i) - f_\theta(\mathbf{x}_j) \|^2_2  \\ + (1-m)\max(0, \epsilon - \|f_\theta(\mathbf{x}_i) - f_\theta(\mathbf{x}_j)\|_2)^2
\label{eq:contrastive}
\end{equation}
where $m$ is 1 if two samples have the same label, otherwise 0, and $\epsilon$ is the upper bound distance.

\paragraph{Reinforcement Learning} RL is another type of learning paradigm that models the learning process as a sequential interaction between an agent and an environment, where a RL agent seeks to learn an optimal policy for sequential decision-making problems. Specifically, at each time interaction step $t$, the agent receives a state $s_t$ in a state space $\mathcal{S}$, and selects an action $a_t$ from an action space $\mathcal{A}$, following a policy $\pi_{\theta}(a_t|s_t): \mathcal{A}\rightarrow\mathcal{S}$ parameterized by $\theta$. 
Then the agent receives a scalar immediate reward $r_t=r(s_t,a_t)$ and the next state $s_{t+1}$ according to the environment dynamics, where $r(s,a)$ is the reward function. 
For each episode, this process continues until the agent reaches a terminal state. After an episode is finished, the RL agent will restart to begin a new episode. 
The return for each state is discounted, accumulated reward with the discount factor $\gamma \in (0,1]$,
$R_t=R(s_t,a_t)= \sum_{k=0}^{\infty} \gamma^k r_{t+k}.$
The agent aims to maximize the expectation of such long-term return from each state, 
\begin{equation}
    \max_{\theta}{\mathbb{E}_{s_t}[R_t|s_t,a_t=\pi_{\theta}(s_t)]}.
\end{equation}

\subsection{Pretraining Tasks for PFMs}
Pretraining is an initialization framework, which generally needs to be used in conjunction with fine-tuning downstream tasks.
In the scheme of pretraining and finetuning, the parameters of the model are trained on pre-set tasks to capture specific attributes, structure, and community information. The pretrained features can assist downstream tasks, provide sufficient information, and speed up the convergence of the model.

\subsubsection{Pretraining Tasks for NLP}
The pretraining tasks can be divided into five categories according to the learning methods: Mask Language Modeling (MLM), Denoising AutoEncoder (DAE), Replaced Token Detection (RTD), Next Sentence Prediction (NSP), Sentence Order Prediction (SOP). 
RTD, NSP, and SOP are contrastive learning methods, which assume that the observed samples are more semantically similar than the random samples. 

\textbf{Mask Language Modeling (MLM).} MLM erases some words randomly in the input sequence and then predicts these erased words during pretraining. Typical examples include BERT~\cite{DBLP:conf/naacl/DevlinCLT19} and SpanBERT~\cite{DBLP:journals/tacl/JoshiCLWZL20}.

\textbf{Denoising AutoEncoder (DAE).} DAE is used to add noise to the original corpus and reconstruct the original input using the corpus containing noise. BART~\cite{DBLP:conf/acl/LewisLGGMLSZ20} is a representative example.

\textbf{Replaced Token Detection (RTD).} RTD is a discriminant task that determines whether the LM has replaced the current token. This task is introduced in ELECTRA~\cite{clark2020electra}. By training the model to distinguish whether a token has been replaced or not, the model can acquire language knowledge.

\textbf{Next Sentence Prediction (NSP).} In order to make the model understand the correlation between the two sentences and capture sentence-level representations, a NSP task is introduced. The PFM inputs two sentences from different documents and checks whether the order of the sentences is correct. A typical example is BERT.

\textbf{Sentence Order Prediction (SOP).} Different from NSP, SOP uses two contiguous fragments from a document as positive samples and the exchange order of the two fragments as negative samples. The PFMs can better model the correlation between sentences, such as ALBERT~\cite{lan2019albert}. 

\subsubsection{Pretraining Tasks for CV} There are many pretraining tasks created for CV to learn the feature space, which is based on SSL. It utilizes pretext tasks that contain human-designed labels, like jigsaw puzzles or the comparison of various patches from images. This enables the generalization of learned representations to a range of downstream tasks.

\textbf{Specific Pretext Task.}
A pretext task also referred to as a predefined task, is created for the encoder networks to perform during the pretraining phase. The network is trained by predicting the answer to a special pretext task. Based on particular features of the data, pseudo labels are generated for the fictitious task. Then, using guided learning techniques, the encoder networks are trained to solve the pretext task.  For example, inpainting aims to pretrain models by predicting the missed center part.

\textbf{Frame Order Learning Task.} Learning frame order from videos involves frame processing through time steps, which can serve as the pretraining task for CV. This issue usually relates to completing pretextual exercises that can aid in the acquisition of visual temporal representations.

\textbf{Data Generation Task.} The representational capabilities within the generative adversarial networks (GANs) can also be used in the pretraining tasks. Projecting data back into the latent space, as demonstrated by BiGANs~\cite{donahue2016adversarial}, is helpful for auxiliary supervised discrimination tasks by acting as feature representations.

\textbf{Data Reconstruction Task.} Since the images can be divided into patches inspired by the natural language, some pretraining tasks for NLP can also be used in CV, e.g., the autoencoder-based masked prediction. The original image is first divided into a few patches and discrete visual tokens are used to encode each patch. The visual tokens from the masked patches are outputted in the second stage to match the corresponding visual tokens from the fixed tokenizer.

\textbf{Miscellaneous.} To train the PFMs in CV, additional pretraining tasks are suggested. For instance, based on contrastive learning, encoder networks are used for pretraining on various data augmentation. The parameters are trained by maximizing the distance between negative pairs (e.g., pairs with different labels) and minimizing the distance between positive pairs (e.g., pairs with the same labels). To pretrain the parameters of the backbone network, the DeepClustering~\cite{caron2018deep} method divides the representations into various clusters and labels these clusters as supervised signals.

\subsubsection{Pretraining Tasks for GL}

The pre-set tasks in GL are similar to other pretext tasks. However, they can be supervised or unsupervised depending on the design. According to the pretraining purpose and potential motivation in GL, such tasks can be divided into the following categories:

\textbf{Graph Information Completion.}
This task refers to firstly masking part of the information in the input graph, and then recovering the masked information based on the analysis of the remaining information distribution.
Similar tasks also exist in CV and NLP, and their goals are to fill in hidden pixels or words, respectively.

\textbf{Graph Property Prediction.}
Different from directly modeling the information of the input graph, this task aims to provide a variety of self-supervised signals by mining the potential properties of the input graph.
Specifically, on the one hand, it considers node attributes, local substructure, and connectivity information to provide predictive regression tasks; on the other hand, it assigns pseudo-labels to nodes through information such as clusters, structure density, and attribute similarity to provide classification tasks.

\textbf{Graph Consistency Analysis.}
The goal of this task is to maximize the consistency between samples with similar semantic information in the graph embedding and minimize the agreement between samples with unrelated semantic information.
In the actual scenario, it can be divided into consistency analysis of context/self/cross-scale according to different model training strategies.

\textbf{Miscellaneous.}
Compared with using only one pretext task, some methods have designed some integration mechanisms to incorporate the advantages of multiple pretext tasks into a unified framework.
Besides, some graph data in specific fields have unique self-supervised signals with practical significance that can be used for pretraining under targeted design.

In summary, the transformer is an important component of the large model architecture, which helps learn the important features and mine intrinsic structure in data. Different learning mechanisms can be used for training PFMs according to the datasets and specific tasks. Especially, SSL is a promising mechanism to learn knowledge embeddings from the data considering the large scale of unlabeled data in various areas. RL provides a new way to fine-tune the PFMs for downstream tasks by optimizing a policy (model) against the reward model. How to design effective and efficient tasks for PFMs to master the knowledge behind the data is an important research topic.


\section{PFMs for Natural Language Processing} \label{Section 3}

NLP is a research field that integrates linguistics and computer science. Its main research tasks include part-of-speech tagging, named entity recognition, semantic role labeling, machine translation, question answering, sentiment analysis, text summarization, text classification, relationship extraction, event extraction, etc. 
The idea of PFM first gained popularity in NLP. Then CV and GL adopt the promising pretraining technology.
The PFM trains on a large benchmark dataset and is fine-tuned on the primary task dataset to obtain a model which can solve new similar tasks.
It models syntactic and semantic representations of words simultaneously and changes the representation of polysemous words dynamically according to different input contexts. PFM learns a rich knowledge of grammar and semantic reasoning with better results. 
Numerous PFMs have been proposed in the past few years, as shown in \textbf{Table~\ref{text}}. 

In this section, we first introduce word representation learning models including the autoregressive language model (LM), contextual LM, and permuted LM. 
Then, we present the neural network architectures for the PFM designing method and the masking designing method. Besides, we summarize boosting methods for enhancing model performance, multi-task learning, and different downstream tasks. Finally, we introduce the instruction-aligning methods, e.g. RLHF and Chain-of-Thoughts, which are applied in PFMs, such as ChatGPT, to provide outputs that more closely match human preferences and are less harmful.

   

\subsection{Word Representations Methods}

 Many large-scale pretrained models have achieved better performance than humans in question answering, machine reading comprehension, and natural language reasoning, which indicates that the current construction approach of PFMs is practical.
The existing pretraining LMs are mainly divided into three branches according to the word representations approach: \emph{(1) autoregressive LM, (2) contextual LM, and (3) permuted LM}. The word prediction direction and contextual information are the most important factors among these three branches. 


\paragraph{Autoregressive Language Model}

The autoregressive LM predicts the next possible word based on the preceding word or the last possible word based on the succeeding word. 
It is selected as a feature extractor and text representations are extracted from the former words.
Thus, it has better performance in NLG tasks such as text summarization and machine translation. 
For a sequence, $T=[w_{1}, w_{2}, \ldots, w_{N}]$, the probability of a given word calculated as follows:
\begin{equation}
p\left(w_{1}, w_{2}, \ldots, w_{N}\right)=\prod_{i=1}^{N} p\left(w_{i} \mid w_{1}, w_{2}, \ldots, w_{i-1}\right),
\end{equation}
where $i>1$ and $N$ is the length of the input sequence.

The GPT~\cite{radford2018improving} adopts a two-stage method of self-supervised pretraining and supervised fine-tuning and uses stacked Transformer~\cite{vaswani2017attention} as its decoder.
As a follow-up, the OpenAI team continues to expand GPT, proposes the GPT-2~\cite{radford2019language} and increases the number of stacked Transformer layers to 48 layers. The total number of parameters reached 1.5 billion. GPT-2 also introduces multi-task learning~\cite{caruana1997multitask}.
The GPT-2 has a considerable model capacity and can be adjusted for different task models rather than fine-tuning them. However, GPT-2 also uses an autoregressive LM. Therefore, it improves the performance of the model without increasing the cost dramatically. Due to the lack of contextual modeling ability with a one-way Transformer, the main performance improvement of GPT-2 comes from the combined effect of multi-task pretraining, super-large datasets, and super-large models.
Task-based datasets for fine-tuning are still needed for specific downstream tasks.
Increasing the training scale of the LM can lead to a significant enhancement in task-independent performance. Hence, GPT-3~\cite{brown2020language} was developed, which features a model size of 175 billion parameters and is trained with 45 Terabytes of data. This enables it to exhibit good performance without the need for fine-tuning for specific downstream tasks.

\paragraph{Contextual Language Model}

The autoregressive LM only uses the information above or below and cannot use the information above and below at the same time. ELMO~\cite{peters2018deep} only uses bi-directional Long Short-Term Memory (LSTM), which is a concatenation of two unidirectional LSTMs in backward and forward.
The contextual LM predictions are based on contextual words.
It uses a Transformer encoder, and the upper and lower layers of the model are all directly connected to each other due to the self-attention mechanism.
For a sequence of words $T$, the probability of a given word calculates as follows
\begin{equation}
p\left(w_{1}, w_{2}, \ldots, w_{N}\right)=\prod_{i=1}^{N} p\left(w_{i} \mid w_{1}, w_{2}, \ldots, w_{N}\right).
\end{equation}

BERT~\cite{DBLP:conf/naacl/DevlinCLT19} uses a stacked multi-layer bi-directional Transformer as the basic structure, and WordPiece~\cite{wu2016google} as a word segmentation method. The model input consists of three parts: word embedding, segment embedding, and position embedding.
It uses a bi-directional Transformer as a feature extractor, which offsets the defect of ELMO and GPT. 
However, the shortcomings of BERT are also not to be ignored. The bidirectional Transformer structure does not eliminate the constraints of the self-encoding model. Its vast number of model parameters are very unfriendly to devices with low computing resources and are challenging to deploy and apply.
Furthermore, the hidden language modeling in pretraining will lead to inconsistencies with the input of the model in the fine-tuning stage.
Most PFMs need more training tasks and a larger corpus.
Aiming at the problem of insufficient training, Liu et al.~\cite{DBLP:journals/corr/abs-1907-11692} propose the RoBERTa. It uses a larger batch size and unlabeled data. Furthermore, it trains the model for a longer time, removes the NSP task, and adds long sequence training. In processing text input, different from BERT, Byte Pair Encoding (BPE)~\cite{sennrich2015neural} is adopted for word segmentation. BPE uses a different mask mode for each input sequence, even if the input sequence is the same.

\paragraph{Permuted Language Model}

The modeling method with a contextual LM can be regarded as the autoencoding model. However, due to the inconsistency in the training stage and fine-tuning stage, the performance of the autoencoding model is poor in the Natural Language Generation (NLG) task.
Permuted LM aims to combine the advantages of the autoregressive LM and the autoencoder LM.
It improves the defects of the two models to a great extent and can be used as a basic idea for the construction of future pretraining target tasks.
For a given input sequence $T=[w_{1},w_{2}...
,w_{N}]$, the formal representation of the target function of the permuted LM is as follows
\begin{equation}
\max _{\theta} \mathbb{E}_{z \sim Z_{N}}\left[\sum_{t=1}^{N} \log p_{\theta}\left(x_{z_{T=t}} \mid x_{z_{T<t}}\right)\right],
\end{equation}
where $\theta$ is the shared parameter in all permutations, $Z_{N}$ represents the set of all possible permutations of the input sequence $T$, and $z_{T=t}$ and $z_{T<t}$ represents the $t$-th element and the $[1, 2, \ldots, t-1]$ elements of a permutation $z \in Z_{N}$.

MLM represented by BERT can implement bi-directional coding well. However, MLM uses the mask marking during pretraining but not during fine-tuning, which resulted in inconsistent data during pretraining and fine-tuning. To achieve bi-directional coding and avoid the problems of MLM, the permuted LM is proposed. permuted LM is based on the autoregressive LM, which avoids the influence of inconsistent data. However, unlike traditional autoregressive models, permuted LM no longer models sequences in order. It gives all possible permutations of sequences to maximize the expected logarithmic likelihood of the sequence. In this way, any position can take advantage of contextual information from all positions, making permuted LM implement bidirectional encoding.
The most common permuted LM models are XLNET~\cite{DBLP:conf/nips/YangDYCSL19} and MPNet~\cite{DBLP:conf/nips/Song0QLL20}. XLNET is a PFM based on a permuted language modeling approach, which incorporates two crucial techniques from Transformer-XL: relative positional encoding and the segment recurrence mechanism. In contrast, MPNet combines Masked Language Modeling (MLM) and permuted language modeling to predict token dependencies, using auxiliary position information as input to enable the model to view a complete sentence and reduce position differences. These two models represent significant advancements in the field of PFMs.

\subsection{Model Architecture Designing Methods}
ELMO adopts a multi-layer RNN structure. Each layer is a bi-directional LSTM structure composed of a forward and backward LM. The maximum likelihood of these two directions is taken as the objective function. Compared with the word vector method, ELMO introduces contextual information and improves the polysemy problem, but ELMO's overall ability to extract linguistic features is weak.

\begin{figure*}[!t]
    \centering
    \includegraphics[width=\linewidth]{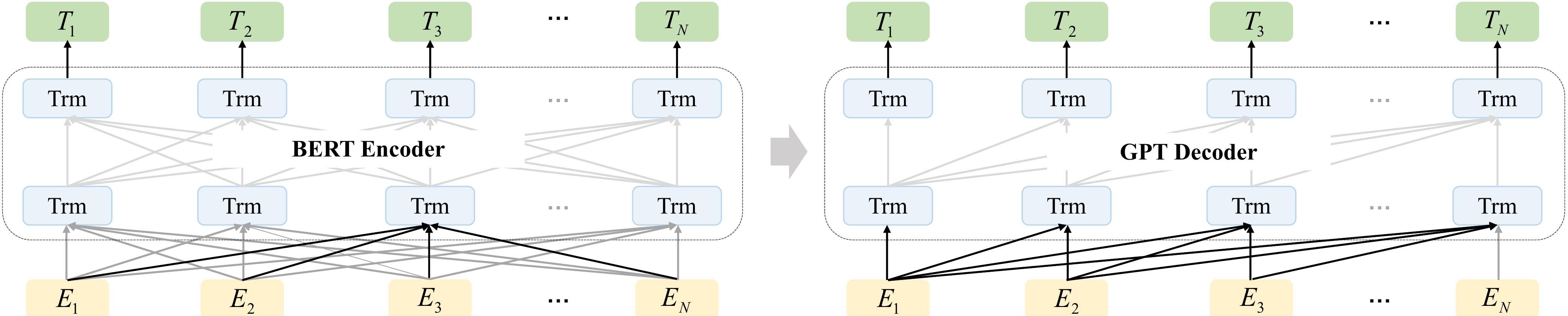}
    \caption{The architectures of BART~\cite{DBLP:conf/acl/LewisLGGMLSZ20}: generalizing BERT (due to the bidirectional encoder), GPT (with the
left-to-right decoder). An autoregressive decoder is used to determine the likelihood of the original document after the corrupted document (on the left) has been encoded using a bidirectional model.}
    \label{ELMO-GPT-BART}
\end{figure*}

The application research of PFMs has two main directions. One is PFMs with fine-tuning (e.g., BERT), and the other one is PFMs with zero/few-shot prompts (e.g., GPT).
\textbf{BERT} uses a bi-directional encoder in Transformer to predict which words are masked and determine whether two sentences are contextual. However, the document is encoded bidirectionally and missing tokens are predicted independently, which reduces the generation ability~\cite{DBLP:conf/acl/LewisLGGMLSZ20}.
\textbf{GPT} uses an autoregressive decoder as a feature extractor to predict the next word based on the first few words and solve downstream tasks using fine-tuning, so it is more suitable for text-generation tasks. However, GPT only uses the former words for prediction, which cannot learn bidirectional interaction information.  

Different from these models, \textbf{BART}~\cite{DBLP:conf/acl/LewisLGGMLSZ20} is a noise-reducing autoencoder built by seq2seq model adopting the encoder-decoder structure, as shown in Fig.~\ref{ELMO-GPT-BART} from ~\cite{DBLP:conf/acl/LewisLGGMLSZ20}. Pretraining mainly includes using noise to destroy text and using the seq2seq model to rebuild the original text.
The encoding layer adopts a bi-directional Transformer.
It adopts five modes of adding noise: (1) single word mask; (2) word deletion; (3) span mask;
(4) sentence rearrangement; (5) document rearrangement.
In the encoder part, the sequence has been masked before inputting it into the encoder. Then, the decoder restores the original sequence according to the encoding representation output by the encoder and the sequence that has not been masked.
The addition of a series of noise patterns makes the performance of BART in sequence generation and natural language reasoning tasks significantly improved.

\subsection{Masking Designing Methods}

The attention mechanism first aggregates essential words into sentence vectors, and vital sentence vectors into text vectors, which allows the model to pay different attention to different inputs~\cite{li2022survey}. 
For BERT, as a bidirectional encoding LM, any two words in an input sentence can see each other.
However, 
it hinders the ability of BERT model to learn NLG tasks. 

\begin{figure*}[!t]
    \centering
    \includegraphics[width=\linewidth]{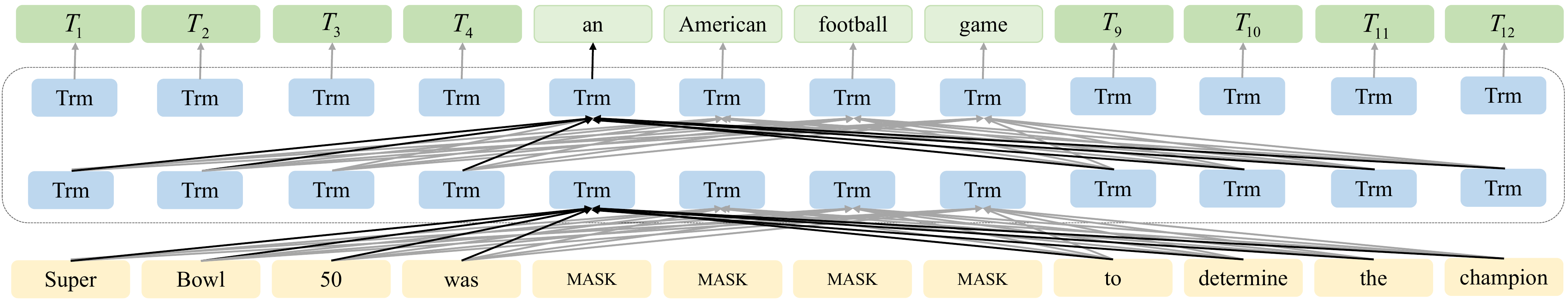}
    \caption{The architecture of SpanBERT~\cite{DBLP:journals/tacl/JoshiCLWZL20}.}
    \label{SpanBERT}
\end{figure*}

Joshi et al.~\cite{DBLP:journals/tacl/JoshiCLWZL20} propose 
SpanBERT based on RoBERTa, which adopts the idea of dynamic masking and single segment pretraining, as shown in Fig.~\ref{SpanBERT} from ~\cite{DBLP:journals/tacl/JoshiCLWZL20}.
The span mask and the Span Boundary Objective (SBO) are also proposed to mask words of a certain length.
The target task of the span-boundary is to restore all the masked span (tokens) by the observed tokens at both ends.
The training stage uses the dynamic mask strategy proposed in the RoBERTa, instead of the mask during the data preprocessing.
Unlike BERT, SpanBERT randomly covers up a continuous text and adds the SBO training target. It predicts the span using the token closest to the span boundary and eliminates the NSP pretraining task.

The BERT and GPT can only separate the training encoder and decoder without joint training in the NLG task. Song et al.~\cite{song2019mass} propose the masked seq2seq pretraining model MASS.
In the training stage, the input sequence of the encoder is randomly masked as a continuous segment of length $k$. The masked segment will be recovered through the MASS decoder.
UniLM~\cite{dong2019unified} completes the learning of the NLG model by designing a different mask for two sentences in the input data.
For the first sentence, UniLM uses the same structure as the Transformer encoder making each word notice its preceding and following words.
For the second sentence, each word can only notice all the words in the first sentence and the preceding words in the current sentence.
Thus, the first and second sentences of the model input form the classic seq2seq pattern.

\subsection{Boosting Methods}
\paragraph{Boosting on Model Performance}
Most of the popular pretraining models need lots of pretraining data, which imposes huge requirements on the hardware, making it challenging to retrain, and only fine-tuning can be done to the model.
To solve these problems, some models appear. For example, ERNIE Tiny released by Baidu is a miniaturized ERNIE~\cite{sun2019ernie}, that reduces the number of layers and increases the prediction speed by 4.3 times with a slight decrease in accuracy.
Lan et al. propose the ALBERT~\cite{lan2019albert} to reduce memory consumption and training speed.
However, it is undeniable that no matter what kind of compression is done for these large-scale models, the performance of the models in these tasks will deteriorate sharply. It requires paying attention to the efficient representation of high-level semantic and grammatical information and lossless compression in future works.
By using word-embedded parameter factorization and hidden parameter sharing between layers, ALBERT significantly reduces the number of parameters of the model without performance loss. It proposes the training task of SOP, which predicts the order of the two sentences to improve the performance.



\paragraph{Boosting for Multi-task Learning}
ERNIE(Baidu)~\cite{sun2019ernie} is mainly composed of two parts, the Transformer encoder and task embedding. In the Transformer encoder, the self-attention mechanism is used to capture the context information of each token and generate context representation embedding. 
Task embedding is a technique that applies different characteristics to a task.
ERNIE 2.0~\cite{sun2020ernie} introduces multi-task learning to realize the pretraining of lexical, grammar, and semantics.
ERNIE 2.0 uses seven different pretraining tasks, covering three aspects: word level, sentence level, and semantic level. It uses continual learning, making the knowledge in the previous training task retained and enabling the model to acquire long-distance memory.
It uses a Transformer encoder and introduces task embedding, enabling the model to distinguish different tasks in the continual learning process.
UniLM~\cite{dong2019unified} uses three pretraining tasks: unidirectional LM, bidirectional LM, and encoder-decoder LM. 
It can simultaneously complete three kinds of target tasks in the pretraining stage through the self-attention layer mask mechanism.
In the training stage, UniLM adopts the small-segment mask strategy proposed by SpanBERT, and the loss function is composed of the loss functions of the above three pretraining tasks. To maintain the contribution consistency on all loss functions, the three pretraining tasks are trained simultaneously.
Modeling and parameter sharing of multiple tasks make LMs achieve good generalization ability in Natural Language Understanding (NLU) and NLG tasks.


\paragraph{Boosting for Different Downstream Tasks}
The pretraining models tend to be large-sized, so how to match different downstream tasks is equally important.
Some pretraining models that are trained on specialized corpora have appeared~\cite{DBLP:journals/taslp/CuiCLQY21, DBLP:conf/emnlp/DiaoBSZW20, tsai2019small}.
Cui et al.~\cite{DBLP:journals/taslp/CuiCLQY21} propose the BERT-whole word masking model (BERT-WWM). They directly use BERT in Chinese to be masked randomly according to the original MLM training, resulting in the loss of semantic information. Since there is no explicit language boundary in Chinese, it is easy to lose significant meaning. ZEN~\cite{DBLP:conf/emnlp/DiaoBSZW20} is a text encoder based on BERT, which adopts N-gram to enhance performance and effectively integrates considerable granular text information with fast convergence speed and good performance.
Tsai et al.~\cite{tsai2019small} propose an oriented multilingual sequence labeling model for sequence labeling tasks. The knowledge distillation method is adopted to achieve better performance in the two tasks: part of speech labeling and morphological attribute prediction for multiple low-resource languages. The inference time is shortened by 27 times.


\paragraph{Examples: ChatGPT and Bard}
\begin{figure*}
    \centering
    \includegraphics[width=1\linewidth]{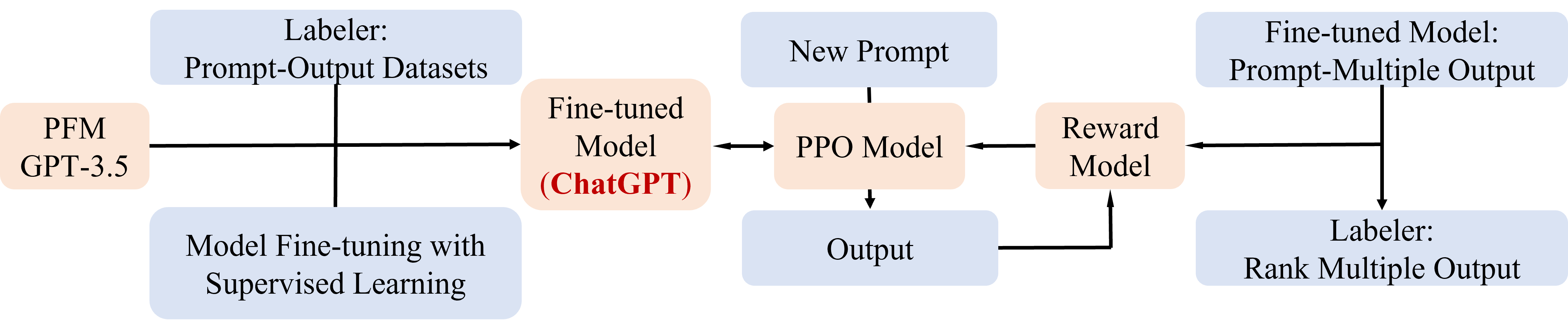}
    \caption{Boosting GPT-3.5 to ChatGPT using Reinforcement Learning from Human Feedback.}
    \label{fig:chatgpt_RL}
\end{figure*}

As shown in Fig.~\ref{fig:chatgpt_RL}, ChatGPT is fine-tuned based on the PFM GPT-3.5 using RLHF. ChatGPT uses a different data collection setup compared to InstructGPT. First, a large dataset with prompts and the desired output behaviors is collected. The dataset is used to fine-tune GPT-3.5 with supervised learning. Second, given the fine-tuned model and a prompt, the model will generate several model outputs. A labeler gives the desired score and ranks the output to compose a comparison dataset, which is used to train the reward model. Finally, the fine-tuned model (ChatGPT) is optimized against the reward model using the Proximal Policy Optimization (PPO)\cite{schulman2017proximal} RL algorithm.

Another experimental conversational PFM, the Bard~\footnote{\url{https://blog.google/technology/ai/bard-google-ai-search-updates/}}, is developed by Google. Bard is based on the LM for Dialogue Applications (LaMDA). LaMDA~\cite{thoppilan2022lamda} is built upon the Transformer, which is pretrained on 1.56T words of dialog data and web text. Safety and factual grounding are two main challenges for conversational AI, LaMDA applies the approaches that fine-tuning with high-quality annotated data and external knowledge sources to improve model performance.  

\begin{table*}[!htbp]
\centering
\caption{Summary of PFMs in NLP. 
The pretraining task includes language model (LM), masked LM (MLM), permuted LM (PLM), denoising autoencoder (DAE), knowledge graphs (KG), and knowledge embedding (KE).
}
\label{text}
\resizebox{\textwidth}{!}{
\begin{tabular}{llllllll}
\hline
\textbf{Year} & \textbf{Conference}       & \textbf{Model}    & \textbf{Architecture}          & \textbf{Embedding}   & \textbf{Training method} & \textbf{Code}                                                                                                                                                                     \\ \hline
2013 & NeurIPS          & Skip-Gram~\cite{mikolov2013distributed}             & Word2Vec               & Probabilistic      & -               & \href{https://github.com/tensorflow/models}{https://github.com/.../models}                                                                                     \\ \hline
2014 & EMNLP            & GloVe~\cite{DBLP:conf/emnlp/PenningtonSM14}         & Word2Vec               & Probabilistic      & -               & -                                                                                                                                                                                 \\ \hline
2015 & NeurIPS          & LM-LSTM~\cite{dai2015semi}                          & LSTM                   & Probabilistic      & LM              & \href{https://github.com/stanfordnlp/GloVe}{https://github.com/.../GloVe}                                                                                      \\ \hline
2016 & IJCAI            & Shared LSTM~\cite{liu2016recurrent}                 & LSTM                   & Probabilistic      & LM              & \href{https://github.com/tensorflow/models/tree/master/research/adversarial\_text}{https://github.com/.../adversarial\_text}                                   \\ \hline
2017 & TACL             & FastText~\cite{bojanowski2017enriching}             & Word2Vec               & Probabilistic      & -               & \href{https://github.com/facebookresearch/fastText}{https://github.com/.../fastText}                                                                           \\ \hline
2017 & NeurIPS          & CoVe~\cite{mccann2017learned}                       & LSTM+Seq2Seq           & Probabilistic      & -               & \href{https://github.com/salesforce/cove}{https://github.com/.../cove}                                                                                         \\ \hline
2018 & NAACL-HLT        & ELMO~\cite{peters2018deep}                          & LSTM                   & Contextual         & LM              & \href{https://allennlp.org/elmo}{https://allennlp.org/elmo}                                                                                                    \\ \hline
2018 & NAACL-HLT        & BERT~\cite{DBLP:conf/naacl/DevlinCLT19}             & Transformer Encoder    & Contextual         & MLM             & \href{https://github.com/google-research/bert}{https://github.com/.../bert}                                                                                    \\ \hline
2018 &                  & OpenAI GPT~\cite{radford2018improving}              & Transformer Decoder    & Autoregressive     & LM              & \href{https://github.com/openai/finetune-transformer-lm}{https://github.com/...transformer-lm}                                                                 \\ \hline
2019 & ACL              & ERNIE(THU)                                                           & Transformer Encoder    & Contextual         & MLM             & \href{https://github.com/PaddlePaddle/ERNIE}{https://github.com/.../ERNIE}                                                                                     \\ \hline
2019 & ACL              & Transformer-XL~\cite{dai2019transformer}            & Transformer-XL         & Contextual         & -               & \href{https://github.com/kimiyoung/transformer-xl}{https://github.com/.../transformer-xl}                                                                      \\ \hline
2019 & ICLR             & InfoWord~\cite{kong2019mutual}                      & Transformer Encoder    & Contextual         & MLM             & -                                                                                                                                                                                 \\ \hline
2019 & ICLR             & StructBERT~\cite{wang2019structbert}                & Transformer Encoder    & Contextual         & MLM             & -                                                                                                                                                                                 \\ \hline
2019 & ICLR             & ALBERT ~\cite{lan2019albert}                        & Transformer Encoder    & Contextual         & MLM             & \href{https://github.com/google-research/ALBERT}{https://github.com/.../ALBERT}                                                                                \\ \hline
2019 & ICLR             & WKLM~\cite{xiong2019pretrained}                     & Transformer Encoder    & Contextual         & MLM             & -                                                                                                                                                                                 \\ \hline
2019 & ICML             & MASS~\cite{song2019mass}                            & Transformer            & Contextual         & MLM(Seq2Seq)    & \href{https://github.com/microsoft/MASS}{https://github.com/.../MASS}                                                                                          \\ \hline
2019 & EMNLP-IJCNLP     & KnowBERT~\cite{peters2019knowledge}                 & Transformer Encoder    & Contextual         & MLM             & \href{https://github.com/allenai/kb}{https://github.com/.../kb}                                                                                                \\ \hline
2019 & EMNLP-IJCNLP     & Unicoder~\cite{huang2019unicoder}                   & Transformer Encoder    & Contextual         & MLM+TLM         & -                                                                                                                                                                                 \\ \hline
2019 & EMNLP-IJCNLP     & MultiFit~\cite{eisenschlos2019multifit}             & QRNN                   & Probabilistic      & LM              & \href{https://github.com/n-waves/multifit}{https://github.com/.../multifit}                                                                                    \\ \hline
2019 & EMNLP-IJCNLP     & SciBERT~\cite{beltagy2019scibert}                   & Transformer Encoder    & Contextual         & MLM             & \href{https://github.com/allenai/scibert}{https://github.com/.../scibert}                                                                                      \\ \hline
2019 & EMNLP-IJCNLP     & BERT-PKD~\cite{sun2019patient}                      & Transformer Encoder    & Contextual         & MLM             & \href{https://github.com/intersun/PKD-for-BERT-Model-Compression}{https://github.com/...Compression}                                                           \\ \hline
2019 & NeurIPS          & Xlnet~\cite{DBLP:conf/nips/YangDYCSL19}             & Transformer-XL Encoder & Permutation        & PLM             & \href{https://github.com/zihangdai/xlnet}{https://github.com/.../xlnet}                                                                                        \\ \hline
2019 & NeurIPS          & UNILM~\cite{dong2019unified}                        & LSTM + Transformer     & Contextual         & LM + MLM        & \href{https://github.com/microsoft/unilm}{https://github.com/.../unilm}                                                                                        \\ \hline
2019 & NeurIPS          & XLM~\cite{lample2019cross}                          & Transformer Encoder    & Contextual         & MLM+CLM+TLM     & \href{https://github.com/facebookresearch/XLM}{https://github.com/.../XLM}                                                                                     \\ \hline
2019 & OpenAI Blog      & GPT-2~\cite{radford2019language}                    & Transformer Decoder    & Autoregressive     & LM              & \href{https://github.com/openai/gpt-2}{https://github.com/.../gpt-2}                                                                                           \\ \hline
2019 & arXiv            & RoBERTa~\cite{DBLP:journals/corr/abs-1907-11692}                       & Transformer Encoder    & Contextual         & MLM             & \href{https://github.com/pytorch/fairseq}{https://github.com/.../fairseq}                                                                                      \\ \hline
2019 & arXiv            & ERNIE(Baidu)~\cite{sun2019ernie}                    & Transformer Encoder    & Contextual         & MLM+DLM         & \href{https://github.com/PaddlePaddle/ERNIE}{https://github.com/.../ERNIE}                                                                                     \\ \hline
2019 & EMC2@NeurIPS            & Q8BERT~\cite{DBLP:conf/nips/ZafrirBIW19}                      & Transformer Encoder    & Contextual         & MLM             & \href{https://github.com/IntelLabs/nlp-architect/blob/master/nlp\_architect/models/transformers/quantized\_bert.py}{https://github.com/.../quantized\_bert.py} \\ \hline
2019 & arXiv            & DistilBERT~\cite{sanh2019distilbert}                & Transformer Encoder    & Contextual         & MLM             & \href{https://github.com/huggingface/transformers/tree/master/examples/research\_projects/distillation}{https://github.com/.../distillation}                   \\ \hline
2020 & ACL              & fastBERT~\cite{liu2020fastbert}                     & Transformer Encoder    & Contextual         & MLM             & \href{https://github.com/autoliuweijie/FastBERT}{https://github.com/.../FastBERT}                                                                              \\ \hline
2020 & ACL              & SpanBERT~\cite{DBLP:journals/tacl/JoshiCLWZL20}     & Transformer Encoder    & Contextual         & MLM             & \href{https://github.com/facebookresearch/SpanBERT}{https://github.com/.../SpanBERT}                                                                           \\ \hline
2020 & ACL              & BART~\cite{DBLP:conf/acl/LewisLGGMLSZ20}            & Transformer            & En: Contextual     & DAE             & \href{https://github.com/huggingface/transformers}{https://github.com/.../transformers}                                                                        \\
     &                  &                                                                      &                        & De: Autoregressive &                 &                                                                                                                                                                                   \\ \hline
2020 & ACL              & CamemBERT~\cite{DBLP:conf/acl/MartinMSDRCSS20}      & Transformer Encoder    & Contextual         & MLM(WWM)        & \href{https://camembert-model.fr}{https://camembert-model.fr}                                                                                                  \\ \hline
2020 & ACL              & XLM-R~\cite{DBLP:conf/acl/ConneauKGCWGGOZ20}        & Transformer Encoder    & Contextual         & MLM             & \href{https://github.com/facebookresearch/XLM}{https://github.com/.../XLM}                                                                                     \\ \hline
2020 & ICLR             & Reformer~\cite{DBLP:conf/iclr/KitaevKL20}           & Reformer               & Permutation        & -               & \href{https://github.com/google/trax/tree/master/trax/models/reformer}{https://github.com/.../reformer}                                                        \\ \hline
2020 & ICLR             & ELECTRA~\cite{clark2020electra}                     & Transformer Encoder    & Contextual         & MLM             & \href{https://github.com/google-research/electra}{https://github.com/.../electra}                                                                              \\ \hline
2020 & AAAI             & Q-BERT~\cite{shen2020q}                             & Transformer Encoder    & Contextual         & MLM             & -                                                                                                                                                                                 \\ \hline
2020 & AAAI             & XNLG~\cite{chi2020cross}                            & Transformer            & Contextual         & MLM+DAE         & \href{https://github.com/CZWin32768/xnlg}{https://github.com/.../xnlg}                                                                                         \\ \hline
2020 & AAAI             & K-BERT~\cite{liu2020k}                              & Transformer Encoder    & Contextual         & MLM             & \href{https://github.com/autoliuweijie/K-BERT}{https://github.com/.../K-BERT}                                                                                  \\ \hline
2020 & AAAI             & ERNIE 2.0~\cite{sun2020ernie}                       & Transformer Encoder    & Contextual         & MLM             & \href{https://github.com/PaddlePaddle/ERNIE}{https://github.com/.../ERNIE}                                                                                     \\ \hline
2020 & NeurIPS          & GPT-3~\cite{brown2020language}                      & Transformer Decoder    & Autoregressive     & LM              & \href{https://github.com/openai/gpt-3}{https://github.com/.../gpt-3}                                                                                           \\ \hline
2020 & NeurIPS          & MPNet~\cite{DBLP:conf/nips/Song0QLL20}              & Transformer Encoder    & Permutation        & MLM+PLM         & \href{https://github.com/microsoft/MPNet}{https://github.com/.../MPNet}                                                                                        \\ \hline
2020 & NeurIPS          & ConvBERT~\cite{DBLP:conf/nips/JiangYZCFY20}         & Mixed Attention        & Contextual         & -               & \href{https://github.com/yitu-opensource/ConvBert}{https://github.com/.../ConvBert}                                                                            \\ \hline
2020 & NeurIPS          & MiniLM~\cite{DBLP:conf/nips/WangW0B0020}            & Transformer Encoder    & Contextual         & MLM             & \href{https://github.com/microsoft/unilm/tree/master/minilm}{https://github.com/.../minilm}                                                                    \\ \hline
2020 & TACL             & mBART~\cite{DBLP:journals/tacl/LiuGGLEGLZ20}        & Transformer            & Contextual         & DAE             & \href{https://github.com/pytorch/fairseq/tree/master/examples/mbart}{https://github.com/.../mbart}                                                             \\ \hline
2020 & COLING           & CoLAKE~\cite{DBLP:conf/coling/SunSQGHHZ20}          & Transformer Encoder    & Contextual         & MLM+KE          & \href{https://github.com/txsun1997/CoLAKE}{https://github.com/.../CoLAKE}                                                                                      \\ \hline
2020 & LREC             & FlauBERT~\cite{DBLP:conf/lrec/LeVFSCLACBS20}        & Transformer Encoder    & Contextual         & MLM             & \href{https://github.com/getalp/Flaubert}{https://github.com/.../Flaubert}                                                                                     \\ \hline
2020 & EMNLP            & GLM~\cite{DBLP:conf/emnlp/ShenMHLTC20}              & Transformer Encoder    & Contextual         & MLM+KG          & \href{https://github.com/THUDM/GLM}{https://github.com/.../GLM}                                                                                                \\ \hline
2020 & EMNLP (Findings) & TinyBERT~\cite{DBLP:conf/emnlp/JiaoYSJCL0L20}       & Transformer            & Contextual         & MLM             & \href{https://github.com/huawei-noah/Pretrained-Language-Model/tree/master/TinyBERT}{https://github.com/.../TinyBERT}                                          \\ \hline
2020 & EMNLP (Findings) & RobBERT~\cite{DBLP:conf/emnlp/DelobelleWB20}        & Transformer Encoder    & Contextual         & MLM             & \href{https://github.com/iPieter/RobBERT}{https://github.com/.../RobBERT}                                                                                      \\ \hline
2020 & EMNLP (Findings) & ZEN~\cite{DBLP:conf/emnlp/DiaoBSZW20}               & Transformer Encoder    & Contextual         & MLM             & \href{https://github.com/sinovation/ZEN}{https://github.com/.../ZEN}                                                                                           \\ \hline
2020 & EMNLP (Findings) & BERT-MK~\cite{DBLP:conf/emnlp/HeZXJLYX20}           & KG-Transformer Encoder & Contextual         & MLM             & -                                                                                                                                                                                 \\ \hline
2020 & RepL4NLP@ACL     & CompressingBERT~\cite{DBLP:conf/rep4nlp/GordonDA20} & Transformer Encoder    & Contextual         & MLM(Pruning)    & \href{https://github.com/mitchellgordon95/bert-prune}{https://github.com/.../bert-prune}                                                                       \\ \hline
2020 & JMLR             & T5~\cite{DBLP:journals/jmlr/RaffelSRLNMZLL20}       & Transformer            & Contextual         & MLM(Seq2Seq)    & \href{https://github.com/google-research/text-to-text-transfer-transformer}{https://github.com/...transformer}                                                 \\ \hline
2021 & T-ASL            & BERT-wwm-Chinese~\cite{DBLP:journals/taslp/CuiCLQY21}                  & Transformer Encoder    & Contextual         & MLM             & \href{https://github.com/ymcui/Chinese-BERT-wwm}{https://github.com/...BERT-wwm}                                                                      \\ \hline
2021 & EACL             & PET~\cite{DBLP:conf/eacl/SchickS21}                 & Transformer Encoder    & Contextual         & MLM             & \href{https://github.com/timoschick/pet}{https://github.com/.../pet}                                                                                           \\ \hline
2021 & TACL             & KEPLER~\cite{DBLP:journals/tacl/WangGZZLLT21}       & Transformer Encoder    & Contextual         & MLM+KE          & \href{https://github.com/THU-KEG/KEPLER}{https://github.com/.../KEPLER}                                                                                        \\ \hline
2021 & EMNLP   & SimCSE~\cite{DBLP:journals/corr/abs-2104-08821}     & Transformer Encoder    & Contextual         & MLM+KE          & \href{https://github.com/princeton-nlp/SimCSE}{https://github.com/.../SimCSE}                                                                                  \\ \hline
2021 & ICML   & GLaM~\cite{du2022glam}     & Transformer    & Autoregressive         & LM          & -                                                                                  \\ \hline
2021 & arXiv   & XLM-E~\cite{chi2021xlm}     & Transformer    & Contextual         & MLM          &                                                                                   \\ \hline
2021 & arXiv   & T0~\cite{sanh2021multitask}     & Transformer    & Contextual         & MLM          & \href{https://github.com/bigscience-workshop/t-zero}{https://github.com/.../T0}                                                                                  \\ \hline
2021 & arXiv   & Gopher~\cite{rae2021scaling}     & Transformer    & Autoregressive         & LM          & -                                                                                 \\ \hline
2022 & arXiv   & MT-NLG~\cite{smith2022using}     & Transformer   & Contextual         & MLM          & -                                                                                 \\ \hline
2022 & arXiv   & LaMDA~\cite{thoppilan2022lamda}     & Transformer Decoder   & Autoregressive         & LM          & \href{https://github.com/conceptofmind/LaMDA-rlhf-pytorch}{https://github.com/.../LaMDA}                                                                                  \\ \hline
2022 & arXiv   & Chinchilla~\cite{hoffmann2022training}     & Transformer   & Autoregressive         & LM          & -                                                                                  \\ \hline
2022 & arXiv   & PaLM~\cite{chowdhery2022palm}     & Transformer   & Autoregressive         & LM         & \href{https://github.com/lucidrains/PaLM-pytorch}{https://github.com/.../PaLM}                                                                                  \\ \hline
2022 & arXiv   & OPT~\cite{zhang2022opt}     & Transformer Decoder  & Autoregressive         & LM          & \href{https://github.com/facebookresearch/metaseq}{https://github.com/.../MetaSeq}                                                                                  \\ \hline

\end{tabular}
}
\end{table*}

\subsection{Instruction-Aligning Methods}

Instruction-aligning methods aim to let the LM follow human intents and generate meaningful outputs. The general approach is fine-tuning the pretrained LM with high-quality corpus in a supervised manner. To further improve the usefulness and harmlessness of LMs, some works introduce RL into the fine-tuning procedure so that LMs could revise their responses according to human or AI feedback. Both supervised and RL approaches can leverage chain-of-thought \cite{weichain} style reasoning to improve the human-judged performance and transparency of AI decision-making.

\paragraph{Supervised Fine-Tuning (SFT)} SFT is a well-established technique to unlock knowledge and apply it to specific real-world, even unseen tasks. The template for SFT is composed of input-output pairs and an instruction \cite{weifinetuned}. For example, given the instruction ``Translate this sentence to Spanish:'' and an input ``The new office building was built in less than three months.'', we want the LM to generate the target ``El nuevo edificio de oficinas se construyó en tres meses.''. The template is commonly humanmade including unnatural instructions \cite{honovich2022unnatural} and natural instructions \cite{wang2022super, mishra2022cross}, or bootstrap based on a seed corpus \cite{wang2022self}. Ethical and social risks of harm from LMs are significant concerns in SFT \cite{weidinger2021ethical}. LaMDA, the largest LM to date, thus relies on crowdworker annotated data for providing a safety assessment of any generated LaMDA response in three conversation categories: natural, sensitive, and adversarial. The list of rules serves further safety fine-tuning and evaluation purposes.

\paragraph{Reinforcement Learning from Feedback} RL has been applied to enhance various models in NLP tasks such as machine translation \cite{kiegeland2021revisiting}, summarization \cite{stiennon2020learning}, dialogue generation \cite{jaques2020human}, image captioning \cite{rennie2017self}, question generation \cite{pang2020text}, text-games \cite{hausknecht2020interactive}, and more \cite{snell2022offline, lu2022quark, uc2022survey}. RL is a helpful method for optimizing non-differentiable objectives in language generation tasks by treating them as sequential decision-making problems. However, there is a risk of overfitting to metrics that use neural networks, leading to nonsensical samples that score well on the metrics \cite{ramamurthy2022reinforcement}. RL is also used to align LMs with human preferences \cite{wu2021recursively, nakano2021webgpt, glaese2022improving}. 

InstructGPT proposes to fine-tune large models with PPO against a trained reward model to align LMs with human preference \cite{ouyang2022training}, which is the same method applied by ChatGPT 
named RLHF. Specifically, the reward model is trained with comparison data of human labelers' manual rankings
of outputs. For each of them, the reward model or machine labeler calculates a reward, which is used to update the LM using PPO. More details are illustrated in Fig. \ref{fig:chatgpt_RL}. 

One of the recent breakthroughs in PFM technology is GPT-4~\cite{openai2023gpt4}, which follows a pretraining approach to predict the subsequent token in a document and then undergoes RLHF fine-tuning. 
As the task complexity increases, GPT-4 outperforms GPT-3.5 in terms of reliability, creativity, and capability to handle more nuanced instructions.

Sparrow \cite{glaese2022improving}, developed by DeepMind, also utilizes RLHF that reduces the risk of unsafe and inappropriate answers. Despite some promising results using RLHF by incorporating fluency, progress in this field is impeded by a lack of publicly available benchmarks and implementation resources, resulting in a perception that RL is a difficult approach for NLP. Therefore, an open-source library named RL4LMs~\cite{ramamurthy2022reinforcement} is introduced recently, which consists of building blocks for fine-tuning and evaluating RL algorithms on LM-based generation. 

Besides human feedback, one of the latest dialogue agents --  Claude favors Constitutional AI \cite{bai2022constitutional} where the reward model is learned via RL from AI Feedback (RLAIF). Both the critiques and the AI feedback are steered by a small set of principles drawn from a ‘constitution’, the specification of a short list of principles or instructions, which is the only thing provided by humans in Claude. The AI feedback focuses on controlling the outputs to be less harmful by explaining its objections to dangerous queries.

\paragraph{Chain-of-Thoughts}
Chain-of-thought (CoT) prompting is a technique for improving the reasoning ability of LLMs by prompting them to generate a series of intermediate steps that lead to the final answer of a multi-step problem. The CoT is a series of intermediate reasoning steps, which can significantly improve the ability of LLMs to perform complex reasoning \cite{weichain, chung2022scaling, kojima2022large}. Besides, fine-tuning with CoT shows slightly more harmless compared to without CoT \cite{bai2022constitutional}. CoT prompting is an emergent property of model scale, meaning it works better with larger and more powerful language models. It is also possible to fine-tune models on CoT reasoning datasets to enhance this capability further and stimulate better interpretability.

In a CoT prompting experiment, a prompt is provided to the model that outlines a multi-step problem. The prompt might pose a question such as ``After selling 30 out of his 100 chickens and 10 out of his 20 pigs, how many animals does a farmer have left?'' The model then generates a sequence of intermediate reasoning steps, for example, ``The farmer has 100-30=70 chickens remaining'' and ``The farmer has 20-10=10 pigs remaining,'' before generating the final answer, such as ``The farmer has 70+10=80 animals remaining.'' CoT prompting has demonstrated its efficacy in improving the performance of LLMs on various reasoning tasks, such as arithmetic, symbolic reasoning, and common sense. It is a promising technique that can enhance the ability of language models to reason about complicated problems.


\subsection{Summary}

The neural probabilistic LM uses a neural network to estimate the parameters of the probabilistic LM, which reduces the size of the model parameters while enlarging the number of context windows. With the help of a neural network, the LM does not need to improve the smoothing algorithm to alleviate the performance bottleneck continuously.
Since the training target is unsupervised, a corpus with a large amount of data is enough for training. The negative sampling technique in the training process provides a new idea for the follow-up study of the target task in the LM.
Furthermore, the neural probabilistic LM promotes the further development of downstream task research because of its good representation capability and training efficiency.
After the pretraining LM, especially the BERT model, is proposed, the research in language modeling has entered a new phase. 
The bidirectional LM, the hidden LM, and the sorted LM adopted by the bidirectional LM have successfully modeled the grammatical and semantic information in natural language at a deeper level.
ChatGPT is another milestone work in PFMs using RL. 
The presentation ability of PFMs is qualitatively better than that of the neural probabilistic LM. It even exceeds that of humans in some tasks.

\section{PFMs for Computer Vision} \label{Section 4}

With the popularity of PFM used in NLP, it motivates researchers to start exploring PFM in CV.
The term ``pretraining'' has not been clearly defined within the realm of deep learning research in CV. This word is first used in convolution-based networks when we adjust the parameters on a more general dataset such as ImageNet, which can make other tasks train to start with a warm-up initialization and thus converge with faster speed. 
In contrast to early CNN-based transfer learning techniques that rely on pretrained datasets with supervised signals, our examination of PFM centers on SSL which utilizes human-designed labels, such as Jigsaw puzzles, or the comparison of different patches from images as pretext tasks. This allows for learned representations to be generalized to various downstream tasks, including classification, detection, recognition, segmentation, etc.

However, it is costly to rely on data annotations when the learning tasks become more complicated, making the labeling process more arduous and time-consuming than the actual learning. This is where SSL is urgently needed and how it can further fuel the progress of deep learning methods. To reduce the dependency on data labeling, unlabeled data are trained with self-supervision by matching, contrasting, or generating in SSL.


\begin{figure*}
    \centering
    \includegraphics[width=1\linewidth]{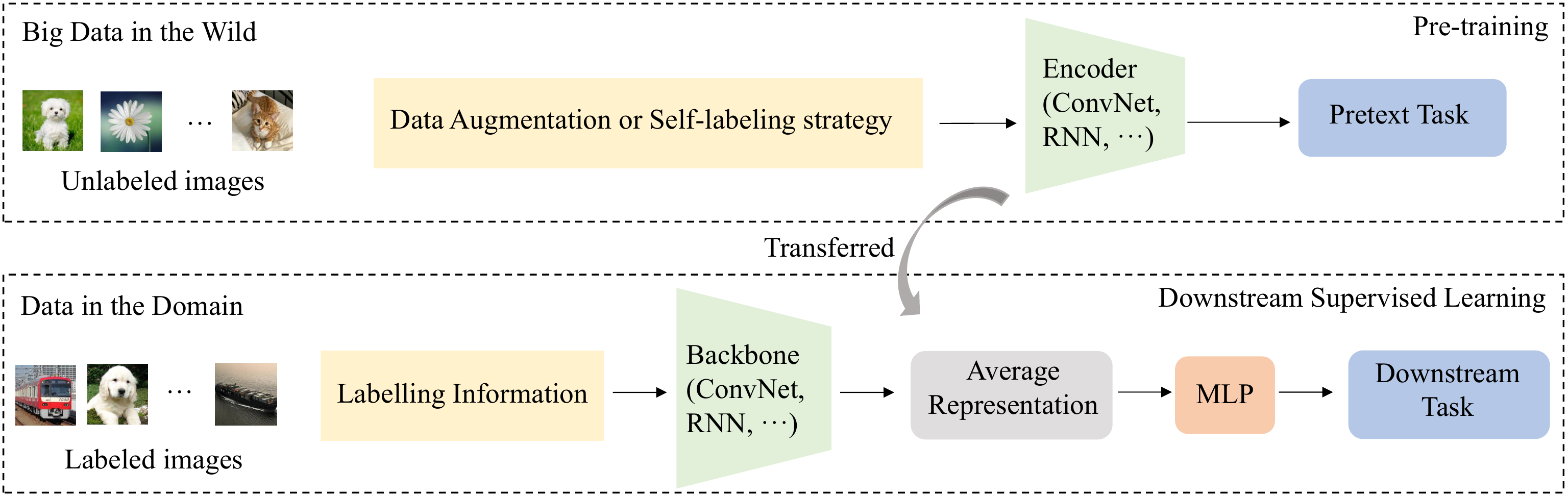}
    \caption{The general pipeline for SSL. The top part represents the pretraining, and the bottom stream obtains transferred parameters from above to learn downstream supervised tasks.}
    \label{fig:pipline}
\end{figure*}

The general pipeline of SSL is shown in Fig. \ref{fig:pipline}. During the pretraining stage, a pretext task is designed for the encoder networks to solve. The artificial labels for this pretext task are automatically generated based on specific attributes of the data, such as image patches from the same origin being labeled as ``positive'' and those from different origins as ``negative''. Then, the encoder networks are trained to solve the pretext task by supervised learning methods. Since shallow layers extract fine-grained details such as edges, angles, and textures, while deeper layers capture task-related high-level features such as semantic information or image contents, learned encoders on pretext tasks can be transferred to downstream supervised tasks. During this stage, the parameters of the backbone are fixed, and only a simple classifier, such as a two-layer Multi-Layer Perceptron (MLP), needs to be learned. Considering the limited workload in the downstream training stage, this learning process is commonly referred to as fine-tuning. In summary, the representations learned during the pretraining stage in SSL can be reused on other downstream tasks and achieve comparable results.  

In this section, we introduce different tasks for pretraining PFMs in CV. The PFMs can be trained by specific pretext tasks, frame order, generation, reconstruction, memory bank, sharing, clustering and so on. We summarize the PFMs proposed in CV in \textbf{Table~\ref{tab:pretraining model for image}}. 


\subsection{Learning by Specific Pretext Task}
In the early stage of unsupervised learning, the network is trained by designing a special pretext task and predicting the answer to this task. Dosovitskiy et al.~\cite{dosovitskiy2014discriminative, DFB16} pretrain the Exemplar CNN to discriminate the different patches from the unlabelled data. The experiments prove the designs can learn useful representations transferred to the standard recognition assignments. In the method based on context prediction~\cite{doersch2015unsupervised}, a handcrafted supervised signal about the position information serves as the label for the pair classification. 
Inpainting~\cite{pathak2016context} aims to pretrain models by predicting the missed center part. 
Because inpainting is a semantic-based prediction, another decoder is linked to the context encoder in this manner. Furthermore, the standard pixel-by-pixel reconstruction process of the decoder can be transferred to any other downstream inpainting tasks.
Specifically, Colorization~\cite{zhang2016colorful} is a method that evaluates how colorization as a pretext task can help to learn semantic representation for downstream tasks. It is also known as the \emph{cross-channel encoding} since different image channels serve as input and the output is discriminated. Similarly, Split-Brain	Autoencoder~\cite{zhang2017split} also learns representations in a self-supervised way by forcing the network to solve cross-channel prediction tasks.
Jigsaw~\cite{noroozi2016unsupervised} is proposed to pretrain the designed Context-Free Network (CFN) in a self-supervised manner by first designing the Jigsaw puzzle as a pretext task.  Completing Damaged Jigsaw Puzzles (CDJP)~\cite{kim2018learning} learns image representation by complicating pretext tasks furthermore, in which puzzles miss one piece and the other pieces contain incomplete color. Following the idea of designing efficient and effective pretext tasks, Noroozi et al.~\cite{noroozi2017representation} use counting visual primitives as a special pretext task and outperform previous SOTA models on regular benchmarks.
NAT~\cite{bojanowski2017unsupervised} learns representation by aligning the output of backbone CNN to low-dimensional noise.
RotNet~\cite{gidaris2018unsupervised} is designed to predict different rotations of images.

\begin{figure*}[!htp]
    \centering
    \includegraphics[width=0.9\linewidth]{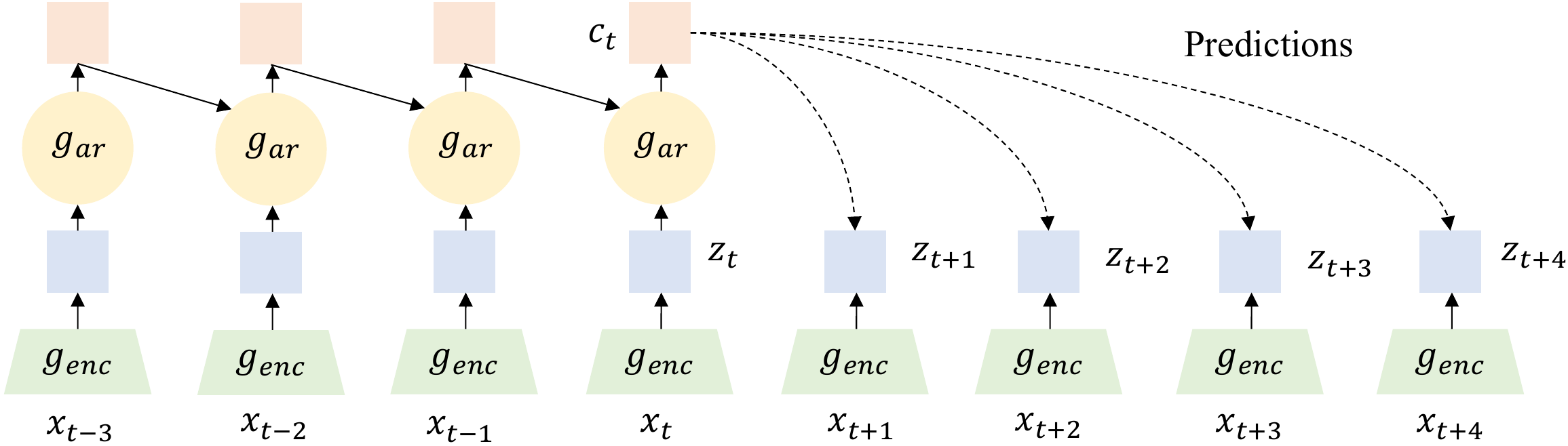}
    \caption{Contrastive Predictive Coding~\cite{oord2018representation}. The input sequence can represent both images and videos.}
    \label{fig:cpc}
\end{figure*}

\subsection{Learning by Frame Order}
The learning of sequence data such as videos always involves frame processing through time steps. This problem often connects with solving pretext tasks that can help to learn visual temporal representations. Contrastive Predictive Coding (CPC)~\cite{oord2018representation} is the first model to learn data representations by predicting the future in latent space. This model can be fed with data in any modalities, like speech, images, text, etc.
The components of CPC are shown in Fig. \ref{fig:cpc} from~\cite{oord2018representation}, where the $x_t$ represents the input sequence of observations, $z_t$ is a sequence of latent representations after the encoder $g_{enc}$, and $c_t$ is a context latent representation that summarizes all the latent sequence $z_{\le t}$ after an autoregressive model $g_{ar}$. Unlike the traditional model predicts future frames $x_{t+k}$ by a generative model $p_k(x_{t+k}|c_t)$, CPC models a "density ratio" $f_k$ to represent the mutual information between the context latent representation $c_t$ and future frame $x_{t+k}$:
\begin{equation}
    f_k(x_{t+k},c_t)\propto p(x_{t+k}|c_t) / x_{t+k}.
\end{equation}
After the encoding of recurrent neural networks, $z_t$ and $c_t$ can both be chosen
for the downstream tasks as needed. The encoder and autoregressive model are trained by InfoNCE~\cite{oord2018representation} as follows
\begin{equation}
    \mathcal{L}=-\mathbbm{E}_X[\log f_k(x_{t+k},c_t) / \sum\nolimits_{x_j\in X}f_k(x_j,c_t)],
\end{equation}
where $X$ denotes the training dataset containing both positive and negative samples. The density ratio $f_k$ can be estimated by optimizing $\mathcal{L}$.
CPC v2 revisits and improves CPC~\cite{henaff2020data} by pretraining on unsupervised representations, and its representation generality can be transferred to data-efficient downstream tasks.

\subsection{Learning by Generation}\label{sec:Learning by Generation}
Although many existing applications are popular after the development of the GAN-based approach, the representation abilities inside the GANs are not entirely exploited due to the absence of a feature encoder. 
Thus, Bidirectional Generative Adversarial Networks (BiGANs)~\cite{donahue2016adversarial} is proposed to project data back into the latent space, which is useful for auxiliary supervised discrimination tasks via serving as feature representations. 

\begin{figure*}[t]
    \centering
    \includegraphics[width=0.8\linewidth]{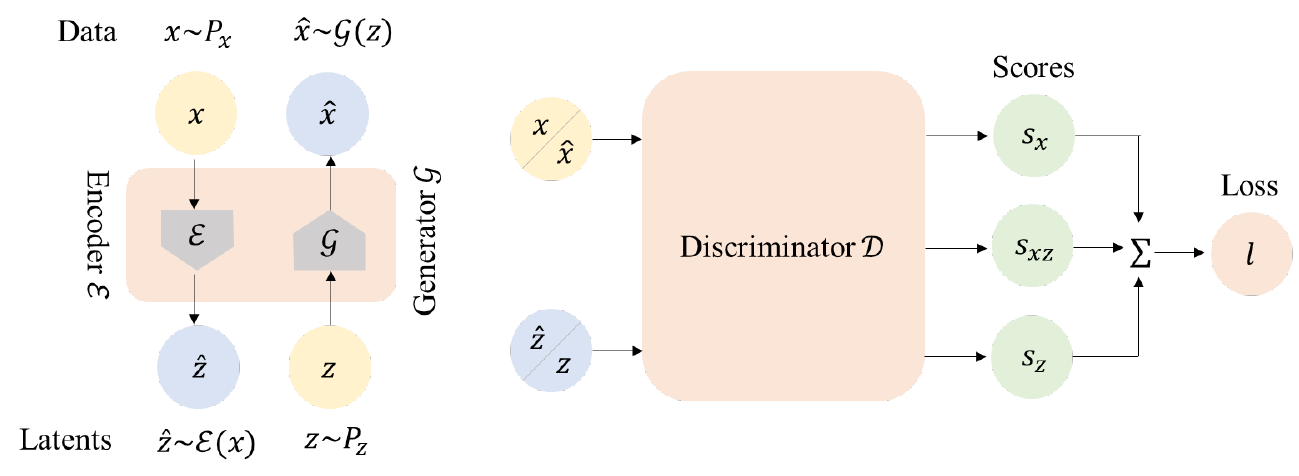}
    \caption{The structure of the BigBiGAN framework~\cite{DBLP:conf/nips/DonahueS19}.}
    \label{fig:bigbigan}
\end{figure*}

Based on BiGANs, BigBiGAN~\cite{DBLP:conf/nips/DonahueS19} first achieves the SOTA in unsupervised representation learning on ImageNet by adding an encoder and modifying the discriminator. As shown in Fig. \ref{fig:bigbigan} from~\cite{DBLP:conf/nips/DonahueS19}, the traditional components of GANs (encoder $\mathcal{E}$ and generator $\mathcal{G}$) are used to produce data-latent pairs, denoted as $(\textbf{x}\sim P_{\textbf{x}},\hat{\textbf{z}}\sim\mathcal{E}(\textbf{x}))$ and $(\hat{\textbf{x}}\sim\mathcal{G}(\textbf{z}),\textbf{z}\sim P_{\textbf{z}})$. The final loss $\ell$ is defined as the sum of data-specific term $s_{\textbf{x}},s_{\textbf{z}}$ and data-joint term $s_{\textbf{xz}}$. The introduced discriminator $\mathcal{D}$ (Adversarially Learned Inference (ALI)~\cite{dumoulin2016adversarially}, or BiGAN~\cite{donahue2016adversarial}) learns to discriminate between pairs from the raw data, latent distribution and encoded vector. 

\subsection{Learning by Reconstruction}
The iGPT \cite{chen2020generative} and ViT \cite{dosovitskiy2020image} models have demonstrated the feasibility of adapting the pretext task of masked prediction using auto-encoder from language to image data. BEiT \cite{bao2021beit} is the first to demonstrate that autoencoder-based masked prediction can outperform DINO \cite{caron2021emerging}, a conventional SOTA method without pretraining techniques. Specifically, BEiT consists of two stages: token embedding with discrete variational autoencoder (dVAE) \cite{ramesh2021zero}, and tokenizer training with masked image prediction. In the first stage, the original image is split into some patches and encoded using discrete tokens, which is different from BERT since image patches don't have off-the-shelf tokens as words in NLP. In the second stage, the BEiT encoder takes a corrupted image containing unmasked and masked patches, and then the visual tokens of the masked patches are outputted to match the corresponding visual tokens from the fixed tokenizer. Despite its success, the separation between masked prediction and autoencoder training induces that the whole framework is not end-to-end and hinders learning effectiveness and efficiency. 

\begin{figure*}
    \centering
    \includegraphics[width=0.8\linewidth]{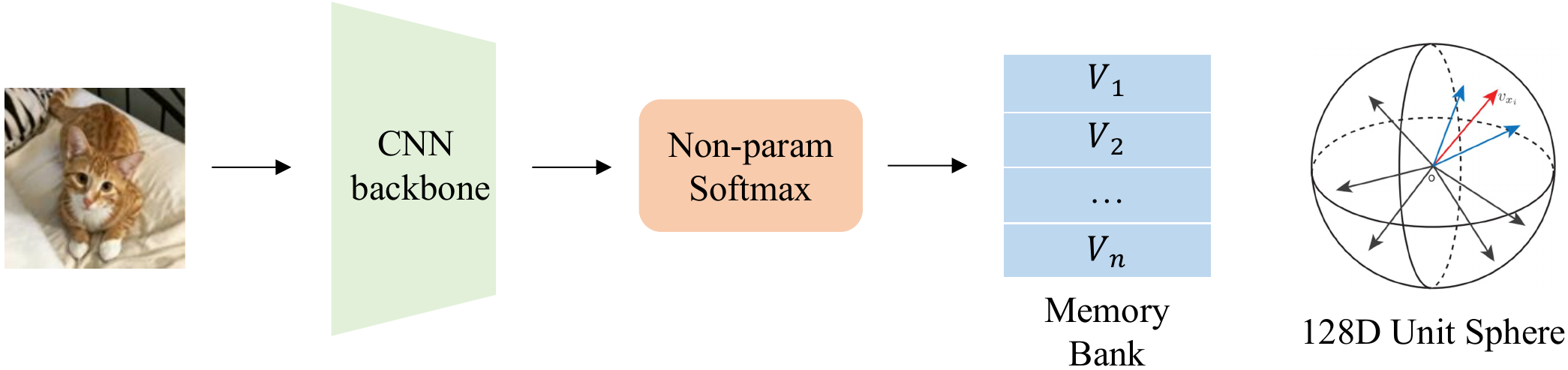}
    \caption{The general pipeline for the Memory Bank Method~\cite{wu2018unsupervised}.}
    \label{fig:memorybank}
\end{figure*}

To migrate this issue, MAE \cite{he2022masked} proposes an end-to-end simple solution by predicting the masked patches directly from the unmasked ones with the Mean Squared Error (MSE) loss. It's worth noting that MAE uses a masking ratio of 75\%, which is significantly higher than that of BERT (typically 15\%). Ablation study suggests that higher masking ratios are beneficial for both fine-tuning and linear probing. Concurrently, SimMIM \cite{xie2022simmim} proposes a similar autoencoder-based solution as MAE, in which they also confirm that a higher marking ratio and leveraging random masking strategy helps improve performance. The major difference is how they partition the responsibility of representation encoding and pretext prediction in the autoencoder. Since the decoder of SimMIM is simple, the encoder of SimMIM synchronously conducts both of them. On the contrary, the encoder in MAE solely undertakes the role of representation encoding, and the decoder is responsible for pretext prediction. Recently, Meta AI announces the Segment Anything Model (SAM)~\cite{kirillov2023segment} which prompts users to specify what to segment in an image, allowing for a wide range of segmentation tasks without the need for additional training. SAM employs an MAE pretrained ViT-H~\cite{dosovitskiy2020image} image encoder that runs once per image and produces an image embedding, as well as a prompt encoder that embeds input prompts such as clicks or boxes. Following that, a lightweight transformer-based mask decoder predicts object masks from image and prompt embeddings. The results show that SAM can generate high-quality masks from a single foreground point that are typically just modestly inferior to the manually annotated ground truth. It routinely achieves strong quantitative and qualitative outcomes on a wide range of downstream tasks using a zero-shot transfer approach and prompt engineering.

Leveraging ViT in MAE poses a serious inefficiency issue, where decreasing the patch size results in a quadratic increase in computing resources. To address the problem, there are two important solutions: (1) hierarchical ViT and (2) local attention. In the first direction, hierarchical ViT (hViT) was introduced, which utilizes a shrinking pyramid structure and techniques like shifted windows \cite{liu2021swin} to reduce computational demands. Unfortunately, hViT cannot be directly applied to enable MAE pretraining because the local window attention used in hViT makes it difficult to handle randomly masked patches as in MAE. Recently, Uniform Masking MAE (UM-MAE) \cite{li2022uniform} is proposed to empower MAE with hViTs, which introduces a two-stage pipeline: sampling and masking. It starts by randomly sampling a portion of patches (25\% reported in the paper) from each block, and then follows by masking additional patches on top of the sampled ones. 
The first step helps to maintain common elements across different local windows, while the second step prevents shortcuts for pixel reconstruction from nearby low-level features, making the task more difficult. Another direction to improve efficiency focuses on reducing the input size by putting the attention of the network into some local small windows of the image. Motivated by the observation that local knowledge is sufficient for reconstructing masked patches, Local masked reconstruction (LoMaR) \cite{chen2022efficient} was proposed. Rather than using the entire image for mask reconstruction, LoMaR samples a number of small windows and focuses attention on local regions, which outperforms MAE on downstream tasks in terms of learning efficiency.



\subsection{Learning by Memory Bank}
Non-Parametric Instance Discrimination (NPID)~\cite{wu2018unsupervised} is the first method that utilizes the instances to learn representations for downstream tasks. The detailed pipeline is shown in Fig. \ref{fig:memorybank}. The feature representations are stored in the memory bank for the convenience of computation because the instance-level classification objective needs all images in the training dataset. For any image $x$ with feature representation $\textbf{v}=f_\theta(x)$, its probability of being recognized as $i$-th example is:
\begin{equation}
    P(i|\textbf{v}) = exp(\textbf{v}_i^{\mathrm{T}}\textbf{v}/\tau) / \sum\nolimits_{j=1}^n exp(\textbf{v}_j^{\mathrm{T}}\textbf{v}/\tau),
\end{equation}
where $\textbf{v}_i$ or $\textbf{v}_j$ is the representation of $i$-th or $j$-th sample, which serves as a substitute for the parametric class prototype (i.e., weights of a classifier). Addtionally, $\tau$ is the temperature parameter borrowed from the knowledege distillation~\cite{hinton2015distilling}. 


Local Aggregation (LA)~\cite{zhuang2019local} is another method that trains a CNN encoder to embed raw images into a lower dimension space -- embedding space.
When a metric of local aggregation is maximized, similar data instances move together in the embedding space while dissimilar instances move apart.

\begin{figure*}
    \centering
    \includegraphics[width=0.85\linewidth]{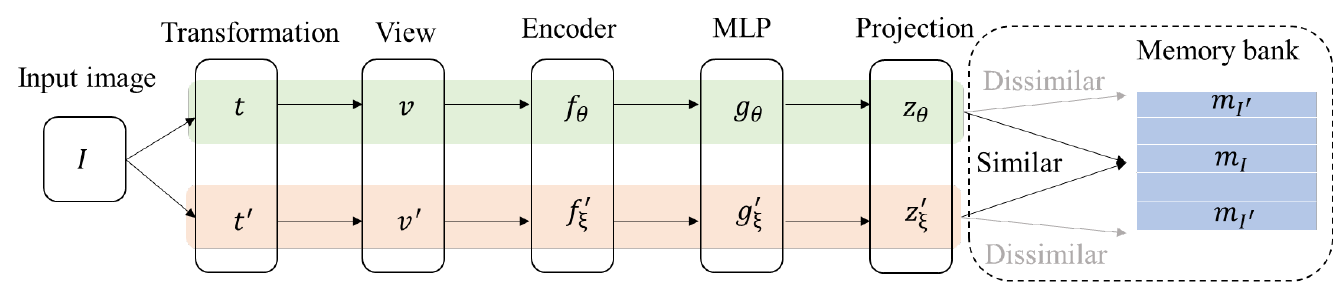}
    \caption{Summary of all two-stream models, including contrastive learning and memory-bank-based methods.}
    \label{fig:two-stream}
\end{figure*}

Based on NPID, Pretext Invariant Representation Learning (PIRL, pronounced as ``pearl'')~~\cite{misra2020self} is proposed to argue that semantic representations are invariant under pretext transformation tasks. Suppose the original view and transformed view of images are denoted as $I$ and $I^{t}$, respectively. These sample views are fed into a CNN encoder, and the total empirical loss on the training dataset $\mathcal{D}$ can be defined as:
\begin{equation}
    \mathcal{L}_{total}(\theta;\mathcal{D})=\mathbbm{E}_{t\sim\mathcal{T}}\left[\frac{1}{|\mathcal{D}|}\sum\nolimits_{\boldsymbol{I}\in\mathcal{D}}\mathcal{L}(\boldsymbol{V_I},\boldsymbol{V}_{\boldsymbol{I}^{t}})\right],
\end{equation}
where $\mathcal{T}$ denotes the different transformations of images. 
The loss encourages the representation of image $\boldsymbol{I}$ to be similar to that of $\boldsymbol{I}^t$, and the representation of $\boldsymbol{I}^t$ to be dissimilar to that of different images $\boldsymbol{I}'$, as shown in the dotted box of Fig.~\ref{fig:two-stream}. 
Therefore, more negative sample pairs contribute to  improving the scalability of the gradient and lead to the final learned encoder with stronger representation ability.
That is the reason why the memory bank is introduced to store more previous representations for subsequent comparison. 


\subsection{Learning by Sharing}

SSL prefers using two encoder networks for the different data augmentation, and then pretrains the parameters by maximizing the distance between negative pairs or minimizing the distance between positive pairs. Fig.~\ref{fig:two-stream} shows the two-stream models for all contrastive learning frameworks. The transformation $t$ on the orginal input image $\boldsymbol{I}$ generates the view $v$, similarly, its counterpart ${t}'$ generates ${v}'$. In general, two different or same encoders $f_\theta$ and $f'_\xi$ are used to extract contrastive representations. The subsequent MLP heads $g_\theta$ and $g'_\xi$ are used to learn more combinations that are beneficial to the contrastive loss. It is noticed that MLP and memory bank could be removed or preserved under different settings. In terms of the shared encoder, SSL can be divided into two categories: 1) Soft Sharing that two encoders share with similar but different parameters ($f_\theta \neq f'_\xi$); 
2) Hard Sharing that two encoders maintain the same architectures and parameters ($f_\theta = f'_\xi$).

\begin{figure*}[t]
    \centering
    \includegraphics[width=0.75\linewidth]{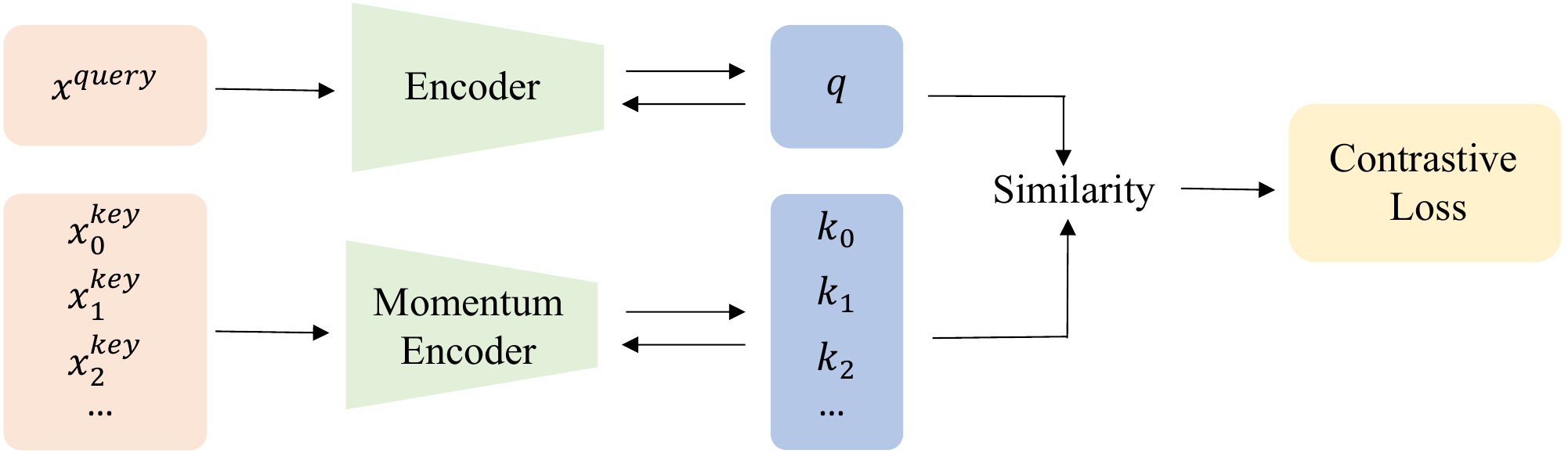}
    \caption{The general pipeline of MoCo~\cite{he2020momentum}, which is also a two-stream framework with different parameters.}
    \label{fig:moco}
\end{figure*}

\paragraph{Soft Sharing.}
Facebook AI Research (FAIR) presents Momentum Contrast (MoCo)~\cite{he2020momentum} 
by using momentum to control the slight difference between two encoders. As shown in Fig. \ref{fig:moco}, one of the encoders is served as a dictionary look-up task that generates a queue of encoded data samples $\{k_0, k_1, \cdots\}$.
Another encoder generates encoded query $\{q_0, q_1, \cdots\}$ with the training batch updated. 
The similarity is measured by the dot product of the new coming encoded query $q$ and the encoded keys stored in the dictionary queue. Suppose there are $K$ keys stored in the queue before the new key comes. The $K$ keys are treated as negative samples to the query of the new key. To combine the contrastive loss on both negative and positive samples, InfoNCE Loss ~\cite{oord2018representation} is used for the pretraining in MoCo.
The key design in MoCo for soft parameter sharing is called momentum update. He et al.~\cite{he2020momentum} suggest that the direct parameter change of key encoder (i.e., momentum encoder) to query encoder loses the necessary consistency and yields poor results. The momentum encoder parameter $\theta_k$ is updated as:
\begin{equation}\label{momentum-update}
    \theta_k = m\theta_k + (1-m)\theta_q,
\end{equation}
where the query encoder parameter $\theta_q$ is learned directly from the gradients of new coming instance, and $m\in[0, 1)$ is a hyper-parameter that controls the consistency ($\theta_k$ is more consistent if $m$ is closer to $1$).

Inspired by the design of SimCLR~\cite{chen2020improved}, in MoCo v2~\cite{chen2020improved}, the FAIR team introduces an MLP projection head after encoders and utilizes more data augmentation techniques to improve the performance. The further improvements are from that: 1) embedded linear classifier bridges the gap between unsupervised and supervised pretraining representations; 2) more contrastive samples are feasible from both the larger training batch and stronger data augmentation.


DeepMind proposed Bootstrap Your Own Latent (BYOL)~\cite{grill2020bootstrap} that contains representation, projection, and discrimination stages to achieve a new SOTA without using negative samples. They understand the discrimination between different views of raw images as necessary prevention from collapse during the pretraining. However, they argue that many negative samples
are not indispensable to prevent this collapse. As shown in the left part of Fig. \ref{fig:two-stream}, there are two streams in BYOL with different parameters. The online network (top green) updates parameters by comparing the prediction generated itself and the regression target provided by the target network. Then the parameters of the target model (bottom red) are updated the same as Eq.~(\ref{momentum-update}), \textit{i.e.}, $\xi\gets\tau\xi+(1-\tau)\theta$, where $\tau$ is the target decay rate to control the degree of parameter changing in the target network. Therefore, the target network can also be understood as a momentum encoder. Here,
$\xi$ in the target model is the parameter $\theta_k$ in momentum encoder, and $\theta$ in the online network denotes the parameter $\theta_q$ in the query encoder.

\paragraph{Hard Sharing.} 
SimCLR~\cite{chen2020simple} is proposed by Brain Team in Google Research which utilizes the hard parameter-sharing architecture. This simple framework can also be concluded in Fig. \ref{fig:two-stream}, in which
we can see that representations of different views of the same image are learned in 
the network $f(\cdot)$. This base encoder shares the parameters with each other.
Thus, memory bank and momentum setting to learn key and query encoders are not necessary, which contributes to a simpler backbone architecture and easier learning strategy.
The loss function to maximize the similarity between different views of the same image (positive pairs) is defined as
\begin{equation}    
\label{nce-loss}
    \ell_{i,j}=-\log exp(sim(z_i,z_j)/\tau) \\ / \sum\nolimits_{k=1}^{2N}\mathbbm{1}_{[k\neq i]}exp(sim(z_i,z_k)/\tau),
\end{equation}
where $(i,j)$ is a pair of positive samples, $\tau$ is an introduced hyper-parameter called temperature parameter~\cite{wu2018unsupervised}, and $\mathbbm{1}_{[k\neq i]}\in\{0,1\}$ is an indicator function to control the denominator containing only negative pairs. 

To avoid the dependence on a large number of explicit pairwise feature comparisons, Swapping Assignments between multiple Views of the same image (SwAV)~\cite{caron2020unsupervised} is proposed as an online algorithm by Inria and FAIR. SwAV introduces clustering to substitute the previous comparison between pairs, which gains more memory with the help of non-queue architecture. In this method, the clustering prototype joins the computation of the defined loss function. This prototype is encoded as the concatenation of vectors learned through the backpropagation in CNNs. Thus, there is no need for SwAV to compare the encoded representations between different views.


Based on the existing SwAV, a novel model called SElf-supERvised (SEER)~\cite{goyal2021self} aims to learn a pretrained encoder from any random image and unbounded dataset in the wild. The base network is RegNetY architectures~\cite{radosavovic2020designing} trained with the SwAV SSL method~\cite{caron2020unsupervised}. This method proves that the SSL is not specific to a curated dataset such as ImageNet, and the scalability of recent RegNet releases the limitation of traditional backbones such as ResNet. In addition, this method encourages the research community to explore more backbones suitable for universal SSL.

Attracting the attention in the recent SSL, 
FAIR conducts empirical experiments on the SSL by utilizing the structure of Simple Siamese (SimSiam) networks. This method~\cite{chen2021exploring} can avoid the design of negative sample pairs, large batches (or memory banks), and momentum encoders in traditional contrastive learning. The two encoders in Fig. \ref{fig:two-stream} with identical parameters that process two different views $t$ and $t^{\prime}$ of image $x$ are substituted by the only siamese network. MLP predictor $g$ is used for one of the view representations, and then the stop-gradient operation is applied to another view representation.
\begin{figure*}[t]
    \centering
    \includegraphics[width=0.7\linewidth]{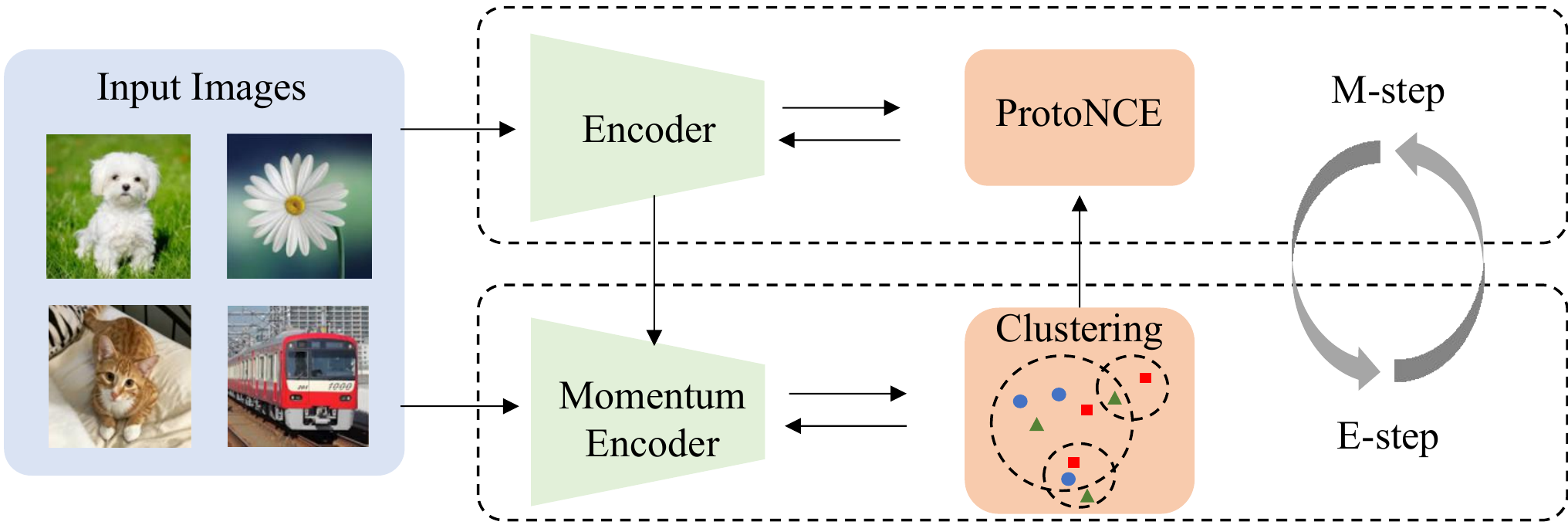}
    \caption{The key pipeline for the DeepCluster model~\cite{caron2018deep}.}
    \label{fig:clustering}
\end{figure*}

\subsection{Learning by Clustering}
DeepCluster~\cite{caron2018deep} is the first model that adopts the clustering algorithm for large-scale dataset learning. This method groups the representations into different clusters and labels these clusters as supervised signals to pretrain the parameters of the backbone network. It demonstrates SOTA performance on a wide range of standard transferred tasks used in unsupervised learning.

When it comes to the connection between contrastive learning and clustering, SwAV~\cite{caron2020unsupervised} has utilized prototypes that serve as a clustering center to help classify the sample pairs during pretraining, while Prototypical Contrastive Learning (PCL)~\cite{DBLP:conf/iclr/0001ZXH21} first targets bridging contrastive learning with clustering. Compared to instance discrimination as pretext tasks learning low-level representations, clustering can help to encode more semantic information. Then more semantic-based downstream tasks will benefit from it. As shown in Fig. \ref{fig:clustering}, prototypical contrastive learning uses prototypes to substitute one of the views of generated samples in NCE loss (Eq. (\ref{nce-loss})), which is the proposed ProtoNCE loss in PCL. In addition, PCL is also a method based on soft parameter sharing, in which the momentum encoder is updated as Eq.(\ref{momentum-update}).

\begin{table*}[t]
	\tiny
	\centering
	\caption{Summary of the PFMs in CV.}
	\label{tab:pretraining model for image}
	\resizebox{\textwidth}{!}{
	\begin{threeparttable}
    \begin{tabular}{llllllll}
		\hline 
		\textbf{Year} & \textbf{Conference} & \textbf{Method} & \textbf{Pretext Task} & \textbf{Architecture} & \textbf{Downstream Task}\tnote{1} & \textbf{Code} \\
		\hline 
		2014 & NeurIPS & Exemplar-CNN~\cite{dosovitskiy2014discriminative,DFB16} & discrimination & CNN & cla, rec & \href{https://lmb.informatik.uni-freiburg.de/resources/binaries/nips2014\_ExemplarCNN.zip}{https://lmb.informatik.uni-freiburg.de/...} \\
		\hline
		2015 & ICCV & Context~\cite{doersch2015unsupervised} & context prediction & CNN & cla, det, clu & \href{https://github.com/cdoersch/deepcontext}{https://github.com/.../deepcontext} \\
		\hline
		2016 & CVPR & Inpainting~\cite{pathak2016context} & inpainting & GAN, CNN &  cla, det, seg, inp & \href{https://github.com/pathak22/context-encoder}{https://github.com/.../context-encoder} \\
		\hline
		2016 & ECCV & Colorization~\cite{zhang2016colorful} & colorization & CNN & cla, det, seg & \href{https://github.com/richzhang/colorization}{https://github.com/.../colorization}  \\
		\hline
		2016 & ECCV & Jigsaw~\cite{noroozi2016unsupervised} & Jigsaw puzzles & CNN & cla, det, seg, ret & \href{https://github.com/MehdiNoroozi/JigsawPuzzleSolver}{https://github.com/.../JigsawPuzzleSolver}  \\
		\hline
		2017 & CVPR & Split-Brain~\cite{zhang2017split} & channel prediction & CNN & cla, det, seg & \href{https://richzhang.github.io/splitbrainauto}{https://richzhang.github.io/splitbrainauto} \\
		\hline
		2017 & ICCV & Counting~\cite{noroozi2017representation} & counting & CNN & cla, det, seg, ret & \href{https://github.com/clvrai/Representation-Learning-by-Learning-to-Count}{https://github.com/clvrai/...} \\
		\hline
		2017 & ICML & NAT~\cite{bojanowski2017unsupervised} & noise & CNN & cla, det & - \\
		\hline
		2017 & ICLR & BiGAN~\cite{donahue2016adversarial} & generation & GAN, CNN & cla, det, seg & \href{https://github.com/jeffdonahue/bigan}{https://github.com/.../bigan} \\
	    \hline
	    2018 & WACV & CDJP~\cite{kim2018learning} & Jigsaw puzzles &  CNN & cla, det, seg & - \\
		\hline
		2018 & ICLR & RotNet~\cite{zhang2016colorful} & rotation &  NIN, CNN & cla, det, seg & \href{https://github.com/gidariss/FeatureLearningRotNet}{https://github.com/gidariss/...} \\
		\hline
		2018 & arXiv & CPC~\cite{oord2018representation} & patch overlapping & CNN, GRU & cla & - \\
		\hline 
		2018 & CVPR & NPID~\cite{wu2018unsupervised} & instance discrimination & CNN & cla & \href{https://github.com/zhirongw/lemniscate.pytorch}{https://github.com/.../lemniscate.pytorch} \\
		\hline 
		2018 & ECCV & DeepCluster~\cite{caron2018deep} & clustering & CNN & cla, det, seg & \href{https://github.com/facebookresearch/deepcluster}{https://github.com/.../deepcluster} \\
		\hline
		2019 & ICCV & LA~\cite{zhuang2019local} & local aggregation & CNN & rec, det & \href{https://github.com/neuroailab/LocalAggregation}{https://github.com/.../LocalAggregation} \\
		\hline
		2019 & NeurIPS & BigBiGAN~\cite{DBLP:conf/nips/DonahueS19} & generation & GAN, CNN & gen, cla & \href{https://tfhub.dev/s?publisher=deepmind\&q=bigbigan}{https://tfhub.dev/...bigbigan}  \\
		\hline
		2019 & CVPR & AET~\cite{zhang2019aet} & transformation & CNN & cla & \href{https://github.com/maple-research-lab/AET}{https://github.com/.../AET} \\
		\hline
		2019 & NeurIPS & AMDIM~\cite{bachman2019learning} & discrimination &  CNN & cla & \href{https://github.com/Philip-Bachman/amdim-public}{https://github.com/.../amdim-public} \\
		\hline
		2020 & CVPR & ClusterFit~\cite{yan2020clusterfit} & clustering & CNN & cla, seg & - \\
		\hline 
		2020 & ICML & CPC v2~\cite{henaff2020data} & patch overlapping & CNN & cla, det & - \\
		\hline 
		2020 & CVPR & PIRL~\cite{misra2020self} & Jigsaw puzzles & CNN & cla, rec, dec & \href{https://github.com/facebookresearch/vissl/tree/master/projects/PIRL}{https://github.com/.../PIRL} \\
		\hline
		2020 & CVPR & MoCo~\cite{he2020momentum} & discrimination & CNN & cla, rec, dec, pos, seg & \href{https://github.com/facebookresearch/moco}{https://github.com/.../moco} \\
		\hline
		2021 & ICLR & PCL~\cite{DBLP:conf/iclr/0001ZXH21} & clustering & CNN & cla, det & \href{https://github.com/salesforce/PCL}{https://github.com/.../PCL} \\
		\hline
		2020 & arXiv & MoCo v2~\cite{chen2020improved} & discrimination & CNN & cla, dec & \href{https://github.com/facebookresearch/moco}{https://github.com/.../moco} \\
		\hline
		2020 & ICLR & SeLa~\cite{asano2019self} & self-labelling & CNN & cla, det, seg & \href{https://github.com/yukimasano/self-label}{https://github.com/.../self-label} \\
		\hline
		2020 & ICML & SimCLR~\cite{chen2020simple} & discrimination & CNN & cla & \href{https://github.com/google-research/simclr}{https://github.com/.../simclr} \\
		\hline
		2020 & NeurIPS & SimCLR v2~\cite{chen2020big} & self-distillation~\cite{hinton2015distilling} & CNN & cla & \href{https://github.com/google-research/simclr}{https://github.com/.../simclr} \\
        \hline
		2020 & ECCV & CMC~\cite{tian2019contrastive} & view matching~\cite{cubuk2019randaugment} & CNN & cla, seg & \href{https://hobbitlong.github.io/CMC/}{https://hobbitlong.github.io/CMC} \\
		\hline
		2020 & NeurIPS & InfoMin~\cite{tian2020makes} & discrimination & CNN & cla, det, loc, seg & \href{https://hobbitlong.github.io/InfoMin/}{https://hobbitlong.github.io/InfoMin} \\
		\hline
		2020 & NeurIPS & SwAV~\cite{caron2020unsupervised} & cropping & CNN, Transformer & cla, det & \href{https://github.com/facebookresearch/swav}{https://github.com/.../swav} \\
		\hline
		2020 & NeurIPS & BYOL~\cite{grill2020bootstrap} & discrimination & CNN & cla, det, seg & \href{https://github.com/deepmind/deepmind-research/tree/master/byol}{https://github.com/.../byol} \\
		\hline
		2021 & arXiv & MoCo v3~\cite{chen2021empirical} & discrimination & CNN, Transformer & cla & - \\
		\hline
		2021 & ICLR & R\textsc{e}LIC~\cite{DBLP:conf/iclr/MitrovicMWBB21} & discrimination & CNN & cla, rel & - \\
		\hline
		2021 & ICLR & PCL v2~\cite{DBLP:conf/iclr/0001ZXH21} & clustering & CNN & cla, det & \href{https://github.com/salesforce/PCL}{https://github.com/.../PCL} \\
		\hline
		2021 & CVPR & SimSiam~\cite{chen2021exploring} & discrimination & CNN & cla, det, seg & \href{https://github.com/facebookresearch/simsiam}{https://github.com/.../simsiam} \\
		\hline
		2021 & ICML & DirectPred~\cite{DBLP:conf/icml/TianCG21} & discrimination & CNN & cla & \href{https://github.com/facebookresearch/luckmatters/tree/main/ssl}{https://github.com/.../ssl} \\
		\hline
		2021 & ICCV & DINO~\cite{caron2021emerging} & discrimination & CNN, Transformer & cla, seg & \href{https://github.com/facebookresearch/dino}{https://github.com/.../dino} \\
		\hline
		2021 & arXiv & MoBY~\cite{xie2021self} & discrimination & CNN, Transformer & cla, det, seg & \href{https://github.com/SwinTransformer/Transformer-SSL}{https://github.com/.../Transformer-SSL} \\
        \hline
        2021 & NeurIPS & MST~\cite{li2021mst} & token prediction & CNN, Transformer & cla, det, seg & - \\
        \hline
		2022 & ICLR & BE\textsc{i}T~\cite{bao2022beit} & token prediction & Transformer & cla, seg & \href{https://github.com/microsoft/unilm/tree/master/beit}{https://github.com/.../beit} \\
        \hline
        2022 & CVPR & MAE~\cite{he2022masked} & reconstruction & Transformer & cla, det, seg & \href{https://github.com/facebookresearch/mae}{https://github.com/facebookresearch/mae}\\ \hline
        
        2022 & CVPR & SimMIM~\cite{xie2022simmim} & reconstruction & Transformer & cla, det, seg & \href{https://github.com/microsoft/SimMIM}{https://github.com/microsoft/SimMIM}\\ \hline
        
        2022 & ArXiv & UM-MAE~\cite{li2022uniform} & reconstruction & Transformer & cla, det, seg & \href{https://github.com/implus/UM-MAE}{https://github.com/implus/UM-MAE}\\ \hline
        2022 & ArXiv & LoMaR~\cite{chen2022efficient} & reconstruction & Transformer & cla, det, seg & \href{https://github.com/junchen14/LoMaR}{https://github.com/junchen14/LoMaR}\\ \hline
        
        2022 & Arxiv & CAE~\cite{chen2022context} & reconstruction & Transformer & cla, det, seg & \href{https://github.com/lxtGH/CAE}{https://github.com/lxtGH/CAE}\\ \hline
        
        2023 & AAAI & PeCo~\cite{dong2021peco} & reconstruction & Transformer & cla, det, seg & -\\ \hline

        2023 & ArXiv & SAM~\cite{kirillov2023segment} & reconstruction & Transformer & det, gen, seg & 
        \href{https://github.com/facebookresearch/segment-anything}{https://github.com/facebookresearch/segment-anything}\\ \hline
        
	\end{tabular}
	\begin{tablenotes}
	    \tiny
	    \item[1] Downstream task types: classification (cla), recognition (rec), detection (det), localization (loc), segmentation (seg), clustering (clu), inpainting (inp), retrieval (ret), generation (gen), pose estimation (pos), reinforcement learning (rel).
	\end{tablenotes}
	\end{threeparttable}
	}
\end{table*}

\subsection{Summary}

This section extensively investigates recent progress in PFMs on images for representation learning, from the early perspective of designing pretext tasks for self-labeling to present contrastive loss-based SSL. The pipelines of the main methods are clearly illustrated. We hope this section
can prepare the incoming researchers to acquire a basic understanding of this novel area and some worthwhile research direction. 
We believe the powerful generalization ability of PFMs would extremely reduce training computation overhead by ``pretraining once and transferring forever''. Recent transformer-based PFMs have gradually outperformed traditional training from scratch on target datasets. 
This discovery will spur further exploration and research into this exciting field.

\section{PFMs for Graph Learning}  \label{Section 5}

With the development of deep learning in graphs, the parameters (i.e., graph embedding) of the model began to increase rapidly.
Therefore, large-scale labeled data is needed for training the models to avoid under-fitting or over-fitting.
However, the cost of constructing large-scale labeled datasets for graphs is too subjective, expensive, and time-consuming, especially in domains that require professional knowledge and timeliness.
While some semi-supervised approaches have temporarily mitigated the reliance of graph embedding models on label scale, they have not fundamentally resolved this problem. In recent times, researchers have turned their attention towards the application of PFMs in the field of graphs, inspired by their success in CV and NLP. However, for most graphs, obtaining large-scale pretraining data directly is challenging due to the unique nature of information such as nodes and edges. Therefore, recent studies have focused on utilizing the inherent information of a graph's attributes, topology, and community to enhance the effectiveness of the node's features. We have summarized the graph-related PFMs in \textbf{Table~\ref{tab:pretraining model for graph}}.

\subsection{Learning by Graph Information Completion}\label{sec:graph_gic}
\begin{figure*}[!t]
 \centering
    \subfigure[Graph Information Completion (GIC).]{
 \includegraphics[width=.4\linewidth]{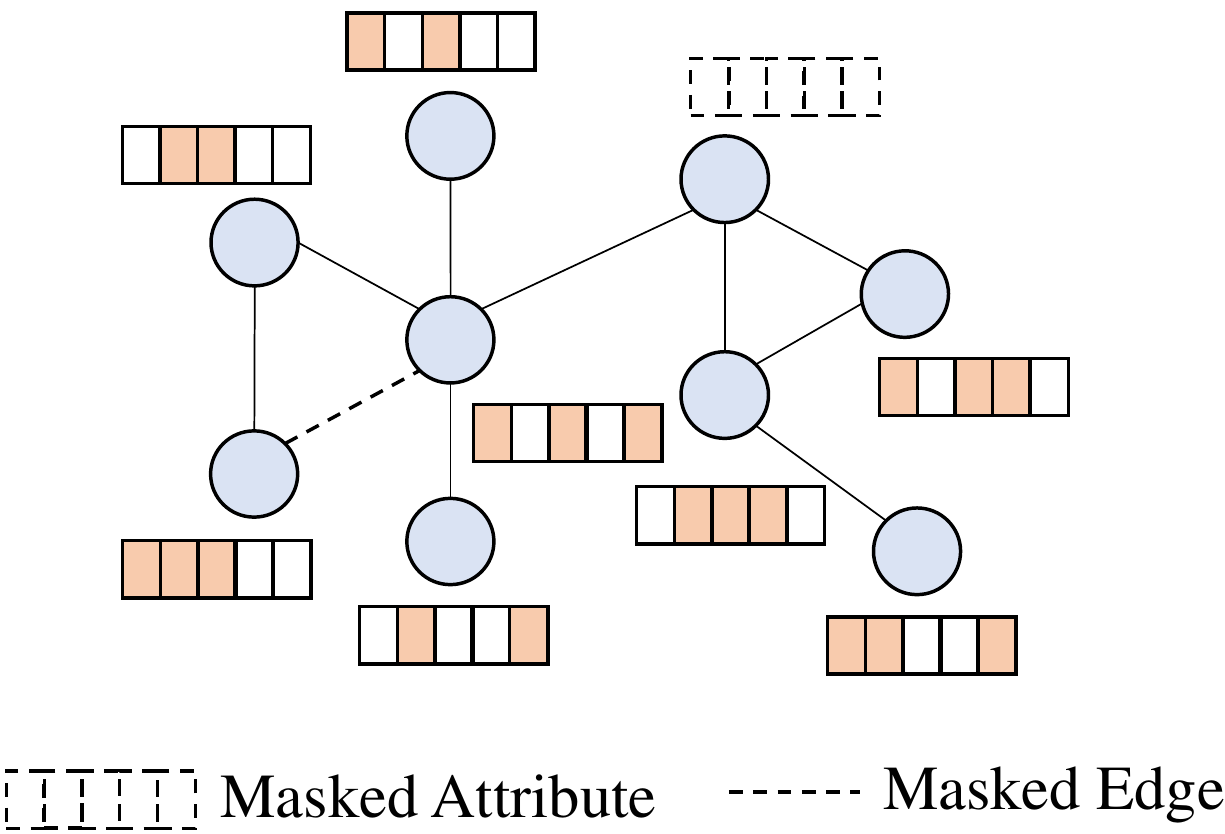}}
 \centering
    \subfigure[Graph Property Prediction (GPP).]{
 \includegraphics[width=.4\linewidth]{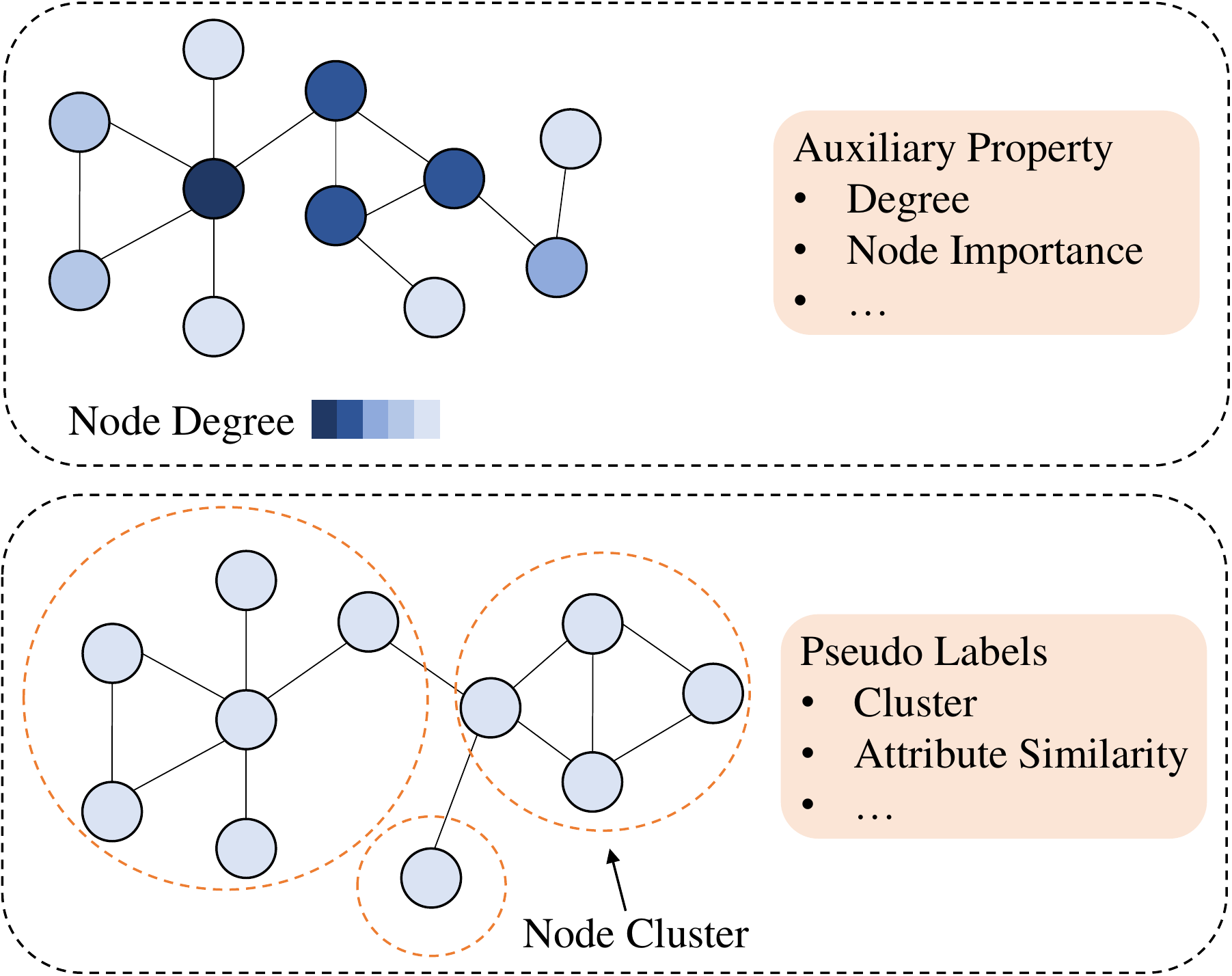}}
	\caption{Graph Information Completion (GIC) and Graph Property Prediction (GPP).}
	\label{fig:GIC_GPP}
\end{figure*}

The essential motivation of pretraining based on graph information completion (GIC) is to mask part of the information of the input graph data and recover the masked information based on the unmasked graph data, so as to pretrain the graph embedding, as shown in Fig. \ref{fig:GIC_GPP}. 
Similar ideas appeared earlier in the field of image and text processing.
For instance, in image processing, information such as image pixels and colors are recovered to pretrain the image encoder; in text processing, many methods implement pretraining of word embeddings and encoders by recovering part of the information in a sentence based on context words.
These methods inspire the design of graph completion tasks on graph PFMs.


Among them, You et al.~\cite{you2020icml} are inspired by image inpainting, and first propose to cover them by removing the features of the target nodes, and then recover/predict the features of the masked nodes.
In order to recover/predict the masked information, GraphCompetion~\cite{you2020icml} is achieved by providing GCNs with unmasked node features (limited to the 2-layer GCNs of the second-order neighbors of each target node).
The purpose of GraphCompetion's pretraining is to help the model better perform feature representation and teach the model to extract features from the context.
You et al.~\cite{you2020icml} propose the attribute mask task (namely, AttributeMask), which masks node attributes randomly, and then requires the self-supervising module to reconstruct the masked attributes.
Jin et al.~\cite{jin2020arxiv} think deeply about SSL on graph data, and propose the edge mask task (namely, EdgeMask), seeking to develop self-supervision in pairs based not only on a single node itself but on the connection between two nodes in the graph.
In particular, EdgeMask randomly masks some edges and then asks the model to reconstruct the masked edges.
In short, EdgeMask is expected to help GNN learn local connectivity information.
Hu et al.~\cite{hu2020iclr} propose a PFM that masks node and edge attributes and then predicts this masked information based on the adjacent structure.

\subsection{Learning by Graph Consistency Analysis }\label{sec:graph_gca}

Different from the aforementioned methods that focus on individual elements in the graph, graph consistency analysis (GCA) mainly explores the consistency of the distribution of two elements in the graph.
Specifically, the consistency of two elements with similar semantics should be significantly stronger than two elements with unrelated semantics, and this characteristic can be used to pretrain the graph model.
According to the judgment object of consistency, such methods can be roughly divided into the following three categories.

\begin{figure*}[!t]
 \centering
    \subfigure[Context Consistency.]{
 \includegraphics[width=.47\linewidth]{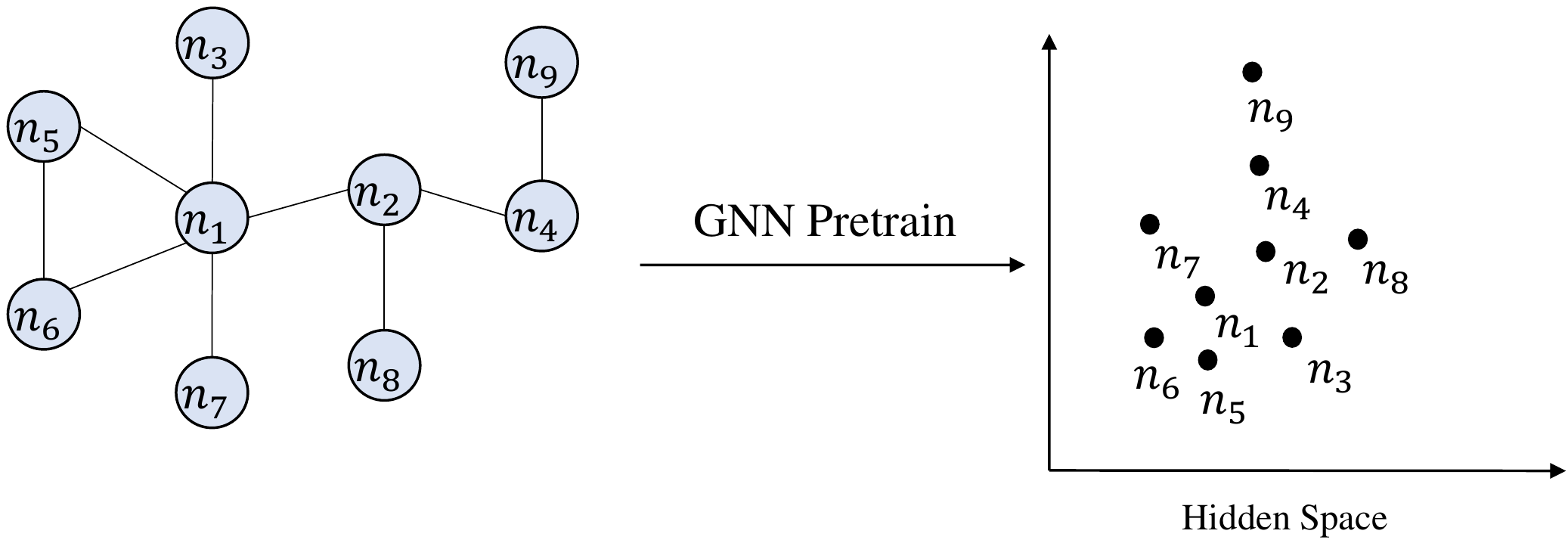}}
 \centering
    \subfigure[Self Consistency.]{
 \includegraphics[width=.47\linewidth]{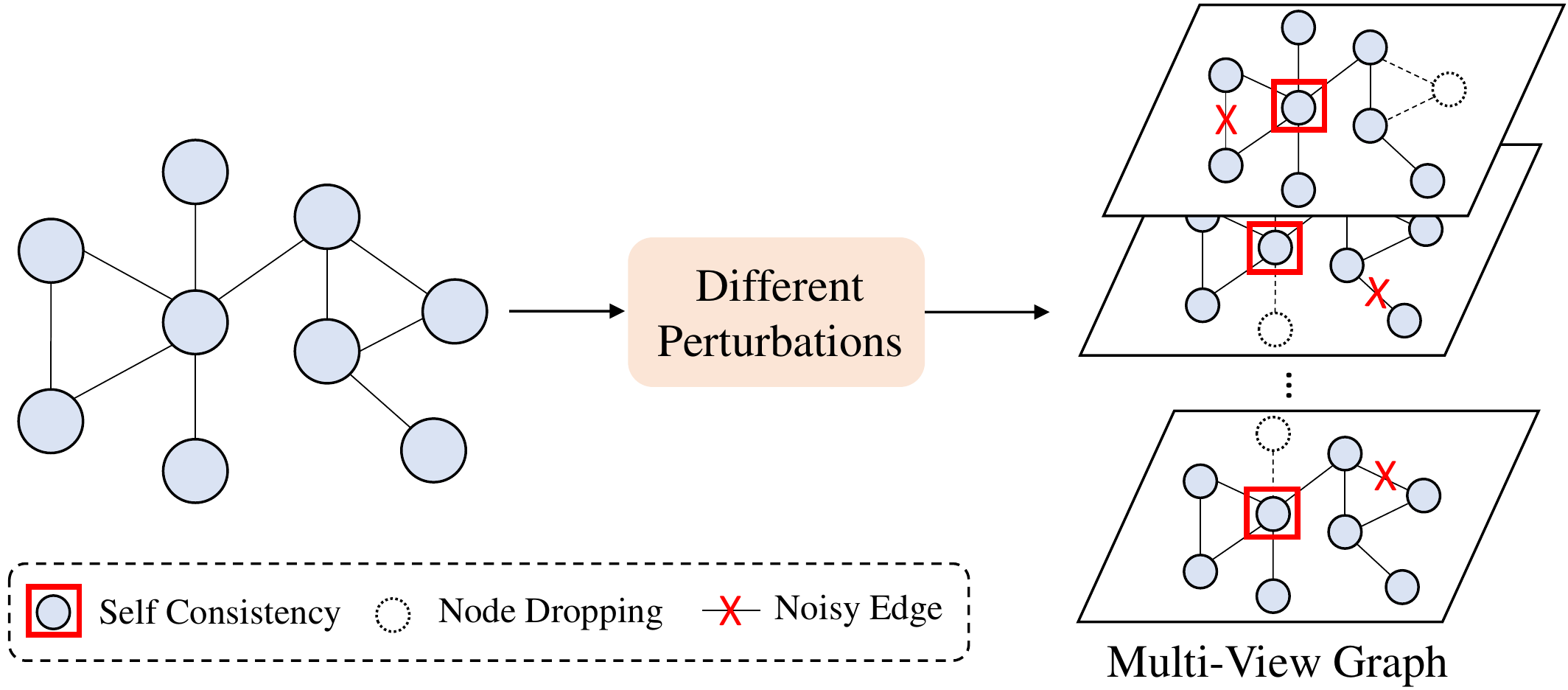}}
	\caption{Graph Consistency Analysis (GCA).}
	\label{fig:GCA}
\end{figure*}

\paragraph{Context Consistency}

Based on the early homogeneity assumption, a mass of graph models tends to project contextual nodes to similar positions in semantic space.
Such consistency of the context in the graph is also applied to the pretraining graph model, which attempts to adjust the node representation by capturing the distribution characteristics of the nodes in the context, as shown in Fig. \ref{fig:GCA} (a).

Random walk is an efficient method to acquire context. 
It can capture the distribution characteristics of different perspectives in the context by designing a variety of walk strategies.
The DeepWalk~\cite{deepwalk2014kdd} adopts a truncated random walk strategy to represent the node context as the form of a sequence of nodes.
By introducing the idea of NLP into the network embedding model, DeepWalk regards the node sequence as a ``sentence''
and models it based on the skip-gram model, providing an unsupervised and scalable training method for node representation.
Furthermore, on the basis of DeepWalk, node2vec~\cite{node2vec2016kdd} uses two different parameter-controlled random walk strategies to obtain deviated node sequences to fully capture the context information.

Different from randomly sampling nodes from the context, some recent methods directly consider the relationship between the node's k-order neighbor distribution (as positive examples) and non-adjacent nodes (as negative examples), and use this to train the graph model.
LINE~\cite{line2015www} respectively proposes first- and second-order proximity to describe the local similarity between pairs of nodes in the graph from different perspectives, and uses it to optimize node representation.
Meanwhile, LINE uses negative sampling and edge sampling techniques to optimize the second-order traversal and excessive training storage overhead.
VGAE~\cite{vgae2016nips} introduces a variational autoencoder to encode graph structure data, and model the node first-order neighbor through a GCN encoder and a simple inner product decoder.


\paragraph{Self Consistency}

In the field of NLP and CV, contrastive learning as an efficient self-supervised mechanism is widely used in the pretraining of models.
In fact, the internal comparison mechanism of such methods is based on the mutual information estimation of the original graph data and the augmented graph data to maintain the consistency of the data itself, as shown in Fig. \ref{fig:GCA} (b).
Inspired by contrastive learning, some studies have begun to generate augmented samples of original data samples in the graph model. 
Among them, two augmented samples from the same original sample are regarded as positive pairs, and two augmented samples from different original samples are regarded as negative pairs.

For node-level tasks, GCC~\cite{gcc2020kdd} devises the pretext task as subgraph instance discrimination in and across networks.
And GCC also enhances the ability of GNNs to learn the intrinsic and transferable structural representations by introducing contrastive learning.
Specifically, GCC samples subgraphs from the whole graph as augmentations via random walk with restart and artificially designs positional node embedding as node initial features.
As a novel graph representation learning model, GCA~\cite{gca2021www} incorporates various priors for topological and semantic aspects of the graph to achieve adaptive contrastive augmentation. 
Specifically, GCA devises an enhancement scheme based on node centrality measures to highlight important connection structures, while corrupting node features by adding noise to specific nodes to lead the pretraining model to recognize underlying semantic information.

For graph-level tasks, some studies have attempted to introduce more diverse contrastive learning strategies.
Among them, You et al.~\cite{graphcl2020nips} introduce four common graph augmentation tasks (i.e., node dropping, edge perturbation, attribute masking, and subgraph sampling) into the GL model based on underlying prior and propose a unified comparative learning framework: GraphCL.
Meanwhile, GraphCL discusses in depth the role of data augmentation in comparative learning and gives experimental demonstration that joint multiple augmentation strategies can improve model performance.

\paragraph{Cross Scale Consistency}

Unlike the above two methods that consider the consistency of elements in the same scale, contrasting elements in graph data of different scales can also be used to train graph models, e.g., node-subgraphs.
Most of such methods have the idea of maximizing mutual information~\cite{gic2021pakdd,sugar2021www}.
Specifically, the readout function is usually used to obtain the summary of the graph/subgraph, and the MI estimator can be calculated using the Jensen-Shannon divergence.

As a representative method, DGI~\cite{dgi2019iclr} relies on maximizing the MI between the patch representation and the summary of the corresponding high-level graphs, which are all derived using the established graph convolutional network architecture, to learn the node representation.
To generate negative samples on a single graph, DGI corrupts the original graph by randomly scrambling node features while keeping the structure unchanged.
Similarly, Hassani and Khasahmadi propose CMVRL~\cite{cmvrl2020icml}, which generates an additional structural view of a sample graph based on graph diffusion.
The sample graph and a regular view are sub-sampled together, and the node representation and graph representation are learned based on two shared MLPs, and then contrast learning is achieved through the consistency loss provided by the discriminator.

SUBG-CON~\cite{subcom2020icdm} samples a series of context subgraphs from the original graph and inputs them to the encoder to obtain the pooled central node and subgraph representation.
For the specified node, the context subgraph is expressed as a positive sample, and other randomly sampled subgraphs are expressed as a negative sample.
The contrast loss of the latent space will force the encoder to identify positive samples and negative samples in order to distinguish different nodes based on regional structure information.

\subsection{Learning by Graph Property Prediction}\label{sec:graph_gpp}

Considering the attribute and structural information of the graph as the target of information completion, pretraining based on graph property prediction (GPP) can also be used to build the graph model in different forms.
One of the most common methods is to generate self-supervised signals by exploring the auxiliary property in the graph data and to take the graph property prediction task as the pretraining task of the graph model.
According to the different settings of the pretext task, it can roughly classify two categories: property regression and property classification.

\paragraph{Property Regression (PR)}

In the graph model, different from the GIC mentioned above, property regression primarily focuses on mining the relationship between the broader numerical structure and property attributes within the graph.
Specifically, this branch of methods extracts richer self-supervised signals in graph data for pretraining graph models.

For example, similar but different from masking node attributes, the goal of NodeProperty~\cite{jin2020arxiv} is to predict each node's auxiliary property in the graph, e.g., degree, local node importance, and local clustering coefficient.
In other words, NodeProperty is used to encourage GNN to capture richer local structural information while optimizing the specific downstream tasks.
Specifically, NodeProperty regards the node degree as a representative local node property, i.e., self-supervised signal, and takes other node properties as future work.
Meanwhile, NodeProperty emphasizes that the intuition of devising self-supervised pretext tasks related to local node property is to ultimately guide the feature embedding of GNN (i.e., node representation) to save this information, which relies on the assumption that the node property information is relevant to the particular task.

\paragraph{Property Classification (PC)}

Different from the property regression task, the task of property classification is usually implemented by defining pseudo-labels based on a certain distribution in the graph data, which is a typical self-supervised method.
Among them, the structure density, similarity of node attributes, and difference between local and global distributions are the most commonly used. 
We will briefly introduce the application of such methods in GL pretraining.

Among these methods, clustering is the most common and effective source of pseudo-labels.
Among them, M3S~\cite{M3S2020aaai} designs a multi-stage training strategy, using the idea of graph clustering to iteratively train the graph encoder, achieving enlarged labeled data with virtual labels in the case of very small samples.
You et al.~\cite{you2020icml} further propose two pretraining strategies.
Among them, Node Clustering assigns $K$ (hyper-parameter) pseudo labels to nodes based on attribute clustering and pretrain node representation by node classification.
In addition, You et al. also present Graph Partitioning based on the topology density assumption.
In Graph Partitioning, the nodes of a graph are divided into approximately equal $K$ (hyper-parameter) subsets to minimize the number of edges connecting nodes among subsets, and then pseudo labels are provided for nodes.

In addition to clustering methods, some researchers generate pseudo labels based on other statistical characteristics of graph data.
For instance, in the molecular field, Rong et al.~\cite{grover2020nips} use the molecular bonds of subgraphs and related statistical information to guide GNN to learn Context-Sensitive Properties (CSP) and then apply them to prediction.
Rong et al.~\cite{grover2020nips} propose a Motif Prediction (MP) task, which can be expressed as a multi-label classification problem, in which each motif corresponds to a label.
Specifically, let's assume that $K$ motifs in molecular data are considered.
For a specific molecule (abstracted as graph $G$), they use RDKit to detect whether each motif appears in $G$, and then take it as the target of the motif prediction task.

\subsection{Learning by Masked Autoencoder}
The masked autoencoder (MAE) is first applied in MAGE~\cite{tan2022mgae}, the masked autoencoders for self-supervised learning on graphs. Following MAE \cite{he2022masked}, MGAE operates on a partial network structure (without masked edges) that is based on convolutions. Besides, the decoder of MGAE is designed to model the cross-correlation between the head and tail nodes of an anchor edge. Empirical results demonstrate that MGAE performs better than traditional graph autoencoders and graph SSL approaches. Furthermore, GMAE \cite{hou2022graphmae} extends this approach by using a transformer instead of convolutions and reconstructing the features of masked nodes rather than masked edges. In addition to empirical improvements, MaskGAE \cite{li2022maskgae} further provides theoretical justifications for the potential benefits of masked graph modeling. Designing algorithms to accommodate graphs of various complex properties is a promising direction. For instance, to tackle the heterogeneous graphs scenario, HGMAE \cite{tian2022heterogeneous} proposes meta-path masking and adaptive attribute masking with a dynamic mask to enable effective and stable learning on complex graph structure. Moreover, several training strategies are developed, including meta-path-based edge reconstruction to incorporate complex structural information, target attribute restoration to utilize various node attributes, and positional feature prediction to encode node positional information. Besides dealing with more complex graph structures, how to improve the learning efficiency of MAE on graph data remains an open question.

\subsection{Other Learning Strategies on Graph Data}
In addition to the above methods, there are lots of pretraining methods that use relatively novel or hybrid strategies. For example, CG$^3$~\cite{cg3aaai2021} generates an improved node representation by designing a semi-supervised consistency loss to maximize the consistency between different views of the same data or data from the same category.
Next, CG$^3$ uses the graph generation loss related to the input feature to extract the potential deterministic relationship between the data feature and the input graph topology as a supplementary supervision signal for SSL.

Based on the attention mechanism, Graph-Bert~\cite{graphbert2020arxiv} trains itself to reconstruct node attributes and topological structure with sampled linkless subgraphs within their local contexts.
GMI~\cite{gmi2020www} extends the traditional mutual information computing idea from the vector space to the graph domain and proposes to jointly maximize feature mutual information (between the node’s embedding and raw features of its neighbors) and edge mutual information (embedding of two adjacent nodes) for graph representation learning.
GPT-GNN~\cite{gptgnn2020kdd} proposes a self-supervised graph generation task to guide itself to capture the topological and semantic attributes of the graph.
GPT-GNN roughly divides the possibility of graph generation into attribute generation and edge generation to untangle the intrinsic dependence between node attributes and graph topology.


\subsection{Summary}
In the graph model, as traditional feature learning methods are often accompanied by information loss in the process of feature learning, and the information taken into consideration is relatively one-sided, the obtained graph representation is relatively rough and loses mass information.
People began to focus on the distribution of data and attributes in the graph data as self-supervised signals to pretrain the graph model so that it can capture more valuable information.
By transforming the distribution of nodes, attributes, and edges in the graph into different pretext tasks, and using GNNs for modeling, the graph model can fully fit the original distribution of the input graph.
In lots of unsupervised or semi-supervised scenarios, such pretrained graph models have been proven to benefit downstream tasks. Besides, federated training large graph models~\cite{wang2023federated} can be a promising solution for building pretrained foundation models.
Currently, with the in-depth study of contrastive learning strategies, some work has attempted to apply contrastive learning in different forms to the pretraining of graph models.
Through the consistency analysis of context, self, and cross-scale, this kind of method greatly improves the performance of the pretrained graph model on different graphs.

\begin{table*}[t]
	\tiny
	\centering
	\caption{Summary of PFMs in GL.}
	\label{tab:pretraining model for graph}
 \scalebox{1.1}{
	\begin{tabular}{llllll}  
		\hline 
		\textbf{Year} & \textbf{Conference} & \textbf{Method} & \textbf{Pretext Task} & \textbf{Encoder} & \textbf{Code} \\
		\hline
		2014 & KDD & DeepWalk~\cite{deepwalk2014kdd} & GC-C & Shallow NN & \href{https://github.com/phanein/deepwalk}{https://github.com/phanein/deepwalk} \\\hline
		2015 & WWW & LINE~\cite{line2015www} & GC-C & Shallow NN & \href{https://github.com/tangjianpku/LINE}{https://github.com/tangjianpku/LINE} \\\hline
		2016 & NeurIPS & VGAE~\cite{vgae2016nips} & GC-C & GCN & - \\\hline
		2016 & KDD & node2vec~\cite{node2vec2016kdd} & GC-C & Shallow NN & \href{https://github.com/aditya-grover/node2vec}{https://github.com/aditya-grover/node2vec} \\\hline
		2017 & NeurIPS & GraphSage~\cite{graphsage2017nips} & GC-C & Shallow NN & \href{https://github.com/williamleif/GraphSAGE}{https://github.com/williamleif/GraphSAGE} \\\hline
		2018 & ICLR & DGI ~\cite{dgi2019iclr} & GC-CS & GCN/SAGE & \href{https://github.com/PetarV-/DGI}{https://github.com/PetarV-/DGI} \\\hline
		
		2020 & ICML & GraphCompetion~\cite{you2020icml} & GIC & GCN & \href{https://github.com/Shen-Lab/SS-GCNs}{https://github.com/Shen-Lab/SS-GCNs} \\\hline
		2020 & ICLR & AttMasking~\cite{hu2020iclr} & GIC & GCN & \href{http://snap.stanford.edu/gnn-pretrain}{http://snap.stanford.edu/gnn-pretrain} \\\hline
		2020 & ICML & AttributeMask~\cite{you2020icml} & GIC & GCN & \href{https://github.com/Shen-Lab/SS-GCNs}{https://github.com/Shen-Lab/SS-GCNs} \\\hline
		2020 & arXiv & EdgeMask~\cite{jin2020arxiv} & GIC & GCN & \href{https://github.com/ChandlerBang/SelfTask-GNN}{https://github.com/ChandlerBang/SelfTask-GN} \\\hline
		2020 & arXiv & NodeProperty~\cite{jin2020arxiv} & GPP-PR & GCN & \href{https://github.com/ChandlerBang/SelfTask-GNN}{https://github.com/ChandlerBang/SelfTask-GN} \\\hline
		2020 & AAAI & M3S~\cite{M3S2020aaai} & GPP-PC & GCN & -  \\\hline
		2020 & ICML & Node Clustering~\cite{you2020icml} & GPP-PC & GCN & \href{https://github.com/Shen-Lab/SS-GCNs}{https://github.com/Shen-Lab/SS-GCNs}  \\\hline
		2020 & ICML & Graph Partitioning~\cite{you2020icml} & GPP-PC & GCN & \href{https://github.com/Shen-Lab/SS-GCNs}{https://github.com/Shen-Lab/SS-GCNs}  \\\hline
		2020 & NeurIPS & CSP~\cite{grover2020nips} & GPP-PC & GCN & -  \\\hline
		2020 & NeurIPS & MP~\cite{grover2020nips} & GPP-PC & GCN & -  \\\hline
		2020 & NeurIPS & SELAR~\cite{selar2020nips} & GC-C & GNN &  \href{https://github.com/mlvlab/SELAR}{https://github.com/mlvlab/SELAR} \\\hline
		2020 & KDD & GCC~\cite{gcc2020kdd} & GC-S & GIN & \href{https://github.com/THUDM/GCC}{https://github.com/THUDM/GCC} \\\hline
		2020 & NeurIPS & GraphCL ~\cite{graphcl2020nips} & GC-S & GCN & \href{https://github.com/CRIPAC-DIG/GCA}{https://github.com/CRIPAC-DIG/GCA} \\\hline
		2020 & ICML & CMVRL ~\cite{cmvrl2020icml} & GC-CS & GCN & - \\\hline
		2020 & ICDM & SUBG-CON ~\cite{subcom2020icdm} & GC-CS & GCN & \href{https://github.com/yzjiao/Subg-Con}{https://github.com/yzjiao/Subg-Con} \\\hline
		2020 & ICLR & InfoGraph ~\cite{infog2020iclr} & GC-CS & GCN & \href{https://github.com/fanyun-sun/InfoGraph}{https://github.com/fanyun-sun/InfoGraph} \\\hline
		2020 & AAAI & DMGI ~\cite{dmgiaaai2020} & GC-CS & GCN & \href{https://github.com/pcy1302/DMGI}{https://github.com/pcy1302/DMGI} \\\hline
		2020 & arXiv & Graph-Bert ~\cite{graphbert2020arxiv} & Hybrid & Transformer & \href{https://github.com/jwzhanggy/Graph-Bert}{https://github.com/jwzhanggy/Graph-Bert} \\\hline
		2020 & WWW & GMI ~\cite{gmi2020www} & Hybrid & GCN & - \\\hline
		2020 & KDD & Gpt-GNN ~\cite{gptgnn2020kdd} & Hybrid & GNN & \href{https://github.com/acbull/GPT-GNN}{https://github.com/acbull/GPT-GNN} \\\hline
		
		2021 & ICML & JOAO ~\cite{joao2021icml} & GC-S & GCN & \href{https://github.com/Shen-Lab/GraphCL\_Automated}{https://github.com/Shen-Lab/GraphCL\_Automated} \\\hline
		2021 & AAAI & CSSL ~\cite{cssl2021aaai} & GC-S & GCN & \href{https://github.com/UCSD-AI4H/GraphSSL}{https://github.com/UCSD-AI4H/GraphSSL} \\\hline
		2021 & PAKDD & GIC ~\cite{gic2021pakdd} & GC-CS & GCN & \href{https://github.com/cmavro/Graph-InfoClust-GIC}{https://github.com/cmavro/Graph-InfoClust-GIC} \\\hline
		2021 & WWW & SUGAR ~\cite{sugar2021www} & GC-CS & GCN & \href{https://github.com/RingBDStack/SUGAR}{https://github.com/RingBDStack/SUGAR} \\\hline
		2021 & ICML & GraphLoG ~\cite{graphlog2021arxiv} & GC-CS & GCN & \href{https://github.com/DeepGraphLearning/GraphLoG}{https://github.com/DeepGraphLearning/GraphLoG} \\\hline
		2021 & WWW & SLiCE ~\cite{slice2021www} & GC-CS & GCN & \href{https://github.com/pnnl/SLICE}{https://github.com/pnnl/SLICE} \\\hline
		2021 & WSDM & BiGI ~\cite{bigi2021wsdm} & GC-CS & GCN & \href{https://github.com/caojiangxia/BiGI}{https://github.com/caojiangxia/BiGI} \\\hline
		2021 & WWW & GCA ~\cite{gca2021www} & GC-S & GCN & \href{https://github.com/CRIPAC-DIG/GCA}{https://github.com/CRIPAC-DIG/GCA}\\\hline
		2021 & KDD & HeCo ~\cite{heco2021kdd} & GC-CS & GCN & \href{https://github.com/liun-online/HeCo}{https://github.com/liun-online/HeCo} \\\hline
		2021 & AAAI & CG$^3$ ~\cite{cg3aaai2021} & Hybrid & GCN & - \\\hline
		2021 & ICLR & SuperGAT~\cite{supergat2021iclr} & GC-C & GAT & \href{https://github.com/dongkwan-kim/SuperGAT}{https://github.com/dongkwan-kim/SuperGAT} \\\hline
		2021 & KDD & MoCL ~\cite{mocl2021kdd} & Hybrid & GNN & \href{https://github.com/illidanlab/MoCL-DK}{https://github.com/illidanlab/MoCL-DK} \\ \hline
		2022 & ArXiv & MGAE ~\cite{tan2022mgae} & Maksed Edge Reconstruction & GCN &-\\ \hline
	   2022 & KDD & GMAE ~\cite{hou2022graphmae} & Maksed Node Reconstruction & Transformer &\href{https://github.com/THUDM/GraphMAE}{https://github.com/THUDM/GraphMAE} \\ \hline
	   2022 & Arxiv & MaskGAE ~\cite{li2022maskgae} & Partial Maksed Node Reconstruction& Transformer &\href{https://github.com/EdisonLeeeee/MaskGAE}{https://github.com/EdisonLeeeee/MaskGAE} \\ \hline
	   2022 & Arxiv & HGMAE ~\cite{tian2022heterogeneous} &  Metapath Masking Reconstruction& Transformer & - \\ \hline
	\end{tabular}
 }
\end{table*}



\section{PFMs for Other Data Modality}\label{Section 6}

With the rapid development of the PFMs, except for text, image, and graph, the PFMs have also carried out a lot of research on speech, video, text-image, and cross-data. Besides, researchers have started investigating the unified PFMs that incorporate all three mentioned domains recently.
Therefore, in this section, we introduce some other advanced and unified PFMs.

\subsection{PFMs for Speech}
In the field of speech, wav2vec~\cite{wav2vec2019isca} obtains speech representation by capturing contextual information on large-scale unlabeled datasets, and fine-tuning on a few samples by noise comparison binary classification task, which greatly improves the performance of downstream tasks.
Furthermore, vq-wav2vec~\cite{vq-wav2vec2020iclr} and wav2vec 2.0~\cite{wav2vec22020nips} propose a discrete unsupervised pretraining method on the basis of wav2vec, discretizing the original continuous speech signal, so that the methods in the mature NLP community can be migrated and applied.
Meanwhile, lots of research have tried to design different mechanisms to use the representation obtained by speech pretraining as the initial input, and apply it to different tasks, e.g., automatic speech recognition~\cite{speechbert2020ieee,wav2vec22020nips}, phoneme recognition~\cite{speechxlnet2020isca}, and speech synthesis~\cite{chung2019ieee}.
In particular, the extensive application of spoken language understanding has promoted the research of joint pretraining of speech and text.
For example, SpeechBERT~\cite{speechbert2020ieee} applies MLM to speech and text pairs to perform representation learning on discrete information.
Unlike~\cite{forcomp2020isca}, which relies on a large amount of labeled data for joint pretraining, SPLAT~\cite{selfsup2021acl} uses unlabeled speech data to pretrain the speech embedding module, and proposes a label-level alignment method suitable for label-level downstream tasks based on sequence alignment.
MusicBERT~\cite{zeng2021musicbert} is a pretrained model designed for processing music data. It was developed by training on a vast symbolic music corpus consisting of over one million songs. To improve the pretraining process with symbolic music data, MusicBERT employs several mechanisms, such as OctupleMIDI encoding and a bar-level masking strategy. Huang et al.~\cite{huang2020pop} suggest incorporating a metrical structure in the input data, which allows Transformers to better recognize the hierarchical structure of music at the beat-bar-phrase level. AudioTransformer~\cite{verma2021audio} is a model that enhances the performance of Transformer architectures by implementing certain techniques, such as pooling, which were previously used in convolutional networks. Verma et al.~\cite{verma2021audio} demonstrate how they leverage multi-rate signal processing ideas based on wavelets to improve the Transformer embeddings and obtain better results.

\subsection{PFMs for Video}
Video is similar to the RGB features of image and sequence information of the text. Many meaningful explorations in self-supervised video representation learning can not only perform efficiently well in video datasets but also generalize to the learning in other areas. Odd-One-Out Networks (O3N)~\cite{fernando2017self} is a technique that targets to predict the odd video subsequence among real subsequences sampled from a video in a training dataset. The experiments are conducted by using different video-clip encoders for O3N to prove consistent improvements of this pretraining design. Similarly, Shuffle and Learn~\cite{misra2016shuffle}  aims to learn the correct temporal order from a sequence of frames in a video. However, Kim et al.~\cite{kim2019self} designed a new self-supervised task called Space-Time Cubic Puzzles to train 3D CNNs. This task requires a pretrained backbone to arrange permuted 3D spatiotemporal crops. The performance of downstream tasks proves that effective video representations have been learned while solving such puzzles.

Inspired by the contrastive learning in images, many pretraining models in the video also utilize the contrastive loss to learn video presentations for downstream tasks. Inter-Intra Contrastive (IIC) framework~\cite{tao2020self} can learn video representations by using positive and negative pairs generated from different videos. Specifically, different modalities in the same video are treated as positive pairs, and video clips from different videos as negative ones. Temporal Contrastive Pretraining (TCP)~\cite{lorre2020temporal} is another contrastive method based on CPC to learn video representations. Different from the existing GAN-based method that generates future frames for the video directly, TCP can predict the latent representation of future frames of the video, which is better for long-term predictions.
Sequence Contrastive Learning (SeCo)~\cite{yao2020seco} is a novel method considering both intra- and inter-frame instance discrimination in sequence order-based task.

\subsection{PFMs for Multimodal}

The multimodal PFM among text and image can be divided into two categories: single-stream model and cross-stream model. The single-stream model refers to integrating text information and visual information at the beginning of the model. The Cross-stream model refers to text information and visual information encoded by two independent coding modules, respectively. Then different modal information is fused by mutual attention mechanism.

\paragraph{Single-Stream Model}
VisualBERT~\cite{DBLP:journals/corr/abs-1908-03557} inputs text and images into the model simultaneously, which are aligned and fused using Transformer's self-attention mechanism. The input of the text is the same as BERT, and the input of the image is the image features extracted by Fasters-RCNN. VisualBERT also does pretraining and then fine-tuning the specific task. It adopts two pretraining tasks, namely MLM and sentence-image prediction, determining whether the input sentence describes the corresponding image.
The structure of Unicoder-VL~\cite{DBLP:conf/aaai/LiDFGJ20} is very similar to VisualBERT, except for the processing of the image. Unicoder-VL extracts the image feature through Faster-RCNN and concatenates the feature with image position-encoding mapping to the same space. It enhances the image label prediction task, which predicts the categories of images.
The pretraining task of VL-BERT~\cite{DBLP:conf/iclr/SuZCLLWD20} is the same as Unicoder-VL.
The image input of VL-BERT includes four parts: the image region features extracted by Fasters-RCNN, the location of the region in the original image, location coding, fragment encoding, and [IMG] encoding.

\paragraph{Cross-Stream Model}

In ViLBERT~\cite{DBLP:conf/nips/LuBPL19}, the text and image modes are first encoded separately, and their outputs go through a standard attention module. This module is based on the Transformer structure. Still, in the self-attention mechanism, each module uses its query to calculate attention with the value and key of another module to integrate the information between different modules.
The model is pretrained on two tasks. The first task is the mask task, which is the same as BERT. On the image side, the goal of the task is that when the region image is masked, the classification distribution of the output of the model can be as consistent as possible with the output distribution of the model used to extract the region features (such as Faster-RCNN). The second task is the language image matching task. DALL-E is a series of deep learning models developed by OpenAI to generate images from natural language prompts. The first version of DALL-E uses a transformer-based architecture, similar to the one used in the GPT LMs, to process the textual prompts and generate image-like representations. The model is trained on a dataset of images and their associated textual descriptions based on GPT-3. DALL-E 2~\cite{ramesh2022hierarchical} is the improved version by employing contrastive language-image pretraining (CLIP) \cite{radford2021learning} for capturing semantic association between image-text pairs and GLIDE diffusion model \cite{nichol2021glide} for text-conditional image synthesis. Furthermore, GPT-4 is proposed by OpenAI recently. It is a large-scale multimodal model which adopts RLHF and demonstrates human-level performance on various professional and academic benchmarks.

Based on the multi-modal data containing more available information than previous single-modality data, thus the performance of these models gets enhanced by combining with the SSL on the benchmark dataset. Cross and Learn~\cite{sayed2018cross} is the first method that reveals crossmodal information as an alternative source of supervision and obtains powerful feature representations from combining crossmodal loss and diversity loss in both RGB and optical flow modalities. 
Different from the existing methods that learn feature representations from only a single task in cross-domain datasets, Ren and Lee et al.~\cite{ren2018cross} propose a novel deep multi-task network to learn more generalizable visual representations to overcome the domain difference and further utilize the cross-domain information in different tasks. In that paper, the cross-domain datasets are real and synthetic datasets generated by a GAN-based network, while the multiple tasks are the predictions of the surface normal, depth, and instance contour in RGB images. This model performs better than any previous single-task-based SSL methods by learning general-purpose visual representations from cross-domain multi-task feature learning. Tian et al.~\cite{tian2020contrastive} believe that a powerful representation is one that models cross-view factors from the perspective of humans view to understand the world. They propose Contrastive Multiview Coding (CMC) to learn a video representation by maximizing mutual information between different views of the same scene.

\subsection{PFM for Code Generation}
Code generation with LLMs involves using pretrained language models to automatically generate code based on natural language descriptions of a desired program. This approach has the potential to greatly improve the efficiency of software development by reducing the need for manual coding and allowing developers to focus on higher-level tasks.

The technique involves training large-scale language models on vast amounts of natural language text and then fine-tuning the models on specific programming tasks. By inputting natural language descriptions of code, the model can generate code snippets that are syntactically and semantically correct. Code generation with LLMs has been applied in various programming domains, including web development, NLP, and data analysis. The models used for code generation include GPT-4, T5, and Codex, among others. For example, Andrei et al.~\cite{zlotchevski2022exploring} have investigated and assessed the fine-tuning of transformer models for personalized code generation. Specifically, they have evaluated the effectiveness of various personalization techniques in the domain of generating unit tests for Java methods and learning to personalize for a specific software project. Shailja et al.~\cite{thakur2022benchmarking} assess the capacity of LLMs to generate Verilog that is useful. To achieve this, pretrained LLMs are fine-tuned on Verilog datasets collected from GitHub and Verilog textbooks. An evaluation framework is then constructed, consisting of test benches for functional analysis and a flow for testing the syntax of Verilog code generated in response to problems of varying degrees of difficulty. An open-source CodeGen LLM that has undergone fine-tuning has been shown to outperform the current leading commercial Codex LLM. The CodeGen~\cite{nijkamp2022codegen} is a group of LLMs that have up to 16.1B parameters and can handle both natural language and programming language data. Additionally, they have released the training library JAX FORMER as open-source. Their work demonstrates that the model can perform as well as the previous state-of-the-art zero-shot Python code generation on HumanEval, showcasing the practical applications of the trained model. Synchromesh, introduced in the study by Poesia et al.~\cite{poesia2022synchromesh}, adopts a novel approach called Target Similarity Tuning (TST) to retrieve a small set of examples from a training bank. Then, Synchromesh utilizes these examples to train a pretrained language model and generates programs by applying Constrained Semantic Decoding (CSD). CSD is a general framework that can restrict the output to valid programs in the target language. In this work, the authors show that the combined use of CSD and TST results in significant improvements in prediction accuracy, as well as preventing runtime errors.

However, there are still some limitations to code generation with LLMs, such as the models' tendency to generate overly verbose or inefficient code and their inability to handle complex programming tasks. Nevertheless, the technology has shown significant promise and has the potential to revolutionize the software development industry.

\subsection{SOTA Unified PFMs}

A big convergence of PFMs handling multiple modalities is emerging, such as backbone architecture, pretraining task, and model scaling up~\cite{wang2022image}. Therefore, many unified PFMs proposed by researchers arise. A unified PFM is a unified model pretrained on unimodal and multimodal
data with single or multiple transformers as the backbone, which has the ability to perform a large variety of downstream AI tasks, including unimodal tasks and multimodal tasks. There are currently three types of SOTA unified models based on model architectures. We defined them as the single-transformer model, multi-transformer model, and comb-transformer model. A single-transformer model refers to a PFM model which only has a large-scale transformer as its backbone, whereas a multi-transformer model refers to a PFM model having multiple transformers. A comb-transformer model is the PFM model with the combination of both single and multiple transformer structures. 

\paragraph{Single-transformer Model}


UNITER~\cite{chen2020uniter} is a large-scale PFM for joint image-text embedding, which consists of an Image Embedder, a Text Embedder, and a multi-layer Transformer. It first encodes visual features and bounding box features for image regions using Image Embedder and tokens and positions using Text Embedder. Then, a Transformer module is applied to learn generalizable contextualized embeddings for images and text through four pretraining tasks. Instead of applying random joint masking to both modalities, conditional masking on pretraining tasks is used. Six vision-language tasks are selected as the downstream tasks.

Uni-Perceiver~\cite{zhu2022uni} is a single siamese model with shared parameters having the ability to process different modalities regarding vision
and language tasks. Different task inputs and targets are encoded into unified token sequences with modality-specific tokenizers, which are then decoded by a modality-agnostic weight-sharing Transformer encoder into the shared representation space. Any perception task is modeled as finding the maximum likelihood target for each input through the similarity of their representations.
Uni-Perceiver is pretrained on unimodal and multimodal tasks. The evaluation results on various downstream tasks show that the performance is close to SOTA methods by conducting prompt tuning on 1\% of downstream task data.

Gato~\cite{reed2022generalist} builds a single large transformer sequence model that works as a multimodal, multi-task, multi-embodiment generalist policy. It can perform various tasks using a single neural network with the same set of weights. Gato is trained on 604 tasks, where different types of data, such as images, text, proprioception, joint torques, and other discrete and continuous observations and actions, are serialized into a flat sequence of tokens, batched, and processed by the transformer. During deployment, sampled tokens are assembled into different actions based on the context.

OFA~\cite{wang2022unifying} is a simple sequence-to-sequence learning framework with a unified instruction-based task representation that unifies various tasks. In the pretraining and finetuning stages, OFA requires no extra task-specific layers for downstream tasks to achieve Task-Agnostic. The Modality-Agnostic compute engine is a Transformer with the constraint that no learnable task- or modality-specific components are added to downstream tasks. OFA is pretrained on small-size image-text pairs to achieve crossmodal tasks while attaining highly competitive performances on unimodal tasks.

UNIFIED-IO~\cite{lu2022unified} is a sequence-to-sequence model using a unified architecture that performs large and diverse tasks. UNIFIED-IO is a transformer model where both the encoder and decoder are composed of stacked transformer layers. The unified architecture does not need specific task or modality branches, which is accomplished by homogenizing the input and output of each task into a sequence of discrete vocabulary tokens. It trains a single transformer-based architecture on over 90 diverse datasets in the vision and language fields. UNIFIED-IO is the first model to perform various tasks and produce strong results across 16 diverse benchmarks without finetuning. 

BEiT-3~\cite{wang2022image} is a general-purpose multimodal pretrained model on language, vision, and vision-language tasks. The big convergence of BEiT-3 can be seen from three aspects, including backbone architecture, pretraining task, and model scaling up. It introduces a shared Multiway Transformer as backbone network performing masked data modeling on both unimodal and multimodal data. To process different modalities, every Multiway Transformer block has a shared self-attention module, and a pool of feed-forward networks. 
It is a giant-size foundation model that contains 1.9B parameters. Experimental results show that BEIT-3 can outperform SOTA models on both vision and vision-language tasks.

\paragraph{Multi-transformer Model}

FLAVA~\cite{singh2022flava} is an alignment model that targets all modalities at once and aims at solving vision and language tasks, and vision-language tasks. It utilizes a common transformer model architecture to learn strong representations from unimodal and multimodal data. An image encoder transformer is used to capture unimodal image representations. A text encoder transformer is adopted to process unimodal text information. A multimodal encoder transformer takes both encoded unimodal images and text as inputs and integrates their representations for multimodal reasoning. During pretraining, masked image modeling (MIM) and MLM losses are applied to the image and text encoders, respectively. On the other hand, masked multimodal modeling (MMM) and image-text matching (ITM) loss are used over paired image-text data. For downstream tasks, classification heads are applied to the outputs from the image, text, and multimodal encoders, respectively, for visual recognition, language understanding, and multimodal reasoning tasks. FLAVA shows good performance on 35 tasks across different domains. A noticeable advantage is that smaller datasets it used compared with other models.  


\paragraph{Comb-transformer Model}

UNIMO~\cite{li2020unimo} can learn both single modality and multimodalities with one model to achieve robust and generalizable representations. It employs multi-layer self-attention Transformers to learn general textual and visual representations simultaneously and unifies them into the same semantic space via cross-modal contrastive learning (CMCL). The main idea behind CMCL is to keep paired image and text representations close to the representation space while keeping non-paired representations far away. All of them are encoded by the same unified-modal Transformer in pairs or individually, and the representations of images and texts are extracted to compute the contrastive loss.


\section{Other Advanced Topics on PFMs} \label{Section 7}
As the number of parameters of the pretraining model increases, the pretraining model requires more memory and computing resources. It increases the training cost of PFMs and limits their deployment on resource-constrained devices.
Therefore, to improve the efficiency of the pretraining model, PFM improves computational efficiency from the following two aspects: model efficiency and model compression.
The model efficiency and compression of the PFM refer to simplifying the redundancy of model parameters and structure. Under the condition that the task completion degree is not affected, the model with fewer parameters and a more concise structure is obtained.

\subsection{Model Efficiency}
Model efficiency devotes to exploring more efficient pretraining methods to pretrain large-scale PFMs with a lower-cost solution. More efficient learning algorithms require more effective training methods and more efficient model architecture. Traditional pretraining tasks may be inefficient. For example, the commonly used masked token prediction task requires the model to predict masked tokens based on context. However, the masked tokens in the samples are usually a subset of the input tokens, and the model can only learn from this part of the tokens, so the training efficiency is low. To solve this problem, ELECTRA~\cite{DBLP:conf/iclr/ClarkLLM20} proposes an RTD task that predicts whether each input marker is replaced by other tokens, which enables the ELECTRA to train against all input tokens. In addition to effective training methods, more efficient architecture can also improve the efficiency of PFMS. For most PFMS based on the Transformer algorithm, a more efficient model architecture can be obtained by reducing the complexity of the Transformer algorithm.

\subsection{Model Compression}
Model compression requires less computing resources and memory. 
It is a potential approach to reduce the model size and enhance computation efficiency. The model compression strategy can be divided into two ways: parameter compression and structure compression.

The methods of parameter compression include parameter pruning, parameter quantization, low-rank decomposition, and parameter sharing. Parameter pruning refers to designing evaluation criteria for model parameters to delete redundant parameters based on a sizeable PFM. For example, Compressing BERT~\cite{DBLP:conf/rep4nlp/GordonDA20} prunes BERT before training while maintaining the performance equivalent to that of the original model. Parameter quantization is the quantization of model parameters from 32-bit full-precision floating-point numbers to lower-order numbers. For example, Q8BERT~\cite{DBLP:conf/nips/ZafrirBIW19} uses 8-bit quantization to compress parameters fourfold with little impact on model performance. Low-rank decomposition is to reduce the dimension of a high-dimensional parameter vector into a sparse low-dimensional vector. Parameter sharing refers to the structured matrix or clustering methods to map model parameters and reduce the number of parameters. For example, the ALBERT~\cite{DBLP:conf/iclr/LanCGGSS20} uses decomposition-embedded parameterization and cross-layer parameter sharing to reduce the parameters in the model.

Structure compression refers to compact networks and knowledge distillation. A compact network means reducing the number of parameters and calculations by designing a new compact network structure. 
Knowledge distillation refers to the transfer of knowledge from the larger teacher model to the smaller student model through the use of a soft label, etc. DistilBERT~\cite{DBLP:journals/corr/abs-1910-01108}, for example, uses the knowledge distillation method to compress BERT, reducing the size of the BERT model by 40\% while retaining 97\% of its language comprehension.

\subsection{Security and Privacy} \label{Section 8}
The security risks, social bias, and data privacy in PFMs become an important research topic. Qiu et al.~\cite{qiu2020pre} recognize that deep neural networks can be attacked by adversarial samples, which mislead the model to produce false predictions. Due to the excellent portability of pretraining models, they have been widely used in NLP, CV, and GL. However, it has been found that the pretraining model is susceptible to the influence of adversarial samples. A tiny interference of the original input may mislead the pretraining model to produce specific false predictions. Meanwhile, it is possible to recover the data samples by querying the PFMs which can cause privacy leakage. 

\paragraph{Generation Adversarial Samples}
The adversarial sample originates from the image. The adversarial samples of the image are hard to recognize with an invisible change. For example, only one pixel of the image is modified. Human beings do not easily detect such disturbance, but the neural network can identify the modified image, which is the original purpose of the adversarial sample.
Some work has found that pretrained LMs are vulnerable in some scenarios. Jin et al.~\cite{jin2020bert} successfully attack the three target models of BERT, CNN, and RNN by generating natural adversarial samples, which indicates that the current language processing model still has a large room for improvement in terms of security. 
However, it is difficult to achieve due to the distinct discreteness of languages in NLP. In particular, the generation of adversarial samples in the text must take into account linguistic characteristics to ensure that the sample's syntax and fluency are not harmed while affecting the model's output.
For example,~\cite{nlp12020aaai} uses adversarial samples to attack the fine-tuning stage of the BERT model for text classification and entailment successfully.~\cite{nlp22020acl} combines the sememe-based word substitution method and search algorithm based on particle swarm optimization to generate adversarial samples.  

\paragraph{Model Defects}
Some unrelated human factors can also mislead the PFM to make wrong predictions. For example,~\cite{nlp32019acl} discovers that the performance of BERT is limited in the reasoning task due to utilizing false statistical information in the dataset, which dramatically affects the performance by destroying this property.~\cite{nlp42019emnlp} defines universal adversarial triggers. When triggers are connected to any input, it can induce the model to generate specific predictions.

\paragraph{Backdoor Attacks}
There are still many methods to manipulate the predicted results of the pretraining model employing a backdoor attack.~\cite{nlp72020acl} demonstrates that it is possible to construct a weight poisoning attack in which pretrained weights are injected. After the fine-tuning stage, the backdoor is exposed. Attackers manipulate model predictions easily by injecting arbitrary keywords.~\cite{nlp82020sp} shows that PFMs in NLP can be manipulated by modifying the model corpus. The ``meaning'' of new words or existing words can be controlled by changing their weight parameters. 

\paragraph{Defense Against Attacks}
The human-in-the-loop method~\cite{nlp52019arxiv, nlp62020acl} has been proposed and applied to generate more natural, efficient, and diversified adversarial samples. 
Some defense approaches have been proposed to defend against such attacks. \cite{nlp112021acl} designs an auxiliary anomaly detection classifier and uses a multi-task learning procedure to defend against adversarial samples. On the other hand, some defects in the PFM may be inherited by the custom models in transfer learning, 
such as the adversarial vulnerabilities and backdoors mentioned above. To mitigate this issue, ~\cite{zhang2022remos} proposes a relevant model slicing technique to reduce defect inheritance during transfer learning while retaining useful knowledge from the PFM.

\paragraph{Data Privacy in PFMs}
LLMs and other PFMs have been trained on private datasets~\cite{carlini2021extracting}. The researchers have discovered that by querying the massive LMs, it is feasible to recover specific training samples. An adversary may, for instance, obtain IRC discussions and personally identifiable information. Even worse, because large models have so many parameters, it is simple for PFM to memorize or learn private information, making larger models more prone to attack than smaller ones.
Many PFMs such as the LLMs have been trained on private datasets. The researchers have found that it is possible to recover individual training examples by querying the LLMs. For instance, an adversary can extract examples including personally identifiable information, and Internet Relay Chat (IRC) conversations. Even worse, because of the billion parameters of large models, it is easy for PFM to learn private information, making the larger model more vulnerable than smaller models. We must take privacy-preserving measures into account during all PFM processes, including data processing, model training, model inference, and system deployment, in order to reduce the risks of privacy leakage.

\section{Future Research Challenges and Open Problems}\label{Section 11}
The PFM can avoid training models from the scratch, which is a breakthrough from weak AI to general AI. At present, due to the characteristics of PFM such as large-scale parameters, a large amount of training data, and high computational complexity, there are still many technical challenges in PFMs.
We summarize the future research challenges of PFMs from four perspectives: data, foundation, model design, and upstream and downstream tasks. Meanwhile, we point out some open problems in the future research direction.


\subsection{Challenges on Data}
Most pretrained datasets are for single modes and single languages. It is very important for the development of PFMs to construct pretrained datasets for multimodal, multi-lingual, and graph data.
For the characteristics of these data, the existing technical challenges are as follows:

\paragraph{Data Deficiencies}
Unlike NLP and CV, except for the reusable nodes in a few molecular and protein networks, most of the nodes and edges in the graph data do not have a large amount of unlabeled data for pretraining.
Meanwhile, the pretraining research of the graph model is still in its initial state. Besides, data from the Internet of Things (IoT) will be enormous and contains rich physical world information. For example, inertial measurement unit sensor data can capture users' social activity information~\cite{wang2018socialite, han2019shad}. 
The theoretical basis, various definitions of the pretext task, and the augmented design of contrastive learning are all imperfect,
and new research urgently needs to be supplemented.

\paragraph{Multimodal PFM}
Some research work has been done on multimodal PFMs, such as text and image, text and audio, etc. These are mostly PFMs between two modalities. At present, 
the learning of multimodal PFMs requires new multimodal datasets, which demand the establishment of the data between different modes. Thus, the construction of multimodal datasets is also an urgent problem to be solved.

\paragraph{Multi-lingual PFM}
The multi-lingual PFM solves the resource shortage problem in multiple languages, and it aids in the achievement of new improvements in QA, text summarization, low-resource neural machine translation, and so on. However, the current PFM is still a mask LM. To improve the performance of the multi-LM, some suitable new tasks need to be added. In addition, multi-lingual vocabularies are much larger than single-language vocabularies, resulting in a sharp increase in model parameters to be learned.

\subsection{Challenges on Foundation}
For a PFM, a theoretical foundation is essential to model performance, whether it is a ``black box'' or ``white box'' method. 
The foundation studied mainly includes theoretical foundation, semantic understanding, and explicable exploration.

\paragraph{Lack of Theoretical Foundation}
SSL in CV learns the experience from the NLP. There is no profound theory to support all kinds of tentative experiments, and further exploration has no handbook. Although there are several theoretical analysis that tries to understand the collapse of pretraining or the generalization ability of the learning representation, the lack of theoretical foundation is still a huge cloud upon the head of SSL.

\paragraph{Semantic Understanding}
Does the pretrained LM learn the meaning of the language, or just rely on corpus learning? Many models perform well on various datasets with helpful information that can be extracted, where some approaches even exceed human levels. 
However, the performance is poor on domain datasets or relatively small datasets. The models cannot reach a better level of stability and match different downstream tasks.
This means that the model cannot serve the real purpose of human language use.

\subsection{Challenges on Model Design}
Most existing structures of PFMs are tried for text, image, and graph.
The primary method is to increase data, improve computation power, and design training procedures to achieve better results.
How to make a trade-off between data, computing resources, and predictive performance is worth studying.

\paragraph{Model Variety} There are many attempts at model design, such as generation-based models in the CV area.
However, GAN-based approaches are not popular for the following two reasons: 1) the discriminator has learned meaningful feature representations, but they are forgotten during training~\cite{chen2019self}; 2) the mode collapse causes the generator to output samples in singular mode to cheat the discriminator. As a result, although researchers attempt to apply GAN-based approaches on SSL for pretraining, the difficulties in the convergence of discriminator and divergence of generator hinder development and progress in this area.

\paragraph{Model Compression}
With the wide application of the Transformer and the pretraining model showing a general trend of growth, the computational complexity of the pretraining model has become the focus of attention. Due to the huge hardware requirements of model training and other reasons, the high threshold makes it difficult for researchers to train from scratch. BERT-base and GPT-3 contain about 108 million parameters and 175 billion parameters, respectively. It is not conducive to the development of relevant research work. There are some works for pretraining model compression, such as ALBERT having fewer parameters and better effect than BERT-base. The improvement models still require powerful computing equipment, making them difficult to apply universally. Reducing the high computing cost is one of the main challenges in future research. 

\paragraph{Model Robustness}
Although many researchers have designed different pretext tasks for the pretraining, the main problem remains on how to design robust pretext tasks and judge the performance before large-scale computations. In addition, how to compare these proposed methods fairly is also a big issue. As for NLP, deep neural networks are vulnerable to adversarial inputs because of their linear characteristics. Although pretraining models perform well on different NLP tasks, most are based on deep neural networks, which generally have poor robustness. Operations such as cutting and rotating do not change the nature of the image in CV. In contrast, operations such as adding, deleting, and substituting a word in the text are likely to affect the semantics of the text. Therefore, how to improve the robustness of the PFM in NLP is a technical challenge.

\paragraph{Model Anti-attack}
The PFMs are vulnerable to attack by adversarial examples, which can easily mislead the model to produce specific false predictions. It is difficult to process due to the unique discreteness of language in the NLP field. Thus, the current PFMs have huge room for improvement in model anti-attack.

\subsection{Challenges on Finetuning and Prompt}

The pretrained model in NLP, CV, and GL fields can achieve good performance in most upstream tasks, but not all good in downstream tasks for fine-tuning and prompt. 
How to achieve consistent results both on upstream and downstream tasks is still a challenge for the PFMs.

\paragraph{Saturation Phenomena}
Google Research~\cite{abnar2021exploring} observed the nonlinear relationship between the performance of upstream and downstream tasks. The higher training accuracy with more data on the upstream tasks does not always lead to better performance on the target downstream tasks. This observation challenges the most intuitive understanding of the pretraining process. Even in the most extreme case, the performance of upstream and downstream is at odds.

\paragraph{Pretext Task}
There are too many self-supervised tasks, also known as pretext tasks. The pretext task can be used for any downstream tasks, such as detection and classification. It is difficult to match the relationship between pretext tasks and downstream tasks.

\paragraph{Task-based Graph}
Much of the pretraining on the graph is based on the task graph. Different tasks construct different graphs, where nodes need to be reused. This makes it impossible to pretrain on the graph by introducing as much data as NLP and CV.

\subsection{Open Problems for Future PFMs}
First of all, a big convergence of text, image, graph, and multimodal pretraining is expected. Till the survey is written, no work has considered the graph in their unified PFMs. All of the SOTA unified models mainly focus on the language, vision, and language-vision tasks, while neglecting the importance of the graph in the data domain. Second, a unified backbone architecture for unified PFMs in future research will become more popular. It can be seen that a unified PFM model which only has a large-scale transformer as its backbone, i.e., a single-transformer model, is more focused by researchers than other types of unified PFMs. Third, a unified PFM is expected to achieve SOTA transfer performance for all different tasks in all data domains, including text, image, graph, and multimodalities. Most unified PFMs are only outstanding in a single data domain, whereas the performance in other domains is not competitive. BEiT-3~\cite{wang2022image} shows a great example in both vision and vision-language tasks towards this research direction. Besides, in terms of RL usage in PFMs, even though ChatGPT build the milestone in NLP, CV and GL do not have significant research published yet. More work in this direction is expected in the future.

\section{Conclusion}\label{Section 12}

Existing PFMs in text, image, and graph domains are principally summarized in this survey. 
Firstly, we introduce the basic components of NLP, CV, and GL.
Then, we provide a summary of existing models designed for pretraining in the  three domains and summarize the necessary information in terms of  model structures.
Furthermore, we study some other research for PFMs, including other advanced and unified PFMs, model efficiency and compression, and security and privacy. 
Finally, we present the main challenges and open problems in  PFM research.



\pagebreak
\appendix
\section{Basic Components} \label{Components}

\subsection{Basic Components on NLP}

\begin{table*}[htbp]
	\centering
	\caption{Commonly used notations on NLP and graph.}
	\label{tab:graph_notation}
	\resizebox{\textwidth}{!}{
	\begin{tabular}{c|c|c|c}  
		\hline 
		 \multicolumn{2}{c}{\textbf{NLP}} &   \multicolumn{2}{|c}{\textbf{Graph}}  \\
		\hline
		\textbf{Notations} & \textbf{Descriptions} & \textbf{Notations} & \textbf{Descriptions} \\
		\hline
	
		$N$ & The length of input text.  &$|\cdot|$ & The length of a set.\\	\hline
	   $w_{i}$ & The i-th word in input text.  &  	$\mathcal{G}$ & The set of graphs. \\	\hline
		$|V|$ &  The word corpus size. & 	$G$ & A graph. \\	\hline
		$H_{x}$& The representation of the input sequence.  &	$V$ & The set of nodes in the graph. \\	\hline
	$\overrightarrow{\boldsymbol{\theta}_{\text {f}}}$	& The parameters for forward modeling. &  	$v$ & A node. \\	\hline
	$\overleftarrow{\boldsymbol{\theta}_{\text {b}}}$	& The parameters for backward modeling. &  	$E$ & The set of edges in the graph. \\	\hline
	 $\theta$ & The shared parameter in all permutations. 	 &	$e_{ij}$ & An edge between $v_{i}$ and $v_{j}$. \\	\hline
	$Z_{N}$ & The set of all possible permutations of $T$.	 &  	$A$ & The adjacency matrix of a graph. \\	\hline
	 $z_{T=t}$ & The $t$-th element of $z$.	 & 	$\mathcal{T}$ & The set of node types in a graph. \\	\hline
	 $z_{T<t}$ & The $[1, 2, \ldots, t-1]$ elements of $z$.  & 	$X$ & The feature matrix of a graph. \\	\hline
	$z$	& A permutation of $T$.  & 	$\mathcal{Y}$ & The set of ground truth labels in a graph. \\	\hline
	$m$ & The dimension of the feature vector. &  	$D$ & The given graph data. \\	\hline
	 $b_{1}$, $b_{2}$ & The bias values of the hidden layer and the output layer.	 & 	$M_{GL}$ & The GL model. \\
		\hline
	\end{tabular}
	}
\end{table*}

\subsubsection{Language Model}

With the rapid development of deep learning, LMs are more and more applicable to the pretraining of NLP models. The LM can estimate the probability of rationality of a paragraph of the text. 
There are two main types of LMs: statistical LM and neural network LM.

\myparagraph{\textbf{Statistical LM}}
The statistical LM is a mathematical model to solve the context-related characteristics of natural language from the perspective of probability and statistics. The core of statistical LMs is to determine the probability of a sentence appearing in a text.
As the theoretical basis of the probabilistic LM, the N-gram model profoundly influences the subsequent LM. It plays a pivotal role in the field of the LM. 
The N-gram LM introduces the Markov hypothesis, which assumes that the probability of the occurrence of the current word only depends on the nearest $n-1$ words.
The maximum likelihood probability of a word $w_{i}$ can be calculated by
\begin{equation}
    p\left(w_{i} \mid w_{1}, w_{2}, \ldots, w_{N} \right)= \\ p\left(w_{i} \mid w_{i-n+1}, w_{i-n+2}, \ldots, w_{i-1} \right) = \\ \frac{C\left(w_{i-n+1}, w_{i-n+2}, \ldots, w_{i}\right)}{\sum_{N} C\left(w_{i-n+1}, w_{i-n+2}, \ldots, w_{i-1} \right)},
\end{equation}
where $T=[w_{1}, w_{2}, \ldots, w_{N}]$ is the text sequence and $C(w_{i-n+1}, w_{i-n+2}, \ldots, w_{i})$ is the co-occurrence frequency of $(w_{i-n+1}, w_{i-n+2}, \ldots, w_{i})$. 
The $p\left(w_{i} \mid w_{1}, w_{2}, \ldots, w_{N} \right)$ is calculated according to the chain rule
\begin{equation}
p\left(w_{1}, w_{2}, \ldots, w_{N}\right)=\prod_{i=1}^{N} p\left(w_{i} \mid w_{1}, w_{2}, \ldots, w_{i-1}\right).
\end{equation}
N-gram uses the probabilities of each word in the sequence to represent the co-occurrence probability of the whole text sequence.
When $N$ is large, it indicates a more vital constraint on the occurrence of the next word in the sequence and leads to more sparse frequency information. When $N$ is small, the statistical results have higher reliability and better generalization ability, but the constraint will be weaker.


\myparagraph{\textbf{Neural LM}}
The statistical LM adopts maximum likelihood estimation, which is intuitive and easy to understand. However, there are still problems such as a lack of long-term dependence, the rapid growth of parameter space and sparse data. Therefore, the neural network is introduced to map the LM to a continuous space. Neural LMs use distributed representations of words to model natural language sequences. Unlike class-based N-gram models, neurolinguistic models are able to recognize two similar words without losing the ability to encode each word as different from the other. It can be directly used for NLP tasks. It mainly introduces Forward Feedback Neural Networks (FFNN), Recurrent Neural Networks (RNN), and pretrained LMs.

\begin{figure*}[!t]
    \centering
    \includegraphics[width=\linewidth]{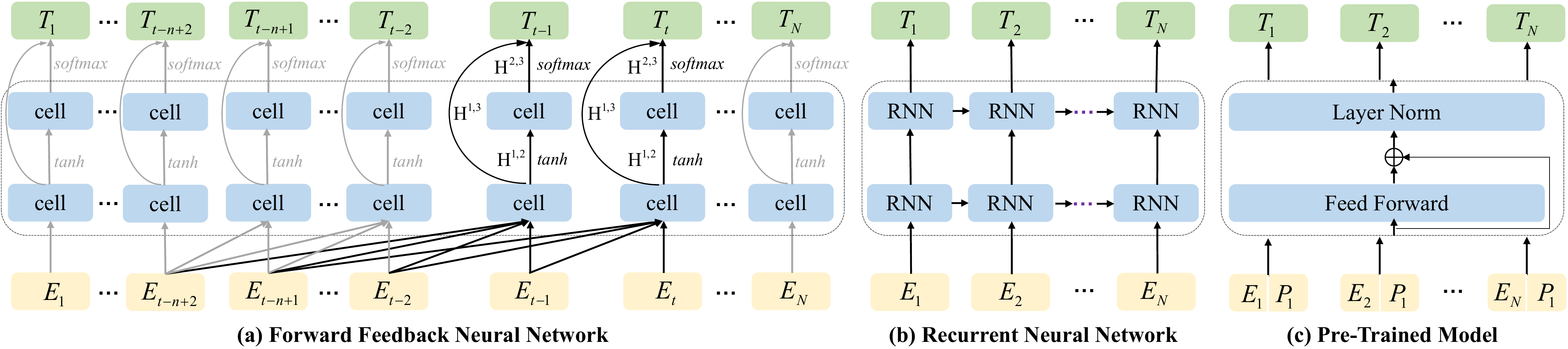}
    \caption{The model architectures of forward feedback neural network, recurrent neural network and Pretrained LMs. $H^{1,2}$, $H^{2,3}$ and $H^{1,3}$ are the weight matrices used to connect each layer.}
    \label{picture1}
\end{figure*}

As shown in Fig.~\ref{picture1} (a), FFNN according to the all former words of $x=[w_{1}, \ldots, w_{i-1}]$ calculates the probability of $w_{i}$. In order to predict the conditional probability of $w_{i}$, $x$ is sharing the projection matrix $M \in R^{|V| \times m}$ to a continuous feature vector space according to the projection index, $|V|$ is word library size, $m$ is the dimension of the feature vector. The output is represented as
\begin{equation}
y=b_{2}+H^{1,3}_{x}+H^{2,3}_{x} \tanh (b_{1}+H^{1,2}_{x}),
\end{equation}
where $H^{1,2}$, $H^{2,3}$ and $H^{1,3}$ are the weight matrices used to connect each layer, and $b_{1}$ and $b_{2}$ are the bias values of the hidden layer and the output layer respectively.

The structure of the FFNN contains only limited information about the foregoing and has some limitations on the length of the input sequence. Therefore, the RNN LM comes into being. As shown in Fig.~\ref{picture1} (b), RNN can accept input of any variable length. When the input window is moved, its internal state mechanism can avoid repeated calculation, and parameter sharing further reduces the number of model parameters. Therefore, compared with FFNN, RNN has a great advantage.

The pretraining LM is to get a set of model parameters by pretraining some tasks. It initializes the model with these parameters and then trains to improve the model performance effectively.
The commonly used pretraining models are fixed embedding (Word2vec~\cite{DBLP:journals/corr/abs-1301-3781}, Glove~\cite{DBLP:conf/emnlp/PenningtonSM14}, etc), variable embedding (Embeddings from LMs (ELMO)~\cite{DBLP:conf/naacl/PetersNIGCLZ18}, Generative Pretrained Transformer (GPT)~\cite{radford2018improving} and Bidirectional Encoder Representations from Transformers (BERT)~\cite{DBLP:conf/naacl/DevlinCLT19}, etc). 
Here, we give an example of the GPT model, as shown in Fig.~\ref{picture1} (c). It adopts a two-stage process. In the first stage, the Transformer decoder is used as the basic unit of the model to perform text prediction. In the second stage, the GPT is initialized differently for different downstream tasks, training the model and fine-tuning the parameters.

\subsection{Basic Components on GL}

Due to the extensive use of graph data in many fields, some communities (e.g., chemistry, protein, and social network) have recently focused on the study of graph pretraining.
These pretraining models encode graph attributes, structures, and other information into node representations from multiple perspectives by designing different pretext tasks, which are used to optimize downstream tasks.
In this section, we introduce the definition of the basic concepts of graphs, and then provide a formal definition of the PFM on the graph.

\subsubsection{Notations and Definitions of Graphs}

Unless particularly specified, the notations used in this article are illustrated in Table~\ref{tab:graph_notation}.
We use $\mathcal{G}=\{G_{i}\}^{N}_{i}$ to represent a set of graphs, where $N$ represents the number of graphs.
Depending to the graph's definition of the edges and nodes, graph data can be classified into the following types.

\begin{Def}
	\label{def: unattr_graph}
	An \textbf{unattributed graph} is $G=(V,E)$, where $v \in V$ is a node, $e \in E$ is an edge, and naturally $E \subseteq V \times V$.
	Adjacency matrix $A \in \mathbb{R}^{n \times n}$ represents the topology of graph $G$, where $n=|V|$. $A_{i,j}=1$ denotes there is an edge between node $v_{i}$ and $v_{j}$, otherwise $A_{i,j}=0$.
\end{Def}

\begin{Def}
	\label{def: attr_graph}
	An \textbf{attributed graph} is $G=(V,E,X_{v},X_{e})$, where $X_{v} \in \mathbb{R}^{n \times d_{v}}$ and $X_{e} \in \mathbb{R}^{m \times d_{e}}$ are the feature matrices of nodes and edges, $|V|=n$, $|E|=m$, $d_{v}$ and $d_{e}$ denotes the feature dimensions of node and edge.
	In fact, in most application scenarios, only nodes have attributes, and edges have no attributes or only weights.
\end{Def}

\begin{Def}
	\label{def: undirected_graph}
	An \textbf{undirected graph} is $G=(V,E)$, where $e_{i,j} \in E$ means an unordered node pair $(v_{i}, v_{j})$.
	In particular, the adjacency matrix $A$ of the undirected graph is a symmetric matrix (i.e., $A_{i,j}=A_{j,i}$).
\end{Def}

\begin{Def}
	\label{def: directed_graph}
	A \textbf{directed graph} is $G=(V,E)$, where $e_{i,j} \in E$ means an ordered node pair $(v_{i}, v_{j})$.
\end{Def}

\begin{Def}
	\label{def: homo_graph}
	$G$ has a node-type mapping function $f_{v}: V \to \mathcal{T}^{v}$ and an edge-type mapping function $f_{e}: E \to \mathcal{T}^{e}$.
	When $|\mathcal{T}^{v}|=|\mathcal{T}^{e}|=1$, the graph $G=(V,E)$ is a \textbf{homogeneous graph}.
	In other words, all nodes in $G$ belong to a type, and all edges also belong to one type.
\end{Def}

\begin{Def}
	\label{def: hete_graph}
	When $|\mathcal{T}^{v}|>1$ and/or $|\mathcal{T}^{e}|>1$, the graph $G=(V,E)$ is a \textbf{heterogeneous graph}.
	In particular, a heterogeneous graph must be an attributed graph.
\end{Def}

\subsubsection{Learning Settings on Graphs}
GL methods are usually used to solve machine learning tasks on graph data, and we introduce different settings (supervision mode and learning mode) for GL. 

Before that, we first provide the notations of the corresponding mathematical formulation of GL.
$C=\{c_{1}, c_{2}, \cdots, c_{K}\}$ is a set of target components defined in a graph set $\mathcal{G}$ ($G^{c_{i}} \in \mathcal{G}$), and $c_{i}$ is associated with a corresponding ground truth label $y_{i} \in \mathcal{Y}=\{1,2,\cdots,N_{y}\}$, where $K$ denotes the total number of target components, and $N_{y}$ is the number of classes being predicted.
Then the graph data can be represented as $D=\{c_{i}, G^{c_{i}}, y_{i}\}^{K}_{i}$, and a complete GL model $M_{GL}$ can also be determined by $y_{i} = M_{GL}(c_{i}, G^{c_{i}})$.
For instance, in a node classification task, $c_{i}$ is the node to be classified, $y_{i}$ denotes $c_{i}$'s label in graph $G^{c_{i}}$. Similarly, in a node clustering task, $c_{i}$ is the node to be clustered, $y_{i}$ denotes the corresponding cluster label in graph $G^{c_{i}}$.

\myparagraph{\textbf{Supervision Mode}}
\begin{figure*}[!t]
	\centering
	\includegraphics[scale=0.15]{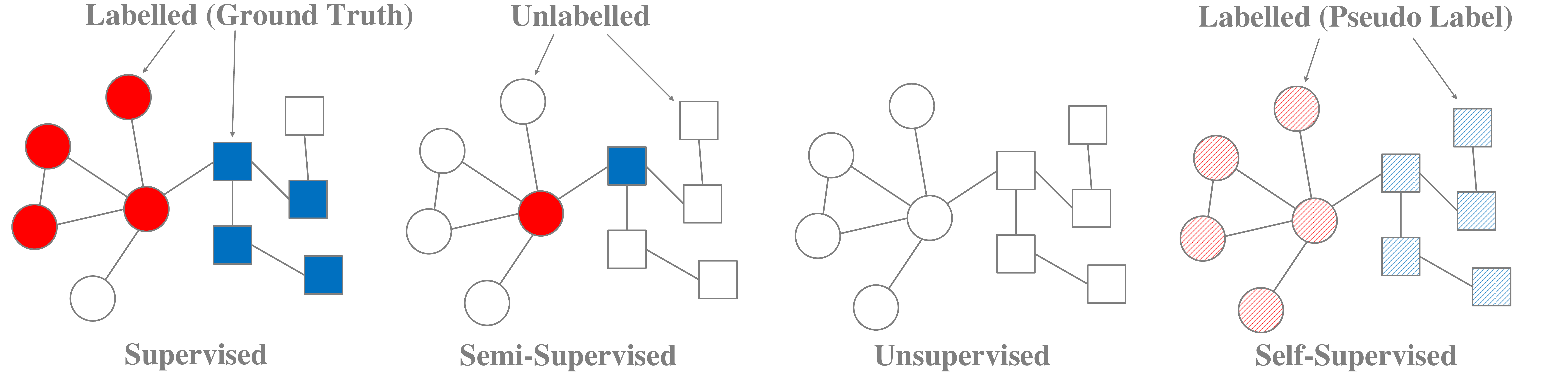}
	\caption{Schematic of different supervision modes.}
	\label{fig:supervision}
\end{figure*}
Depending on the source and scale of the training data, the supervision settings of GL can be divided into four types as shown in Figure~\ref{fig:supervision}.
\textbf{Supervised} GL is the most common mode in the real scenario.
Given the target component $c_{i}$ and the corresponding ground truth label $y_{i}$, the goal is to minimize the loss function between the predicted label of the GL model (i.e., $y^{pred}_{i}=M_{GL}(c_{i}, G^{c_{i}})$) and the expected label $y_{i}$ of all $c_{i}$.
Compared with supervised learning, \textbf{unsupervised} GL refers to situations in which no labeled data is provided, only the attributes and structure distribution of graph data (i.e., $(c_{i}, G^{c_{i}})$) can be used.
\textbf{Self-supervised} GL is a special case of both supervised and unsupervised learning. 
Specifically, self-supervised learning mainly uses pretext tasks (e.g., clustering, completion, and partition) to mine its own supervised information (i.e., pseudo-labels) from large-scale unsupervised graph data, and trains the GL model $M_{GL}$ through the self-supervised information, so that it can learn to the valuable features of downstream tasks.
In other words, the supervised information of self-supervised learning is not manually labeled, but the pretext tasks automatically construct supervised information from large-scale unsupervised data for supervised learning or training.
\textbf{Semi-supervised} learning is a combination of unsupervised and supervised learning, who aims at learning data distribution to predict unlabeled data to solve the problem of difficulty in obtaining labeled data in real scenarios.
In GL, semi-supervised learning refers to the realization of pattern recognition given a few labeled data and mass unlabeled data.

\myparagraph{\textbf{Learning Mode}}
The GL model $M_{GL}$ is optimized by the given training samples, and adjusted on the validation samples to participate in the test.
According to the visibility of the graph data at different stages, the learning settings of GL model $M_{GL}$ can be classified into two categories: inductive learning and transductive learning.

\begin{Def}
	\label{def: ind_learning}
	\textbf{Inductive Learning}, which is the most common setting in machine learning tasks, trains the model on labeled data and then tests on samples that have never appeared in the training stage.
	Formally, given a training sample $\{(c_{i}, G^{c_{i}}, y_{i})\}^{N_{l}}_{i=1}$, $\{(c_{j}, G^{c_{j}})\}^{N_{u}}_{j=1}$, where $N_{l}$ and $N_{u}$ are the numbers of labeled/unlabeled samples.
	Inductive learning learns a function $f^{ind}: \mathcal{G} \mapsto \mathcal{Y}$ so that $f^{ind}$ is expected to be a good classifier on the future graph data $\{(c_{k}, G^{c_{k}})\}$, beyond $\{(c_{j}, G^{c_{j}})\}^{N_{u}}_{j=1}$.
\end{Def}

\begin{Def}
	\label{def: trans_learning}
	\textbf{Transductive Learning} is different from inductive learning in that all samples are visible during both the training and testing stages.
	Formally, given a training sample $\{(c_{i}, G^{c_{i}}, y_{i})\}^{N_{l}}_{i=1}$, $\{(c_{j}, G^{c_{j}})\}^{N_{u}}_{j=1}$, transductive learning learns a function $f^{trans}: \mathcal{G}^{l+u} \mapsto \mathcal{Y}^{l+u}$ so that $f^{trans}$ is expected to be a good classifier on the unlabeled data $\{(c_{j}, G^{c_{j}})\}^{N_{u}}_{j=1}$.
\end{Def}

Under the supervised setting (including semi-/self-supervised), the unified classifier optimization methods of inductive learning and transductive learning can be written as:
\begin{equation}\label{eq:uniformloss}
	\mathcal{L}= \frac{1}{K} \sum^{K}_{i=1} \mathcal{L}(f^{(\cdot)}_\theta(c_{i}, G^{c_i}), y_i),
\end{equation}
where $\mathcal{L}$ is the cross-entropy loss, $c_{i}$ can be node, edge or subgraph of its associated graph $G^{c_{i}}$, and $f^{(\cdot)}_\theta$ denotes inductive/transductive function with parameter $\theta$.

Compared with using only one pretext task, some methods have designed some integration mechanisms to incorporate the advantages of multiple pretext tasks into a unified framework.

\section{Traditional Learning Methods} 

\subsection{Traditional Text Learning}

NLP is a research field that integrates linguistics and computer science. Its main research tasks include part-of-speech tagging, named entity recognition, semantic role labeling, machine translation, question answering, sentiment analysis, text summarization, text classification, relationship extraction, event extraction, etc. 
The LM can be considered the cornerstone of the downstream NLP tasks. 
It experiences four processes: grammar rule LM, probabilistic LM, neural network LM, and pretraining LM.
A PFM trains on a large benchmark dataset to obtain a model which can solve new similar tasks, which has become a new hotspot in current LM research.


Word representations play a significant role in downstream tasks, which is the basis of NLP.
The N-gram model preprocesses text features and encodes adjacent $N$ words as a group, which makes it overly dependent on the richness of the training corpus. Otherwise, data-sparse is likely to occur, and the computational complexity will increase exponentially with the increase of $N$.
Neural Network LM (NNLM)~\cite{DBLP:journals/jmlr/BengioDVJ03} adopts the idea of word vector for the first time, and the low-dimensional word vector of distributed representation can solve the discrete problem caused by word embedding well. However, it is still challenging to solve the problem of high computational complexity.
The computational complexity of the word2vec model is independent of the selected window size but is determined by the dictionary size and the word vector dimension.
Many downstream tasks can be significantly improved by training on a large corpus using word vector embedding after initial training.
However, the problem of polysemy for the static word vector is still unsolved, and it still belongs to the shallow LM~\cite{DBLP:conf/nips/MikolovSCCD13}~\cite{DBLP:journals/jmlr/CollobertWBKKK11}. Therefore, more effective models are urgently needed to deal with the dataset more flexibly.
To capture high-level concepts of context, such as polysemy elimination, syntactic structure, etc.
Neelakantan et al.~\cite{DBLP:conf/emnlp/NeelakantanSPM14} propose to learn multiple embeddings per word type. Zhou et al.~\cite{DBLP:conf/coling/ZhouQZXBX16} integrate the features on both dimensions of the matrix to enrich semantics by using subword information.
Based on the Continuous Bag Of Words (CBOW)~\cite{DBLP:journals/corr/abs-1301-3781} in word2vec, Hui et al.~\cite{DBLP:journals/2017Hui} fine-tune the generated word vectors for emotion and obtain the word vectors containing both semantic meaning and emotional tendency, which significantly improved the performance in the Weibo sentiment classification task.
Liu et al.~\cite{liu2017hierarchical} propose a model of hierarchical translation for machine translation. It uses the neural LM based on RNN as the word vector generation model. 
Liang et al.~\cite{DBLP:journals/2018N} propose an approach based on the double-layer self-attention mechanism for machine reading comprehension, and the model is divided into three parts: single document encoder, multi-document encoder, and answer prediction.
In the single document encoder, the problem of the context information is represented by the Gated Recurrent Unit (GRU) model.
Zhang et al.~\cite{zhichang2019user} propose an INDependent RNN (INDRNN) and attention mechanism for user intention classification, using word vectors generated by word2vec as input.
The model introduces a word-level attention mechanism to effectively quantify the contribution of domain vocabulary to the intention category.

\subsection{Traditional Image Learning} 
There are several types of neural networks in the deep learning era, from the beginning of most famous convolutional neural networks (CNNs) to the subsequent Attention- and Transformer-based networks. A deep neural network refers to an artificial neural network with more hidden layers, and more parameters are used to represent the target model, which leads to the SOTA performance on the benchmark dataset from image to video. Here, we introduce the milestone networks in CV chronologically.

\subsubsection{Convolution-Based Networks.}
 ImageNet~\cite{deng2009imagenet}, as one of the most important databases in computer vision, has aroused many milestone network architectures in image classification, including AlexNet~\cite{krizhevsky2012imagenet}, NIN~\cite{lin2013network}, VGG~\cite{simonyan2014very}, GoogLeNet~\cite{szegedy2015going}, ResNet~\cite{he2016deep}, DenseNet~\cite{huang2017densely}, etc. When it comes to object detection and semantic segmentation, researchers explore R-CNNs~\cite{girshick2014rich, girshick2015fast, ren2015faster, he2017mask}, FCN~\cite{long2015fully}, SSD~\cite{liu2016ssd}, YOLOs~\cite{redmon2016you, redmon2017yolo9000, redmon2018yolov3, bochkovskiy2020yolov4, chen2021you}, SegNet~\cite{badrinarayanan2017segnet}, PSPNet~\cite{zhao2017pyramid}, Deeplabs~\cite{chen2014semantic, chen2017deeplab, chen2017rethinking, chen2018encoder}, RefineNet~\cite{lin2017refinenet}, etc. on common benchmark datasets, such as PASCAL VOC~\cite{everingham2015pascal, everingham2010pascal}, MS COCO~\cite{lin2014microsoft}, etc. 

There are several shared features among these popular convolution-based networks: 1) \emph{data augmentation}. Deep models require much more data to fit a complicated model, thus the data augmentation technique such as flipping, rotation, cropping, scaling, translation, and even adding noises enlarges the training dataset; 2) \emph{convolution}. The convolutional kernel is used to extract the features of original image data, which maintains the spatial structure for the adjacent pixels; 3) \emph{deep architecture}. The deep architecture contains more parameters, which enhance the capability of the model. These common features contribute to the SOTA performance of convolutional neural networks (CNNs) in computer vision for nearly recent 10 years. 

\subsubsection{Recurrent neural networks}
Different from CNNs targeting 2D-dimensional image applications, recurrent neural networks (RNNs)~\cite{rumelhart1986learning, jordan1997serial, elman1990finding} try to use recursive cells to process pictures in sequence, i.e., video data. However, the weaknesses of gradient explosion and long-term dependencies restrict further development of this model. To handle these problems embedded inside the RNN-based models, long short-term memory (LSTM)~\cite{hochreiter1997long} was proposed by Hochreiter and Schmidhuber in 1997. In addition, the improved capability of LSTMs produces popularity and attracts attention both in NLP and CV~\cite{graves2008novel, vinyals2015show, sutskever2014sequence, graves2013generating, sundermeyer2012lstm}.

\subsubsection{Generation-Based Networks}
Generative Adversarial Networks (GANs)~\cite{goodfellow2014generative} have provided a paradigm to learn representations for unlabelled data, and spawn many GAN-based approaches on downstream tasks. In image translation, pix2pix software~\cite{isola2017image} first proposes the conditional adversarial networks as a solution to the image-to-image translation problems, and achieves reasonable results on real-world datasets. 
Markovian Generative Adversarial Networks (MGANs)~\cite{li2016precomputed} is a method to generate texture synthesis, which can be applied to style transfer and video
stylization. CycleGAN~\cite{zhu2017unpaired} provides a learning algorithm to translate an original image from the source domain to a target domain without containing pairs of images in datasets for supervised learning. StyleGAN~\cite{karras2019style} is a style-based generator to serve as an alternative architecture for traditional GANs. Pixel Recurrent Neural Networks (PixelRNN)~\cite{van2016pixel} aims to complete images by modeling full dependencies between the color channels. DiscoGAN~\cite{kim2017learning} is designed to learn relations between different domains. 

GANs have also provided a novel direction to study data synthesis because it perfectly simulates the distribution of the original data. Laplacian Pyramid of Adversarial Networks (LAPGAN)~\cite{denton2015deep} uses a cascade of convolutional networks to generate images in a coarse-to-fine fashion. Similarly, Stacked Generative Adversarial Networks (SGAN)~\cite{huang2017stacked} decompose variations into multiple levels and gradually resolve uncertainties by stacking several GANs in a top-down way.

\subsubsection{Attention-Based Networks}
Based on the success of CNNs in the area of CV, the attention module is designed to equip with the popular CNNs. For example, SENet~\cite{hu2018squeeze} proposes a channel attention module, which won first place in the competition of ILSVRC2017. In addition,  CBAM~\cite{woo2018cbam} sequentially infers attention maps along both channel and spatial dimensions. Many innovative works, such as GCNet~\cite{cao2019gcnet} and CCNet~\cite{huang2019ccnet}, are inspired by this idea of soft-attention mechanism, which outperforms the traditional CNNs on major benchmarks for both recognition and segmentation tasks. In particular, the self-attention mechanism~\cite{zhang2019self}, calculating the response at a position among all entities in a sequence by attending to all positions within the same sequence, is proposed to estimate the relevance of one position to other positions in feature maps. To control the expected entities and model more complex relations among different elements in the sequence, masked self-attention and multi-head attention~\cite{vaswani2017attention} are the key components proposed to substitute the function of convolutions in the era of transformers. 

\subsubsection{Transformer-Based Networks}
Recently, inspired by the self-attention mechanism and subsequent success of the transformer in NLP, researchers in CV also try to use the transformer as an alternative to the convolution. Self-attention-based transformer models always operate in a two-stage training mechanism: 1) pretraining on a primitive dataset (always big but not well labeled) by defining pretext tasks; 2) transferring the pretrained weights to the downstream tasks and adjusting the parameters on the target domain dataset by finetuning. Vision Transformer (ViT)~\cite{dosovitskiy2020image} is applied on CV and achieves the SOTA performance on major benchmark datasets. Data-efficient image Transformers (DeiT)~\cite{touvron2020training}was proposed by Facebook AI to train image transformers more efficiently and maintain the SOTA performance simultaneously. DEtection TRansformer (DETR)~\cite{carion2020end} significantly outperforms competitive baselines in both object detection and semantic segmentation. LeViT~\cite{graham2021levit} outperforms existing benchmarks with respect to balancing the accuracy and training speed. Image GPT~\cite{chen2020generative} is inspired by a sequence transformer in NLP, which can compete with several self-supervised benchmarks on ImageNet. On the basis of this research, DeepViT~\cite{zhou2021deepvit} explores a deeper architecture to improve performance consistently by making the transformer go deeper. Moreover, many researchers try to apply the transformer to more specific tasks. Pyramid Vision Transformer (PVT)~\cite{wang2021pyramid} introduces the pyramid structure to overcome the difficulties of porting the transformer to various dense prediction tasks, and achieves the SOTA performance on major benchmark datasets. M3DeTR~\cite{guan2021m3detr} is a novel research on multi-representation, multi-scale, and mutual-relation 3D object detection with transformers. Medical Transformer (MedT)~\cite{valanarasu2021medical} has focused on medical image segmentation and outperforms previous CNN-based and transformer-based architecture. In conclusion, the transformer has become a novel and popular research area in CV and its performance is proved by many existing works.

\subsection{Traditional Graph Learning} 

GL aims to embed the graph as a low-dimensional representation while preserving the desired properties of the original graph data.
Classical GL methods are usually implemented using statistical methods or artificially designed components.

\myparagraph{\textbf{Dimension Reduction}}
As a commonly used method in feature engineering, dimension reduction aims to reduce the dimension of high-dimensional attribute graph data into a lower-dimensional representation.
In GL, it highlights the remaining information at the cost of losing part of the attributes.
According to different dimensionality reduction strategies, such methods can be classified into two types.
The first type is subspace learning under the linear assumption.
Based on the assumption that the principal components~\cite{pca1976tse} related to the larger variance represent important structural information, and those smaller variances represent noise, principal component analysis calculates a low-dimensional representation that maximizes the variance of the data.
Linear Discriminant Analysis (LDA)~\cite{lda2004nips} achieves dimension reduction by maximizing the ratio of inter-class scattering and intra-class scattering to obtain a linear projection matrix.
Multi-Dimensional Scaling (MDS)~\cite{mds1995amj} is a distance-maintaining manifold learning method. It produces a mapping in a lower dimension to preserve dissimilarities between nodes as much as possible.
The second type is nonlinear dimension reduction, which aims to automatically learn nonlinear topology to achieve manifold learning.
Isomap~\cite{isomap2006prl} first constructs a neighborhood graph on the manifold and calculates the shortest path between pairs of nodes, and then uses MDS to construct a low-dimensional embedding.
Locally Linear Embedding (LLE)~\cite{lle2000sci} first allocates neighbors for each node. 
Then, it calculates the weighted $W_{i,j}$, the best linear reconstruction feature $X_{i}$ from its neighbors. 
Finally, calculate the low-dimensional embedding for the optimal reconstruction of $W_{i,j}$.

\myparagraph{\textbf{Matrix Factorization}}
Greatly influenced by the idea of dimension reduction, the models based on matrix factorization emerged in the early research of GL.
Such models aim to reconstruct the adjacency matrix of the graph to achieve dimension reduction while maintaining structural information.
Although these models have significant limitations, in fact, their ideas still inspire many current studies.
Depending on how the matrix is constructed, such methods often append specific constraints.
Graph Laplacian eigenmaps~\cite{ledrdr2003nc} minimizes a loss function to ensure that nodes close to each other on the manifold are mapped into the low-dimensional space and still maintain the local distances.
Node proximity matrix factorization~\cite{rl2008kdd} minimizes the objective function $|W-YY^{cT}|$ through matrix factorization to approximate the proximity of nodes in the low-dimensional space, where $Y$ and $Y^{c}$ are the embeddings for nodes and context nodes, and $W$ is the default node proximity matrix.
GraRep~\cite{gragep2015cikm} aims to preserve the high-order proximity of graphs in the embedding space, thus it derives a $k$-th order transition matrix, $A^{k}$, by multiplying the adjacency matrix to itself $k$ times. 
The transition probability from node $v_{i}$ to node $v_{j}$ is the entry in the $i$-th row and $j$-th column of the $k$-th order transition matrix, i.e., $p_{k}(v_{i}|v_{j})=A^{k}_{i,j}$.
Then GraRep defines the loss function using the skip-gram model and negative sampling.
To capture the high-order proximity between node pairs, HOPE~\cite{hope2016kdd} preserves asymmetric transitivity in approximating
the high-order proximity.
Specifically, the goal of HOPE is to minimize the objective function $||S-WC^{T}||^{2}_{F}$, where the elements $s_{i,j} \in S$ represent a certain edge feature (e.g., Katz index, the Rooted Page-Rank, the Common Neighbors, and the Adamic-Adar) between the corresponding node pairs $(v_{i}, v_{j})$, $W$ is the node representation matrix, and $C$ is the embedding of the node as the context.
To reconstruct the matrix $S$ more simply and elegantly, HOPE proposes to obtain $W$ and $C$ directly based on the low-rank singular value decomposition (SVD). 

\myparagraph{\textbf{Graph Kernel}}
The kernel method is an important algorithm in pattern recognition and machine learning.
Its basic idea is to give the graph embedding $x \in X$ in the original low-dimensional space $X$, and maps the embeddings to a high-dimensional feature space $H$ through a nonlinear function $f^{ker}$.
Then the nonlinear problem in $X$ can be solved by constructing a linear algorithm in $H$.
There are two main types of kernel methods on graph data.
The first type uses the embedding method to convert the graph data into vectorial representation, and then directly implements the application based on the kernel function.
However, due to the loss of mass graph structure information when transforming graphs into vectorial representation, such methods do not perform well in real scenarios.
The second type of method introduces the graph kernel function to solve this problem. 
Based on retaining the advantages of the original kernel function, it directly represents the structural information of the graph data in the high-dimensional Hilbert space.
The definition of the traditional method of graph kernel comes from R-convolution. 
According to the difference between the contrast substructure and the decomposition method of the graph structure, a large number of methods based on graph kernel have been proposed.
For example, the work of~\cite{rd12015nips,frd2012siam} proposed a random-walk kernel based on calculating the number of common synchronization between two graph structures,
To reduce the computational complexity and optimize the random walk strategy, a graph kernel based on comparing the shortest path information between two graph structures is proposed.
To capture more complex topological information, the Weisfeiler-Lehman subtree graph kernel is proposed, which is based on a one-dimensional Weisfeiler-Lehman isomorphism test algorithm to find isomorphic subtree structures in a bunch of graph structures~\cite{wlgk2011jmlr}.

\section{PFMs Theory} \label{Section 9}


Since pretraining has received great attention from the research community, the investigation in the theory-backed explanation is similarly eye-catching. During the unsupervised pretraining era before SSL, Erhan et al.~\cite{erhan2009difficulty,erhan2010does} shed some light on the theoretical explanation for the confirmation and clarity of learning difficulties. \cite{erhan2009difficulty} researches the influence of pretraining with respect to architecture depth, model capacity, and the number of training samples, and demonstrates the robustness of pretraining from the perspective of both the optimization and the regularization. \cite{erhan2010does} further prove the regularizer role of the unsupervised pretraining in the downstream supervised tasks.

\subsection{Different Perspectives}

\myparagraph{\textbf{Pretext Tasks}} 
\cite{lee2020predicting} posits a mechanism based on approximate conditional independence (CI) to connect pretext and downstream task data distributions, which suggests that pretext tasks can self-supervisedly learn the representations from unlabelled data that reduce the sample complexity of downstream supervised tasks. The experiments both on CV and NLP task supports this theory. Representation Learning via Invariant Causal Mechanisms (R\textsc{e}LIC)~\cite{DBLP:conf/iclr/MitrovicMWBB21} also provides a theoretical understanding from the perspective that the explicit invariance constraints across augmentations can yield improved generalization guarantees.

\myparagraph{\textbf{Multi-View Redundancy}}
From the perspective of a multi-view setting, \cite{tosh2021contrastive} understands contrastive learning as exploiting multiple views of data for representation learning.
This theory provides a theoretical analysis that the linear functions of these representations from pretraining are still competitive compared with the non-linear optimal predictor of the label. In other words, the linear functions of the learned representations are nearly optimal on downstream prediction tasks whenever the different views provide redundant information about the label.

\subsection{Different Categories}

\myparagraph{\textbf{Contrastive Learning}}
Although experimental results show us that previous designs such as contrastive loss or momentum updating can produce impressive performance in SSL. However, one of the most important questions that remain in SSL is why these methods can maintain representation consistency during the pretraining process. A naive view is the minimization between positive pairs can boost invariance learning, while the maximization between negative pairs contributes to avoiding representational collapse. \cite{arora2019theoretical} shows that contrastive learning can achieve competitive bound via intra-class concentration, thus leading to the reduction of sample complexity on downstream tasks from the benefit of transferred representations. This research also provides a framework that can be utilized both on the guarantees of the quality of learning representations during the pretraining phase and the future assumptions added to the framework that allow tighter guarantees.

\myparagraph{\textbf{Non-Contrastive Learning}}
While contrastive learning shows an effect by capturing the similarity and dissimilarity among the unlabelled examples, and further converging to an average local optimum which represents the general representations, recent non-contrastive SSL methods such as BYOL and SimSiam also shows the SOTA performance without the design of comparison between negative pairs. Based on the analysis of the eigenspaces, Tian et al.~\cite{DBLP:conf/icml/TianCG21} study the behavior of non-contrastive SSL training and prove that the effects are from both the predictor and stop-gradient signal. Based on this theory, a novel and simple \textbf{DirectPred} method is proposed as a by-product of this theoretical exploration.

\section{Pretext Task Taxonomy on CV}
Pretext tasks are always designed to use pseudo labels generated from the data itself to pretrain the proxy model. 
There are five categories of pretext tasks for self-supervised: 1) generation-based methods; 2) transformation-based methods; 3) context-based methods; 4) semantic-based methods; 5) view-based methods.

\myparagraph{\textbf{Generation-Based Methods}} This type of method is GAN-based in the deep learning era. For image generation, there are several applications including image colorization~\cite{zhang2016colorful, anwar2020image}, image super-resolution~\cite{ledig2017photo}, image editing~\cite{perarnau2016invertible}, context encoders~\cite{pathak2016context}, image-to-image translation~\cite{zhu2017unpaired}, etc. On the other hand, video generation tasks contains future prediction~\cite{oord2018representation}, video action recoginition~\cite{lorre2020temporal}, video generation~\cite{vondrick2016generating, tulyakov2018mocogan}, and video representaion~\cite{wang2015unsupervised}.

\myparagraph{\textbf{Transformation-Based Methods}} Transformation is a typical technology that serves as a data augmentation method to enlarge the training dataset in traditional DL. However, if transformations of the same image are labeled as positive samples and others as negative samples, this pretext task can be used for self-supervised pretraining~\cite{chen2020simple}. Popular transformation in self-supervised learning (SSL) contains color transformation (such as Jitter, Gaussian blur, and adjusting brightness) and geometric transformation (such as flipping, cropping, scaling, and rotation).

\myparagraph{\textbf{Context-Based Methods}} Basically, the design and construction of many artificial tasks, such as solving Jigsaw puzzles~\cite{noroozi2016unsupervised}, comparing context similarity, and discriminating sequence order. Solving Jigsaw puzzles is defined as identifying the correct position of patches from an image. This task can help the model to learn an encoder for transfer learning~\cite{wei2019iterative, kim2018learning}, and the feature representations are effective after the pretrained dataset is big enough. In addition, the design of video Jigsaw is also proposed for unsupervised learning~\cite{ahsan2019video}. Differently, context similarity tries to label the patches from the same images as positive samples and others as negative samples, then use a predefined similarity function to scale the distance between different pairs~\cite{caron2018deep}.

\myparagraph{\textbf{Semantic-Based Methods}} Semantic-based methods contain object detection, semantic segmentation, and depth prediction. These tasks also involve pretext tasks because their pixel-based labels can learn a more robust feature representation than simpler tasks. These pre-text tasks always establish on video dataset~\cite{pathak2017learning, croitoru2017unsupervised}.

\myparagraph{\textbf{View-Based Methods}} This type of method contains both single-modal data and multi-modal data. For the single-modal data, the original data is treated as the anchor and different viewpoints generate its positive pair samples. Sometimes the time slices in sequence-based data are treated as negative pairs because the scene is changed as time goes~\cite{sermanet2018time}. In addition, multi-modal data is usual in view-based methods, which are also called cross-modal-based methods here. Such as audio-video cooperative learning~\cite{korbar2018cooperative}, RGB and optical flow cross-modal distance training~\cite{sayed2018cross}.

\section{PFMs for Reinforcement Learning}
The success of pretraining learning methods in the supervised learning domain has spurred interest in the reinforcement learning (RL) domain to study whether the same paradigms can be adapted to RL algorithms. General pretraining RL can include broad directions, such as Reward-Free RL \cite{stadie2015incentivizing, achiam2017surprise, pathak2017curiosity, tang2017exploration}, Goal-condition RL \cite{dey2019manipulating,han2021learning, ding2019goal}, and Representation Learning in RL \cite{shah2021rrl, xiao2022masked, schwarzer2021pretraining, schwarzer2020data}. Here we focus the Representation Learning in RL. Specifically, this direction seeks to \textit{improve the performance by pretraining the visual perception competent of RL agent, i.e., the state encoder, with some large-scale datasets using unsupervised/self-supervised data augmentation techniques}. The pretraining process empowers the state encoder to capture the essential structure information from the raw inputs (pixel-level input for CV). An RL policy network is built based on the pretrained state encoder to learn the specific downstream control tasks in the fine-tuning stage. Recent studies have demonstrated that  can greatly benefit both in sample efficiency and learning effectiveness from unsupervised \cite{haWorldModels2018, jaderbergReinforcementLearningUnsupervised2016, higginsDARLAImprovingZeroShot2018}, semi-supervised \cite{finn2016generalizing}, and self-supervised \cite{shahRRLResnetRepresentation2021,schwarzerDataEfficientReinforcementLearning2021} learning techniques. Specifically, this direction could be roughly classified into the following two categories: 
Model-based Pretraining RL and Contrastive-like Pretraining RL. 

\paragraph{Model-based Pretraining RL} Model-based Pretraining RL aims to first pretrain a generative world model to capture the underlying structure of the environment and then leverage the world model as a state encoder or simulator during fine-tuning. World Models \cite{haWorldModels2018} is the first work that proposes to learn a compressed spatial and temporal representation of the environment in an unsupervised manner using a simple Variational Autoencoder, which greatly improves the sample efficiency compared to training from scratch. However, learning the world model without being aware of the environment's dynamic might lead to ignorance of some key information in the environment. Dreamer \cite{hafner2019dream, hafner2020mastering} proposed to learn latent dynamics by approximating the representation, transition, and reward model. They then train RL agents purely by imagination in a latent space, which is more efficient since it brings a low memory footprint and enables fast predictions of thousands of imagined trajectories in parallel.
Furthermore, DreamerPro \cite{deng2022dreamerpro} proposes a reconstruction-free approach based on prototypical representations to migrate the task-irrelevant visual distractions problem in the latent dynamics modeling. DreamerPro significantly outperforms previous SOTA methods when there are complex background distractions. To verify whether learning accurate world models for the real world is promising, Daydreamer \cite{wu2022daydreamer} applies Dreamer to the real-world physical robots problem and empirically demonstrates significant learning efficiency gains. 

\paragraph{Contrastive-like Pretraining RL} Contrastive-like Pretraining RL techniques seek to improve the representation ability of state encoders by pretraining the state encoder with a large amount of out-of-domain data or adding some auxiliary loss using unsupervised learning or data augmentation techniques. 
CURL \cite{laskin2020curl} combines instance contrastive learning and  by using MoCo \cite{he2020momentum} mechanism, which significantly improves the data efficiency of RL agents. Furthermore, RAD \cite{laskin2020reinforcement} proposes an implicit approach that directly trains the RL objective on multiple augmented observations views, which outperforms CURL on some of the environments in the DeepMind Control Suite. Concurrent to RAD, DrQ \cite{kostrikov2020image} introduces a simple regularization term, which applies image augmentation to compute current and target Q values. They demonstrate that data efficiency can be significantly improved after applying it to DQN. DrQ-v2 \cite{yarats2021mastering} further extends this approach to solve complex humanoid locomotion tasks by inserting similar techniques into the DDPG algorithm. Orthogonal to this direction, \cite{xiao2022masked,shah2021rrl,nair2022r3m, parisi2022unsurprising} demonstrate that pretraining the vision part of RL agent using supervised or unsupervised methods on out-of-domain data can improve the learning efficiency of downstream RL control tasks. Besides ensuring consistency across different views of observation, SPR \cite{schwarzer2020data} additionally trains a dynamics model which enforces the representations to be temporally predictive. Based on SPR, SGI \cite{schwarzer2021pretraining} proposes to pretrain representations using a combination of latent dynamics modeling, unsupervised goal-conditioned, and inverse dynamics modeling. Compared to previous methods, SGI can better capture the environment's dynamics and facilitate downstream RL control task training. 



\section{Evaluation Metrics}\label{Evaluation_Metrics}




\paragraph{Classification Task}
The classification task, according to a labeled training document, determines the relationship between document features and document categories. The learned relationship model is then used to determine the category of new documents.

\myparagraph{Accuracy and Error Rate} 
The key metrics for a text classification model are Accuracy and Error Rate. The terms Accuracy and Error Rate are defined as follows:
\begin{equation}
Accuracy =\frac{(\mathrm{TP}+\mathrm{TN})}{N},
\end{equation}
\begin{equation}
Error Rate = 1 - Accuracy =\frac{(\mathrm{FP}+\mathrm{FN})}{N},
\end{equation}
where $\mathrm{TP}$ and $\mathrm{FP}$ denote true positive and false positive, $\mathrm{TN}$ and $\mathrm{FN}$ stand for true negative and false negative.

\myparagraph{Precision, Recall and F1}
Regardless of the standard type and error rate, there are very important metrics used for unbalanced testing sets. These metrics are similar to the concept of the class label in the testing samples. F1 is defined as the harmonic average of Precision and Recall. Thus, Accuracy, Recall, and F1 can be represented as:

\begin{equation}
Precision =\frac{\mathrm{TP}}{\mathrm{TP}+\mathrm{FP}}, \quad
Recall =\frac{\mathrm{TP}}{\mathrm{TP}+\mathrm{FN}},
\end{equation}
\begin{equation}
F1 =\frac{2 \mathrm { Precision \times Recall }}{\mathrm{ Precision }+\mathrm{Recall}}.
\end{equation}


When the accuracy, F1, and recall values hit 1, the desired results are obtained. On the other hand, when the values turn 0, we get the worst consequence. For the multi-class classification task, the precision and recall values of each class can be determined independently, and then the individual and overall performance can be analyzed.

\myparagraph{{$ Micro-F1$}} The $ Micro-F1$~\cite{DBLP:books/daglib/0021593} 
is a metric that measures all labels' overall accuracy and recall.
We denote $ Micro-F1$ as:
\begin{equation}
 Micro-F1=\frac{2 \mathrm{P}_{t} \times R_{t}}{\mathrm{P}+\mathrm{R}},
\end{equation}
\begin{equation}
 P=\frac{\sum_{t \in \mathcal{S}} T P_{t}}{\sum_{t \in S} T P_{t}+F P_{t}},\quad  R=\frac{\sum_{t \in S} T P_{t}}{\sum_{t \in \mathcal{S}} T P_{t}+F N_{t}}.
\end{equation}
where $T P_{t}$ and $F P_{t}$ mean true and false positive of the $t$ th label on a text.

\myparagraph{{$ Macro-F1$}} 
The $ Macro-F1$ calculates the average $F1$ of all labels by giving equal weight to them. $ Macro-F1$ is denoted as:

\begin{equation}
{Macro}-F1=\frac{1}{\mathcal{S}} \sum_{t \in \mathcal{S}} \frac{2 \mathrm{P}_{t} \times R_{t}}{\mathrm{P_{t}}+\mathrm{R_{t}}},
\end{equation}
\begin{equation}
P_{t}=\frac{T P_{t}}{T P_{t}+F P_{t}},\quad R_{t}=\frac{T P_{t}}{T P_{t}+F N_{t}}.
\end{equation}
where $T N_{t}$ and $F N_{t}$ represent true and false negative of the $t$ th label. $\mathcal{S}$ stands for the label set of all samples.

\myparagraph{Mean Reciprocal Rank (MRR)} 
The MRR is commonly used to evaluate the performance of ranking algorithms on Question Answering (QA) and Information Retrieval (IR) tasks. MRR is represented as

\begin{equation}
\mathrm{MRR}=\frac{1}{Q} \sum_{i=1}^{Q} \frac{1}{{rank}_{i}},
\end{equation}
where ${rank}_{i}$ is the ranking of the $i$ th ground-truth answer. 
The number of predicted labels on each text is denoted by $Q$.
Moreover, there are some metrics, such as EM, Hamming-loss~\cite{DBLP:journals/ml/SchapireS99}, P@K and NDCG@K.

\paragraph{Generation Task}
Generation task uses LMs to predict the next most likely word or sentence based on input data.

\myparagraph{Bilingual EvaLuation Understudy (BELU)} 
BLEU compares the generated sentences to the reference sentence and makes predictions using automatic machine translation algorithms. The language creation problem is also supported by deep learning technologies such as speech recognition, image caption generation, and text summarization. They can't discover anything better, but it has a few advantages: it's simple to comprehend, correlates well with human judgment, and is language-independent.
As a bilingual evaluation aid, BLEU is mainly used to evaluate the quality of machine translation~\cite{reiter2018structured}.
BLEU compares the degree of overlap between the N-gram in the candidate text and the N-gram in the reference text. The higher overlap indicates better translation quality.
The formula for the computation is:
\begin{equation}
B L E U=BP \times \operatorname{exp}\left(\sum_{n=1}^{N} W_{n} \log P_{n}\right),
\end{equation}
where $N$ represents N-gram, $BP$ is penalty factor, $P_{N}$ is multivariate precision, and $W_{N}=1/N$ is the corresponding weight of multivariate precision.
$r$ represents the length of the shortest reference translation, and $c$ represents the length of the candidate translation, then the specific calculation method of penalty factor $BP$ is as follows:
\begin{equation}
BP= \begin{cases}1, & l_{t}>l_{a} \\ e^{1-l_{a} / l_{t}}, & l_{t} \leq l_{a}\end{cases},
\end{equation}
where $l_{t}$ is the number of words in machine translation and $l_{a}$ is the number of words in reference answer.
The penalty factor is mostly used to penalize large gaps between machine and reference translations.

\myparagraph{ROUGE (Recall-Oriented Understudy for Gisting Evaluation)} 
ROUGE stands for N-gram co-occurrence statistics, which are used in automatic evaluation methods. It is expanded on the similarity of N-grams, which means that an N-gram is a subsequence of the main document text in terms of $N$ words.
There are four types of ROUGE, including ROUGE-N, ROUGE-L, ROUGE-W, and ROUGE-S.
The first two are commonly used, and the $N$ in rouge-N refers to N-gram, which is calculated similarly to BLEU, except BLEU is based on accuracy, while ROUGE is based on recall.
$L$ in ROUGE-L refers to the Longest Common Subsequence, which is calculated as the Longest Common Subsequence between the candidate abstract and the reference abstract. Thus, the longer the length, the higher the score, based on the $F$ value.
The calculation formula of ROUGE-N and ROUGE-L is mainly introduced. The calculation formula of ROUGE-N is as follows:
\begin{equation}
    ROUGE-N= \\ \frac{\sum_{S \in\{\text {ReferenceSummaries}\}} \sum_{\text {gram}_{n} \in S} \text {Count}_{\text {match}}\left(\operatorname{gram}_{n}\right)}{\sum_{S \in\{\text {ReferenceSummaries}\}} \sum_{\text {gram}_{n} \in S} \text {Count}\left(\text {gram}_{n}\right)},
\end{equation}
where $N$ stands for N-gram, $Count(gram_{n})$ represents the frequency of occurrence of an N-gram, and $Count_{match}(gram_{n})$ represents the frequency of co-occurrence of an N-gram.
The calculation formula of ROUGE-L is as follows:
\begin{equation}
ROUGE-L=F_{lcs}=\frac{\left(1+\beta^{2}\right) R_{\mathrm{lcs}} P_{\mathrm{lcs}}}{R_{1 \mathrm{cs}}+\beta^{2} P_{\mathrm{lcs}}},
\end{equation}
\begin{equation}
R_{\mathrm{lcs}}=\frac{L C S(X, Y)}{M},
\end{equation}
\begin{equation}
P_{\text {lcs }}=\frac{L C S(X, Y)}{N},
\end{equation}
where $X$ is the candidate abstract, $Y$ represents the reference abstract, $LCS(X, Y)$ table indicates the length of the Longest Common Subsequence (LCS) of the candidate abstract and references abstract, $M$ stands for the length of reference abstract, and $N$ denotes the length of the candidate abstract.
The ROUGE method is characterized by N-gram co-occurrence statistics, based on recall rate (ROUGE-N) and F-value (ROUGE-L).
They are often used in text summaries.
It is worth noting that ROUGE is word-based correspondence rather than semantic-based correspondence, but this can be mitigated by increasing the number of reference summaries.

\myparagraph{METEOR} 
METEOR, also known as an explicitly sorted translation evaluation metric~\cite{denkowski2014meteor}, is an improved version of the BLEU standard that aims to address some flaws in the BLEU standard.
Using WordNet to calculate matching relationships between specific sequences, synonyms, roots, affixes, and definitions improves BLEU performance and makes it more relevant to manual discrimination.
The calculation formula is as follows:
\begin{equation}
METEOR=(1-Pen) \times F_{\text {m}},
\end{equation}
\begin{equation}
F_{\text {m}}=\frac{P R}{\alpha P+(1-\alpha) R},
\end{equation}
\begin{equation}
P=\frac{m}{\sum_{k} \text h_{k}(c_{i})},
\end{equation}
\begin{equation}
R=\frac{m}{\sum_{k} \text h_{k}(s_{ij})},
\end{equation}
where $Pen=\gamma(\frac{ch}{m})^{\theta}$ is a penalty factor, which punishes the word order in candidate translation that is different from that in reference translation. $ch$ refers to the number of chunks, which are clustered units of matched units adjacent to each other in both the candidate translation and the candidate reference translation. $\alpha, \beta, \theta$ is the adjustable parameter, $m$ is the number of unary groups that can be matched in the candidate translation, $c$ is the length of the candidate translation, $ h_{k}(c_{i})$ is the number of occurrences in candidate translations $c_{i}$, and  $ h_{k}(s_{ij})$ is the number of occurrences in reference translations $s_{ij}$.

\myparagraph{Perplexity} 
Perplexity is also called the degree of confusion~\cite{lin2003automatic}.
Its core idea is: first, according to the testing sentence, learn a LM $P$.
Then, according to the LM $P$, the score of the optional sentence is calculated.
Finally, the above scores are standardized according to sentence length.
The calculation formula is as follows:
\begin{equation}
P P L(W)=P\left(w_{1}, w_{2}, \ldots, w_{M}\right)^{-\frac{1}{M}},
\end{equation}
where $W$ is the candidate translation, $M$ is the length of the candidate translation, $P$ is the LM obtained according to the reference translation, and $P(w_{1}, w_{2}, \ldots, w_{M})$ is the score calculated by the LM for the candidate translation.
The Perplexity assessment indicator is based on a LM.
The lower the degree of confusion, the better the translation quality, which is often used in machine translation and LMs.
Its disadvantages are as follows: the larger the dataset is, the faster the degree of confusion decreases; the punctuation in the data will impact the PPL of the model; and the interference of common words.

\section{Datasets}\label{datasets}

\subsection{Downstream Tasks and Datasets on NLP}
There are many available datasets in the NLP domain, divided according to different tasks. We summarize them in Table~\ref{datasets1}.
It mainly comprises two categories: the task of classification of texts and the task of generating texts.
The text classification tasks mainly include Sentiment Analysis (SA), News Classification (NC), Topic Labelling (TL), Natural Language Inference (NLI), Named Entity Recognition (NER), Question Answering (QA), Dialogue Act Classification (DAC), etc. The generation tasks mainly include text summaries and machine translation.

\begin{table*}[]
\caption{The statistics of the datasets on NLP. For the QA task, the class represents the sum number of candidate answers and the correct answer. For dialogue, class is the number of slots. Length means the average tokens in turn.}\label{datasets1}
\resizebox{0.9\textwidth}{!}{
\begin{tabular}{cclllll}
\hline
\textbf{Type}     & \textbf{Task}   & \textbf{Datasets} & \textbf{Class} & \textbf{Length}  &\textbf{ Number}     & \textbf{Related Papers} \\ \hline
Classification & Sentiment Analysis& MR & 2     & 20 & 10662 &  \cite{DBLP:conf/emnlp/Kim14, DBLP:conf/acl/KalchbrennerGB14, DBLP:conf/emnlp/YangZYLZZ18, DBLP:conf/aaai/YaoM019, DBLP:conf/www/WangSH0Z18}  \\ \cline{3-7} 
   &  & SST-1    & 5 & 18 & 11,855 & \cite{DBLP:conf/emnlp/SocherPWCMNP13, DBLP:conf/emnlp/Kim14, DBLP:conf/acl/TaiSM15, DBLP:conf/icml/ZhuSG15, DBLP:conf/emnlp/0001DL16}  \\ \cline{3-7} 
   &  & SST-2    & 2 & 19 & 9,613 & \cite{DBLP:conf/emnlp/SocherPWCMNP13, DBLP:conf/emnlp/Kim14, DBLP:conf/emnlp/LiuQCWH15, DBLP:conf/ijcai/LiuQH16, DBLP:conf/naacl/DevlinCLT19}   \\ \cline{3-7} 
   &  & MPQA     & 2 & 3 & 10,606 & \cite{DBLP:conf/emnlp/SocherPHNM11, DBLP:conf/emnlp/Kim14, DBLP:conf/iclr/ShenZL0Z18}   \\ \cline{3-7} 
   &  & IMDB     & 2 & 294 & 50,000 &  \cite{DBLP:conf/icml/LeM14, DBLP:conf/acl/IyyerMBD15, DBLP:conf/emnlp/LiuQCWH15, DBLP:conf/ijcai/LiuQH16, DBLP:conf/iclr/MiyatoDG17, DBLP:conf/nips/YangDYCSL19}   \\ \cline{2-7} 
   & News Classification& 20NG     &20 & 221 & 18,846 & \cite{DBLP:conf/aaai/LaiXLZ15, DBLP:conf/icml/JohnsonZ16, DBLP:conf/iclr/BaoWCB20, DBLP:conf/aaai/YaoM019, DBLP:conf/icml/WuSZFYW19, DBLP:conf/coling/ZhouQZXBX16} \\ \cline{3-7} 
   &  & AG News  & 4 & 45/7 &127,600 & \cite{DBLP:conf/nips/ZhangZL15, DBLP:conf/acl/JohnsonZ17, DBLP:conf/ijcai/WangWZY17, DBLP:conf/emnlp/YangZYLZZ18, DBLP:conf/nips/YangDYCSL19}      \\ \cline{3-7} 
   &  & R8 &8 & 66&7,674 & \cite{DBLP:conf/aaai/YaoM019, DBLP:conf/icml/WuSZFYW19, DBLP:conf/emnlp/HuangMLZW19}   \\ \cline{3-7} 
   &  & R52& 52 & 70 & 9,100 & \cite{DBLP:conf/aaai/YaoM019, DBLP:conf/icml/WuSZFYW19, DBLP:conf/emnlp/HuangMLZW19}   \\ \cline{2-7} 
   & Topic Labeling& DBPedia  & 14 &55 & 630,000 & \cite{DBLP:conf/nips/ZhangZL15, DBLP:conf/acl/JohnsonZ17, DBLP:conf/iclr/MiyatoDG17, DBLP:conf/cncl/SunQXH19}   \\ \cline{3-7} 
   &  & Ohsumed  & 23 &136 &7,400 &\cite{DBLP:conf/aaai/YaoM019, DBLP:conf/icml/WuSZFYW19, DBLP:conf/emnlp/HuangMLZW19}   \\ \cline{3-7} 
   &  & YahooA   & 10 & 112 & 1,460,000 & \cite{DBLP:conf/nips/ZhangZL15, DBLP:conf/naacl/YangYDHSH16}   \\ \cline{2-7} 
   & Natural Language Inference     & SNLI     & 3&   - &570,152 & \cite{DBLP:conf/emnlp/BowmanAPM15, DBLP:conf/ijcai/WangHF17, DBLP:journals/corr/abs-1907-11692, DBLP:conf/acl/LiuHCG19, DBLP:conf/naacl/DevlinCLT19, DBLP:conf/naacl/PetersNIGCLZ18} \\ \cline{3-7} 
   &  & MNLI     & 3& -   & 433,000&  \cite{DBLP:conf/naacl/WilliamsNB18, DBLP:conf/naacl/DevlinCLT19, DBLP:conf/nips/YangDYCSL19, DBLP:journals/corr/abs-1907-11692, DBLP:conf/iclr/LanCGGSS20} \\ \cline{3-7} 
   &  & QNLI     &2 & -   & 115,667&   \cite{DBLP:conf/naacl/DevlinCLT19, DBLP:conf/nips/YangDYCSL19, DBLP:conf/iclr/LanCGGSS20} \\ \cline{3-7} 
   &  & WNLI     & 2&  -  &852 & \cite{DBLP:conf/acl/LiuHCG19, DBLP:conf/iclr/LanCGGSS20}  \\ \cline{3-7} 
   &  & RTE& 2&  - & 5,768& \cite{DBLP:conf/iclr/LanCGGSS20}   \\ \cline{3-7} 
   &  & SICK     &3 & -   & 10,000&  \cite{DBLP:conf/semeval/MarelliBBBMZ14}  \\ \cline{3-7} 
   &  & MSRP     &2 &  -  & 5,801& \cite{DBLP:conf/coling/DolanQB04}   \\ \cline{2-7} 
   & Named Entity Recognition    &  CoNLL 2003  & 4&  -  &2,302 &   \cite{DBLP:conf/naacl/PetersNIGCLZ18, DBLP:conf/naacl/DevlinCLT19, DBLP:conf/emnlp/FuLN20, DBLP:conf/emnlp/LesterPHCB20, DBLP:conf/emnlp/LuoZZ20, DBLP:conf/acl/LiFMHWL20}  \\ \cline{3-7} 
    & &OntoNotes 4.0&18&-&-&  \cite{DBLP:conf/acl/ZhangY18, DBLP:conf/nips/MengWWLNYLHSL19} \\ \cline{3-7} 
   & &OntoNotes 5.0&18&-&2,945,000&  \cite{DBLP:conf/naacl/DevlinCLT19, DBLP:conf/emnlp/FuLN20, DBLP:conf/emnlp/LesterPHCB20, DBLP:conf/acl/LiFMHWL20}  \\ \cline{3-7} 
   & &MSRA & 3&-&- & \cite{DBLP:conf/acl/ZhangY18, DBLP:conf/naacl/DevlinCLT19, DBLP:conf/nips/MengWWLNYLHSL19, DBLP:conf/acl/LiFMHWL20}  \\ \cline{3-7} 
   & &ACE 2004 &7 &-&443 & \cite{DBLP:conf/naacl/KatiyarC18, DBLP:conf/emnlp/WangL18, DBLP:conf/naacl/LuanWHSOH19, DBLP:journals/tacl/ShibuyaH20, DBLP:conf/acl/LiFMHWL20}  \\ \cline{3-7}
   & &ACE 2005 &7 &-&437 & \cite{DBLP:conf/naacl/KatiyarC18, DBLP:conf/emnlp/WangL18, DBLP:conf/naacl/LuanWHSOH19, DBLP:conf/acl/LinLHS19, DBLP:conf/acl/LiFMHWL20} 
   \\ \cline{3-7} 
   & &KBP2017  & -&-&- & \cite{DBLP:conf/acl/LinLHS19, DBLP:conf/acl/LiFMHWL20}  \\ \cline{2-7} 
   & Question Answering& QQP&2 &    &799,266 &  \cite{DBLP:conf/naacl/DevlinCLT19, DBLP:conf/iclr/LanCGGSS20} \\ \cline{3-7} 
   &  & MRPC     & 2&  -  & -&  \cite{DBLP:conf/iclr/LanCGGSS20}  \\ \cline{3-7} 
   &  & SQuAD    & - & 5,000 &  5,570 & \cite{DBLP:conf/naacl/PetersNIGCLZ18, DBLP:journals/corr/abs-1907-11692, DBLP:conf/iclr/LanCGGSS20}    \\ \cline{3-7} 
   &  & RACE     &5 &  -  & 100,000&   \cite{DBLP:conf/emnlp/LaiXLYH17, DBLP:conf/nips/YangDYCSL19, DBLP:conf/acl/LiuHCG19, DBLP:conf/iclr/LanCGGSS20} \\ \cline{3-7} 
   &  & TREC  &6&10&6,400& \cite{DBLP:conf/acl/KalchbrennerGB14, DBLP:conf/emnlp/LiuQCWH15, DBLP:conf/ijcai/WangWZY17, DBLP:conf/coling/ZhouQZXBX16, DBLP:conf/emnlp/YangZYLZZ18, DBLP:conf/cncl/SunQXH19}  \\ \cline{3-7} 
   &  & WikiQA   & - &  873 &  243 &\cite{DBLP:conf/emnlp/YangYM15, DBLP:journals/corr/SantosTXZ16}   \\ \cline{2-7} 
   & Dialog Act Classification     & DSTC 4   &89 & - &30,000 &\cite{DBLP:conf/naacl/LeeD16, DBLP:conf/iwsds/KimDBWH16}  \\ \cline{3-7} 
   &  & MRDA     & 5 & - &62,000 &\cite{DBLP:conf/icassp/AngLS05, DBLP:conf/naacl/LeeD16}    \\ \cline{3-7} 
   &  & SwDA     & 43 & - &1,022,000 &\cite{DBLP:conf/naacl/LeeD16, DBLP:conf/bigdataconf/WanYGZWY18, DBLP:conf/naacl/RahejaT19}   \\ \cline{1-7} 
 Generation  & Text Summarization &  NYT&- &  -  &109,910 &  \cite{DBLP:conf/acl/XuGCL20, DBLP:conf/emnlp/ZouZLWZ20}  \\ \cline{3-7} 
   &  & CNN&- & 760   & 92,579&  \cite{DBLP:conf/aaai/LiuLYQZL18, DBLP:conf/aaai/YangQTSZ019, DBLP:conf/emnlp/BhandariGALN20, DBLP:conf/emnlp/DongWGCCL20, DBLP:conf/emnlp/HuangCYBWXZ20} \\ \cline{3-7} 
   &  & Dailymail    &- & 653   &219,506 & \cite{DBLP:conf/emnlp/KryscinskiPXS18, DBLP:conf/aaai/YangQTSZ019, DBLP:conf/acl/XuGCL20, DBLP:conf/emnlp/KryscinskiMXS20, DBLP:conf/emnlp/DongWGCCL20}\\ \cline{3-7} 
   &  & Gigaword    &- & -   &3,991,000 & \cite{DBLP:conf/acl/KourisAS19, DBLP:conf/aaai/YangQTSZ019} \\ \cline{2-7} 
   
    & Machine Translation&   WMT14  &- &  -  & -& \cite{DBLP:conf/acl/ChenWUS20, DBLP:conf/emnlp/LinPWQFZL20}\\ \cline{3-7} 
    &  &   WMT16  &- &  -  &- &\cite{DBLP:conf/acl/BugliarelloO20, DBLP:conf/emnlp/LinPWQFZL20} \\ \cline{3-7} 
    &  &   WMT17  &- & -   &- & \cite{DBLP:conf/acl/AjiBHS20, DBLP:conf/emnlp/BaziotisHB20, DBLP:conf/acl/BugliarelloO20, DBLP:conf/acl/ChenWUS20, DBLP:conf/aaai/CuiHLGZHC21}\\ \cline{3-7} 
    &  &   WMT18  &- & -   &- & \cite{DBLP:conf/acl/AjiBHS20, DBLP:conf/acl/BugliarelloO20, DBLP:conf/emnlp/BaziotisHB20} \\ \cline{2-7} 
    &Dialogue  &   DSTC2  &- &  -  & 3,000& \cite{DBLP:conf/emnlp/WuHSX20} \\ \cline{3-7} 
    &  &   MWOZ  &35 &  15.03  & 10,438& \cite{DBLP:conf/emnlp/WuHSX20, DBLP:conf/acl/CampagnaFML20, DBLP:journals/corr/abs-2103-10518}\\ \cline{3-7} 
    &  &   GSIM  &- &  -  &3,008 & \cite{DBLP:conf/emnlp/WuHSX20} \\ \cline{3-7} 
    &  &   OOS  &151 &  -  & 23,700& \cite{DBLP:conf/emnlp/WuHSX20} \\ \cline{1-7} 
\end{tabular}
}
\end{table*}

\paragraph{Sentiment Analysis (SA)}
It consists of judging the emotional polarity and dividing it into several classes.
Depending on the granularity of sentiments, the SA is divided into three categories: dichotomy (positive and negative), trichotomy (positive, negative, and neutral), and multiple categories.
Here we introduce several datasets in detail.

\myparagraph{Stanford sentiment treebank (SST)~\cite{sentiment}} The dataset is an extension of MR~\cite{DBLP:conf/acl/PangL05}. SST-1 is a version of SST. It is divided into five categories and the number of training texts and testing texts is 8,544 and 2,210, respectively. It also consists of 20 average tokens.
The SST-2~\cite{socher2013recursive} contains 9,613 movie reviews including 6,920 training texts, 872 development texts, and 1,821 testing texts.

\myparagraph{Semantic textual similarity benchmark (STS-B)~\cite{cer2017semeval}} It is used in semantic textual similarity tasks organized in the SemEval context between 2012 and 2017~\cite{DBLP:conf/naacl/HendrickxKKNSPP09}.
It consists of text from image titles, news titles and forums.
On a scale of 1 to 5, STS-B displays the semantic similarity of two sentences.
It includes 5,749 training sets, 1,379 development sets, and 1,377 testing sets.

\myparagraph{Multi-Perspective Question Answering (MPQA)~\cite{DBLP:journals/lre/WiebeWC05, mpqa}} This is an opinion dataset which has two categories. 
It contains 10,606 sentences from various news sources that have been manually annotated for opinions and other private states.
It is worth noting that there are 3,311 positive articles and 7,293 negative articles, having no labels for each article.

\myparagraph{IMDB reviews~\cite{DBLP:conf/kdd/DiaoQWSJW14}} The dataset is the world’s most authoritative source for binary sentiment classification of film reviews. The number of content in each class is the same and it can be divided into training and testing sets whose number of comments is 25,000 on average. 


\paragraph{News Classification (NC)}
As one of the most vital information sources, news content exerts a critical effect on people.
The NC facilitates users to acquire essential knowledge in real time. Its applications mainly include news topic identification and recommendation of relevant news based on user interests.
Here we introduce several datasets in detail.

\myparagraph{20 Newsgroups (20NG)~\cite{datasets-for-single-label-textcategorization}} 
20NG is a text dataset derived from newsgroups. There are 20 classes with the same number of articles per class, including 18846 articles in total. The average number of tokens is 221.

\myparagraph{AG News~\cite{DBLP:conf/nips/ZhangZL15, AG-News}} This is an academic news search engine, which is divided into four categories. It contains news headlines and introductions. It includes 120,000 training texts and 7,600 testing texts. The number of average tokens is 45/7.

\myparagraph{R8 and R52~\cite{textmining}} They come from Reuters~\cite{nlp-reuters}. R8 contains 8 classes  consisting of 66 average tokens and includes 2,189 and 5,485 testing and training courses.
 There are 52 classes in R52, which consists of 70 average tokens. It is divided into 6,532 and 2,568 training and testing texts.

\paragraph{Topic Labeling (TL)}
The task mainly obtains the meaning of the file by defining complex file themes.
It is a critical component of topic analysis technology, which aims at simplifying topic analysis by assigning each article to one or more topics. 
Here, we introduce a few in detail.

\myparagraph{DBpedia~\cite{DBLP:journals/semweb/LehmannIJJKMHMK15}} It is a large-scale multilingual knowledge base generated by Wikipedia's most commonly used information boxes. It releases DBpedia every month, adding or removing classes and attributes in each version. The most popular version of DBpedia has 14 categories, separated into 560,000 training data and 70,000 testing data. The number of average tokens is 55.

\myparagraph{Ohsumed~\cite{ohsumed}} This is a biomedical literature database. The number of texts is 7,400. It has 23 cardiovascular disease categories and consists of 136 average tokens. 
All texts are medical abstracts that are categorized into one or more classes.

\myparagraph{Yahoo answers (YahooA)~\cite{DBLP:conf/nips/ZhangZL15}} The dataset is a topic labeling task having 10 categories. The number of average tokens is 136. There are 140,000 training data and 5,000 testing data. Each text in YahooA has question titles, question contexts, and best answers.

\paragraph{Natural Language Inference (NLI)}
This task is used to forecast whether the meaning of a text can be inferred from another. Interpretation is a broad form of NLI. 
By comparing the semantic similarity of sentence pairings, it determines whether a sentence is the interpretation of another one.
Here we introduce several primary datasets in detail.

\myparagraph{The Stanford Natural Language Inference (SNLI)~\cite{DBLP:conf/emnlp/BowmanAPM15}} 
It is commonly used in NLI takes. It contains 570,152 human-annotated sentence pairs, which are annotated with three sorts of relationships: neutral, derived, and conflicting. 
Multi-genre Natural Language Inference (MNLI)~\cite{williams2017broad} has 3 categories and consists of 430,000 sentence pairs annotated with textual information, which is usually used in textual inference tasks. 
Question Natural Language Inference (QNLI)~\cite{rajpurkar2016squad}, whose task with 2 classes is to determine whether a given text pair is a question-answer.
Winograd Natural Language Inference (WNLI)~\cite{levesque2012winograd} which consists of 2 categories is a dataset that captures the standard reference information between two paragraphs.


\myparagraph{Microsoft Research Paraphrase (MSRP)~\cite{DBLP:conf/coling/DolanQB04}} The dataset contains sentence pairs for the text-similarity task, including 1,725 training and 4,076 testing sets. A binary label annotates each pair, discriminating whether they are paraphrases.

\myparagraph{Sentences Involving Compositional Knowledge (SICK)~\cite{DBLP:conf/semeval/MarelliBBBMZ14}} It includes nearly 10,000 English sentence pairs, marked with similarity, and the scale range is 1-5. It has neutral, entailment, and contradictory three categories.

\paragraph{Named Entity Recognition (NER)}
This is a fundamental task of NLP to identify people, places, organizations, and other entities in text.
It is a crucial primary tool for many NLP tasks, including information extraction, question answering, semantic parsing, machine translation, etc.

\myparagraph{CoNLL 2003~\cite{DBLP:conf/naacl/PetersNIGCLZ18}}
It consists of newswire text from the Reuters RCV1
corpus. It contains four different entity types (Location, Organization, Person, and Miscellaneous) and includes 1,393 English news articles, and 909 German news articles.


\myparagraph{OntoNotes 5.0~\cite{DBLP:conf/naacl/DevlinCLT19}}
The dataset consists of 174,5K English, 900K Chinese, and 300K Arabic text data. It comes from telephone conversations, news agencies, radio news, radio conversations, and blogs. It has 18 entity classes containing 11 types, seven values, and 2,945,000 text data.

\myparagraph{MSRA~\cite{DBLP:conf/acl/ZhangY18}}
This is a Chinese dataset that is obtained from the news domain. It has three types of entities and is used as a shared task on SIGNAN back in 2006.



\paragraph{Question Answering (QA)}
There are two types of QA systems: the extraction guidance system and the generation guidance system. 
The extractive QA can be regarded as a particular case of text classification. 
Here we detail several datasets.

\myparagraph{Microsoft Research Paraphrase Corpus (MRPC)~\cite{dolan2005automatically}} It contains 5,800 sentence pairs extracted from Internet news, and the task type is similar to the QQP dataset. Sentence pairs are derived from comments on the same news item and determine whether the two sentences are semantically the same. The assessment criteria were classification accuracy and F1 score.

\myparagraph{Stanford Question Answering Dataset (SQuAD)~\cite{DBLP:conf/naacl/PetersNIGCLZ18} } This is a large-scale machine-reading comprehension dataset that contains two tasks.
SQuAD 1.1~\cite{rajpurkar2016squad} provides questions and corresponding answers, and the dataset contains 100,000 samples in total, while SQuAD 2.0~\cite{rajpurkar2018know} adds unanswered questions and expands the scale to 150,000.

\myparagraph{RACE~\cite{lai2017race}} The dataset has 5 categories, containing nearly 100,000 questions extracted from middle and high school English tests, with corresponding answers given by experts.
The average length of RACE text is more significant than 300, which is longer than other reading comprehension datasets (such as SQuAD) sequences.



\paragraph{Dialog Act Classification (DAC)}
The dialogue act is a specific verbal component, which marks the dialogue according to the meaning category of the dialogue. DAC categorizes tags according to the meaning of the dialogue to help understand the speaker's intentions.

\myparagraph{Dialog State Tracking Challenge 4 (DSTC 4)~\cite{DBLP:conf/iwsds/KimDBWH16}}
It belongs to the dialog act classification task and mainly focuses on dialog state tracking on human-human dialogs. It is divided into 89 training classes and contains 24,000 training texts and 6,000 test texts.

\myparagraph{ICSI Meeting Recorder Dialog Act (MRDA)~\cite{DBLP:conf/icassp/AngLS05}}
It includes about 75 hours of speech from 75 naturally occurring meetings among 53 speakers. The number of categories is 5, and it contains 51,000 training texts, 11,000 test texts, and 11,000 validation texts.

\myparagraph{Switchboard Dialog Act (SwDA)~\cite{manualarticle}}
The dataset extends the dialogue behavior label with rounds/discourses. The label summarizes the sentence structure, and relevant and pragmatic information of the relevant turn. The SwDA is split into 43 training classes and includes 1,003,000 training texts, 19,000 test texts, and 112,000 validation texts.

\paragraph{Text Summarization}
Text summarization is a summary of given single or multiple documents. It is kept as concise as possible while ensuring that it reflects the critical content of the original document.
It can be divided into extractive summarization and generative summarization.
Extractive summarization is generated by extracting and splicing the critical sentences in documents. Generative summarization is generated by a model, which summarizes documents according to the required content expressed in documents.

\myparagraph{NYT~\cite{DBLP:conf/acl/XuGCL20}}
The dataset comes from the corpus annotated by the New York Time. The named entities are annotated using the Stanford NER tool in conjunction with the Freebase knowledge base. It contains 9,076 articles, with the remaining 100,834 divided into a training set (96,834 examples) and a validation set (4,000 samples).

\myparagraph{CNN/Daily Mail~\cite{DBLP:conf/aaai/LiuLYQZL18}}
It is used for the passage-based question-answering task, and it is popular in assessing ATS systems.
The dataset consists of CNN/Daily Mail news stories paired with multi-sentence human-generated summaries. There are 287,226 training instances, 13,368 validation instances, and 11,490 testing instances in total.

\myparagraph{Gigaword~\cite{DBLP:conf/acl/KourisAS19}}
This is a dataset of English news chapters consisting of nearly 950 pieces.
Headlines -- stories from multiple sources, including the New York Times -- include some articles with a one-sentence, short news feed.

\paragraph{Machine Translation (MT)}
It refers to the task of translation from one language to another with its semantic equivalence by a computer.
There are three categories, rule-based machine translation, statistics-based machine translation, and neural network-based machine translation.

\myparagraph{WMT14~\cite{DBLP:conf/acl/ChenWUS20}}
It is a grouping of datasets used in the Ninth Workshop on Statistical Machine Translation shared tasks, including a news translation task, a quality estimation task, a metrics task, and a medical text translation task.

\myparagraph{WMT16~\cite{DBLP:conf/emnlp/LinPWQFZL20}}
This dataset is a grouping of datasets used in the First Conference on Machine Translation shared tasks. It has ten shared tasks, including a news translation task, an IT domain translation task, a biomedical translation task, an automatic post-editing task, a metrics task, a quality estimation task, a tuning task, a pronoun translation task, a bilingual document alignment task, and a multimodal translation task.

\myparagraph{WMT17~\cite{DBLP:conf/acl/ChenWUS20}}
The dataset includes three MT tasks (news, biomedical, and multimodal), an automatic post-editing task, a quality estimation task, a task dedicated to the training of neural MT systems, a task on bandit learning for MT, an automatic post-editing task, and a metrics task.

\myparagraph{WMT18~\cite{DBLP:conf/acl/AjiBHS20}}
It mainly features six shared tasks: a news translation task, a biomedical translation task, an automatic post-editing task, a metrics task, a quality estimation task, and a multimodal translation task. Participants must evaluate their approaches to the machine translation topic using the standard datasets created for the shared tasks.

\paragraph{Dialogue}
As an essential way of man-machine interaction, 
the dialogue system offers a wide range of applications.
The existing dialogue systems can be grouped into task-oriented dialogue systems and non-task-oriented dialogue systems from application scenarios. Among them, the non-task type of conversation system can also be called a chatbot.


\myparagraph{DSTC2~\cite{DBLP:conf/emnlp/WuHSX20}}
This is a multi-round dialogue dataset of restaurant reservation fields, including 1,612 training data, 506 verification data, and 1,117 test data.
It allows the user's goals to change compared to DSTC1. DSTC2 is also richer in terms of the conversation state representation, including the slot value pairs of the user's targets and the ways to find them. 

\myparagraph{MWOZ~\cite{DBLP:conf/emnlp/WuHSX20}}
It contains 8,420/1,000/1,000 conversations for training, validation, and test sets, respectively. It contains 30 pairs in seven domains being a multi-domain fully-labeled corpus. Every sample includes a goal, multiple user and agent utterances, and annotations regarding slot values.


\myparagraph{Out-Of-Scope (OOS)~\cite{DBLP:conf/emnlp/WuHSX20}}
The dataset includes 15,100 training, 3,100 validation, and 5,500 test sets, respectively. It contains 151 intent classes, containing 150 in-scope and one out-of-scope intent. The out-of-scope intent indicates that a user utterance failed to classify to given predefined objectives. 






\subsection{Downstream Tasks and Datasets on CV}

\begin{table*}[htbp]
\centering
\caption{The statistics of the datasets used on downstream tasks.}\label{datasets2}
\resizebox{\textwidth}{!}{
\begin{tabular}{ccccccl}
\hline
\textbf{Type} & \textbf{Name} & \textbf{Usage} & \textbf{Domain} & \textbf{Class} & \textbf{Size} & \textbf{Related Papers} \\ \hline
    \multirow{2}{*}{Classification} & \multirow{2}{*}{ImageNet} & Pretrain \& & \multirow{2}{*}{-} & \multirow{2}{*}{1000+} & \multirow{2}{*}{1,200,000+} & \multirow{2}{*}{ \begin{tabular}[c]{@{}l@{}}\cite{doersch2015unsupervised, pathak2016context, noroozi2016unsupervised, kim2018learning, zhang2017split, noroozi2017representation, bojanowski2017unsupervised, zhang2016colorful, oord2018representation,li2020prototypical, yan2020clusterfit, tian2020makes, bachman2019learning, caron2021emerging, xie2021self, DBLP:conf/icml/TianCG21, henaff2020data, wu2018unsupervised}\\ \cite{ zhuang2019local, misra2020self, he2020momentum, chen2020improved, grill2020bootstrap, donahue2016adversarial, donahue2019large, caron2018deep, zhang2019aet, chen2020simple, caron2020unsupervised, chen2021exploring, asano2019self, tian2019contrastive, chen2020big, chen2021empirical, DBLP:conf/iclr/MitrovicMWBB21, he2021masked}\end{tabular}} \\
    & & Downstream & & & &
\\ \cline{2-7} 
    & CIFAR-10 & Downstream & - & 10 & 60,000 & 
\cite{dosovitskiy2014discriminative, DFB16, zhang2016colorful, grill2020bootstrap, zhang2019aet, asano2019self, chen2020simple, bachman2019learning, DBLP:conf/icml/TianCG21}   
\\ \cline{2-7} 
    & CIFAR-100 & Downstream & - & 100 & 60,000 &   
\cite{grill2020bootstrap, asano2019self, chen2020simple, bachman2019learning}
\\ \cline{2-7} 
    & STL-10 & Downstream & - & 10 & 6,000 &
\cite{dosovitskiy2014discriminative, DFB16, tian2019contrastive, tian2020makes, bachman2019learning, DBLP:conf/icml/TianCG21}
\\ \cline{2-7} 
    & Caltech-101 & Downstream & object & 101 & 9,146 & 
\cite{dosovitskiy2014discriminative, DFB16, grill2020bootstrap, chen2020simple} 
\\ \cline{2-7} 
    & MNIST-10 & Downstream & digit & 10 & 60,000 & 
\cite{donahue2016adversarial, tian2020makes}  
\\ \cline{2-7}
    & SVHN & Downstream & digit & 10 & 73,257 & 
\cite{asano2019self}
\\ \cline{2-7} 
 & Places205 & Downstream & scene & 205 & 2,448,873 &
\cite{zhang2016colorful, zhang2017split, noroozi2017representation, zhuang2019local, misra2020self, caron2018deep, zhang2019aet, li2020prototypical, asano2019self, caron2020unsupervised, bachman2019learning, yan2020clusterfit, he2021masked}
\\ \cline{2-7} 
    & SUN397 & Downstream & scene & 899 & 130,519 &
\cite{chen2020simple}
\\ \cline{2-7} 
    & HMDB51 & Downstream & action & 51 & 7000 & 
\cite{tian2019contrastive}
\\ \cline{2-7} 
    & UCF101 & Downstream & action & 101 & - & 
\cite{tian2019contrastive}  
\\ \cline{2-7} 
    & Food-101 & Downstream & food & 101 & 101,000 &   
\cite{grill2020bootstrap, chen2020simple}
\\ \cline{2-7} 
    & Birdsnap & Downstream & bird & 500 & 49,829 &
\cite{chen2020simple}
\\ \cline{2-7} 
    & Cars & Downstream & car & 196 & 16,185 &
\cite{chen2020simple, grill2020bootstrap}
\\ \cline{2-7} 
    & Aircraft & Downstream & aircraft & 102 & 10,200 &
\cite{grill2020bootstrap, chen2020simple}
\\ \cline{2-7} 
    & Pets & Downstream & pet & 37 & 7,400 &
\cite{grill2020bootstrap, chen2020simple}
\\ \cline{2-7} 
    & Flowers & Downstream & flower & 102 & 8,189 &
\cite{grill2020bootstrap, chen2020simple}
\\ \cline{2-7} 
    & DTD & Downstream & texture & 47 & 5,640 &
\cite{grill2020bootstrap, chen2020simple}
\\ \cline{2-7} 
    & iNaturallist2018 & Downstream & species & 8,000+ & 450,000+ &
\cite{misra2020self, caron2020unsupervised, yan2020clusterfit, he2021masked}
\\ \cline{2-7} 
    & JFT-300M & Pretrain & - & 3,000+ & 300,000,000+ &
\cite{dosovitskiy2020image, he2021masked}
\\ \cline{1-7} 
Detection & COCO & Downstream & object & 80 & 200,000 & 
\cite{noroozi2017representation, he2020momentum, chen2020improved, tian2020makes, caron2020unsupervised, chen2021exploring, xie2021self, he2021masked}  
\\ \cline{2-7}
    & VOC07 & Downstream & object & 20 & 9,963 & 
\cite{pathak2016context, zhang2016colorful, noroozi2016unsupervised, zhang2017split, noroozi2017representation, bojanowski2017unsupervised, zhang2016colorful, henaff2020data, zhuang2019local, misra2020self, he2020momentum, chen2020improved, grill2020bootstrap, caron2018deep, li2020prototypical, asano2019self, caron2020unsupervised, chen2021exploring, kim2018learning} 
\\ \cline{1-7}
Segmentation & VOC12 & Downstream & object & 20 & 2,913 & 
\cite{pathak2016context, zhang2016colorful, noroozi2016unsupervised, zhang2017split, noroozi2017representation, caron2018deep, kim2018learning}
\\ \cline{2-7} 
    & NYU-Depth V2 & Downstream & scene & 894 & 1,449 & 
\cite{zhang2017split, grill2020bootstrap, tian2019contrastive}
\\ \cline{2-7} 
    & VOC11 & Downstream & object & 20 & 3,334 & 
\cite{doersch2015unsupervised}
\\ \cline{2-7} 
    & ADE20K & Downstream & scene & 3,688 & 27,574 & 
\cite{xie2021self, he2021masked}
\\ \cline{2-7} 
    & Cityscapes & Downstream & scene & 25 & 25,000+ & 
\cite{he2020momentum}
\\ \cline{2-7} 
    & LVIS & Downstream & vocabulary & 1,200+ & 160,000+ & 
\cite{he2020momentum}
\\ \cline{2-7} 
    & DAVIS & Downstream & scene & 150 & - & 
\cite{caron2021emerging}
\\ \cline{1-7} 
Inpainting & Paris StreetView & Downstream & scene & - & 15,000 & 
\cite{doersch2015unsupervised, pathak2016context}
\\ \cline{1-7}
Sequence & Moving-MNIST & Downstream & digit & - & 10,000 & 
\cite{tian2020makes}
\\ \cline{1-7}
- & YFCC100M & Pretrain & multimedia & - & 100,000,0000+ & 
\cite{caron2018deep}
\\ \cline{1-7}
\end{tabular}
}
\end{table*}

The datasets in CV mainly contain three types from the perspective of tasks: classification, detection, and segmentation. The popular datasets are concluded in Table \ref{datasets2}, and some infrequently mentioned datasets in long tails are discussed in the text.

\paragraph{Classification} In this part, we first cover the popular large-scale datasets used frequently in both the pretext and downstream tasks. Then the domain datasets only used for the downstream tasks are unfolded.

\myparagraph{MNIST~\cite{mnist}} 
It's a collection of handwritten digits that includes $60,000$ samples in training and $10,000$ in testing.
The images are fixed-size with $28\times28$ pixels. The pixel values are from $0$ to $255.0$ in which pixel values smaller than 255.0 can be understood as background (white) and 255 means foreground (black). The labels are from 0 to 9 and only one of these digits exists in an image. Both traditional and deep learning methods are based on this most popular dataset despite advanced methods showing perfect results. Thus, Geoffrey Hinton has described it as "the drosophila of machine learning".

\myparagraph{Street View House Numbers (SVHN)~\cite{svhn}} In the domain of digit numbers, it collects real-world digit numbers from house numbers in Google Street View images. It includes $73,257$ digits for training, $26,032$ digits for testing, and $531,131$ additional. All of them are $32\times32$ color images with both class labels and character-level bounding boxes.

\myparagraph{CIFAR}~\cite{cifar} As more advanced methods show perfect results on the simple datasets, more sophisticated datasets such as CIFAR-10 and CIFAR-100 are conducted. These two datasets are closer to the real-world object. The CIFAR-10 contains $50,000$ training images and $10,000$ testing images, with $6,000$ images per class and $32\times32$ pixels in each RGB color image. The CIFAR-100 is similar to the CIFAR-10 but with more detailed label information. There are $100$ classes containing $500$ training images and $100$ testing images in each class. In addition, these $100$ "fine" classes are grouped equally into $20$ "coarse" classes. Researchers can adapt it to suitable learning methods.

\myparagraph{STL-10~\cite{coates2011analysis}} Inspired by the CIFAR-10 dataset, STL-10 is another $96\times96$ color image dataset containing similar $10$ real-world classes. Each class has $500$ training images and $800$ testing images. The biggest difference is that STL-10 has $100,000$ unlabeled images for unsupervised learning. More construction information can be seen in~\cite{stl-10}.

\myparagraph{Caltech-101~\cite{caltech-101}} It collects roughly $300\times200$ color images of objects belonging to 101 categories, with 40 to 800 images per category and 50 on average. The outlines of the objects in the pictures are annotated for the convenience of different learning methods.

\myparagraph{ImageNet~\cite{deng2009imagenet}} This is one of the most popular and large-scale datasets on computer vision. It is built according to the hierarchical structure of WordNet~\cite{miller1998wordnet}. The full ImageNet dataset contains $14,197,122$ images and $21,841$ synsets indexed, attaching on average $1,000$ images to demonstrate each synset. The most frequently-used subset of ImageNet is the ImageNet Large Scale Visual Recognition Challenge (ILSVRC) dataset from 2010 to 2017, containing tasks of classification, localization, and detection. The number of samples in training and testing datasets and the labels of images are determined by the specific task, more details are seen in~\cite{imagenet}.

\myparagraph{HMDB51~\cite{hmdb, Kuehne11}} In addition to the popular MNIST, there still exist many domain datasets used for the downstream tasks in the classification problem. HMDB51 is an action video database for a total of $7,000$ clips in 51 action classes. It contains five types of facial actions and body movements.

\myparagraph{UCF101~\cite{ucf101}} It is another action video dataset designed for more realistic action recognition. It is an extension of the UCF50~\cite{ucf50} dataset containing only 50 action categories with 101 action categories, collected from YouTube. What makes it a famous recognition dataset is the workshop in ICCV13 with UCF101 as its main competition benchmark. 

\myparagraph{Food-101~\cite{bossard2014food}} This is a real-world food dataset of $101$ food categories, with $750$ and $250$ images per class in training and testing dataset respectively.

\myparagraph{Birdsnap~\cite{Berg_2014_CVPR}} 
It is a fine-grained visual categorization of birds on a broad scale,
with bounding boxes and the locations/annotations of 17 parts in the object. It contains $49,829$ images of the 500 most common species in North America, with each species containing $69$ to $100$ images and most species having 100. In addition, some images are also labeled as male or female, immature or adult, and breeding or non-breeding plumage. 

\myparagraph{SUN397} To target the scene categorization, the extensive Scene UNderstanding (SUN) database~\cite{xiao2010sun, xiao2016sun} fills the gap of the existing dataset with the limited scope of categories. This database has $899$ categories and $130,519$ images, and only images with more than $200\times200$ pixels were kept. SUN397 is a more well-sampled subset that maintains 397 categories with at least 100 images per category, in which other categories containing relatively few unique photographs are discarded. 

\myparagraph{Places205} Places205~\cite{places205} dataset is another large scale scene dataset consists of $2,448,873$ images from 205 scene categories.

\myparagraph{Cars~\cite{cars}} The dataset in the domain of cars contains $16,185$ color images of $196$ classes (at the level of Make, Model, Year) of cars. For convenience, this dataset is split into training and testing sets in roughly equal quantities. 

\myparagraph{Aircraft~\cite{maji13fine-grained}} It is another fine-grained visual classification designed for aircraft (also known as FGVC-Aircraft). A popular form of this dataset is the fine-grained recognition challenge 2013 (FGComp2013)~\cite{fgcomp2013} ran in parallel with the ILSVRC2013. There exist four-level hierarchies: Model, Variant, Family, Manufacturer, from finer to coarser to organize this database. The more detailed information is shown in~\cite{fgvc-aircraft}. 

\myparagraph{Pets~\cite{pets}} It represents The Oxford-IIIT Pet Dataset that collects 37 pet categories with roughly 200 images per category. All images have an associated ground truth annotation of breed for classification, head ROI for detection, and pixel-level trimap for segmentation. 

\myparagraph{Flowers~\cite{flowers}} Similarly, Flowers is another domain dataset in flowers also collected by Oxford; it contains Oxford-17 Flowers of 17 categories and Oxford-102 Flowers of 102 categories. 

\myparagraph{Describable Textures Dataset (DTD)~\cite{dtd}} This is an evolving collection of textural images in the wild, which consists of $5,640$ images of 47 categories, with 120 images per category.

\myparagraph{iNaturalist2018~\cite{inaturalist2018}} It is a large-scale species classification competition conducted on the FGVC5 workshop at CVPR2018. This dataset contains over 8,000 species categories, with more than $450,000$ images in the training and validation dataset collected from iNaturalist~\cite{inaturalist}.

\myparagraph{JFT-300M~\cite{sun2017revisiting}} JFT-300M is an internal Google dataset introduced by Sun et al~\cite{sun2017revisiting} and well-known from ViT Model~\cite{dosovitskiy2020image}. It is labeled by algorithms that utilize human-computer communications and target classification tasks. This dataset finally contains 300M images with over 1000M labels, thus leading to the multiple labels attached to this large-scale dataset.

\paragraph{Detection} The detection is a popular task in the CV, and almost all the research is conducted on COCO and PASCAL VOC datasets.

\myparagraph{COCO~\cite{lin2014microsoft}} This is a large-scale dataset for object detection, segmentation, and caption; it contains $330,000$ RGB images, with more than $200,000$ labeled. There are 1.5 million object instances of 80 object categories involved. Thus, it is one of the most popular benchmark dataset in detection and segmentation in parallel with the following PASCAL VOC.

\myparagraph{PASCAL VOC~\cite{voc}} 
From 2005 through 2012, the dataset has run challenges assessing performance on object class recognition and has provided standardized image datasets for object class recognition.
The main datasets used in self-supervised learning are VOC07, VOC11, and VOC12. Main competitions in VOC07~\cite{voc07} contain classification and detection tasks; both of them consist of 20 objects and contain at least one object in each image. Thus, it is common to use VOC07 to serve as the downstream task for the detection.

\paragraph{Segmentation} The segmentation is a semantics-based pixel-level classification. These datasets are difficult to obtain and annotate, thus they are always used as a downstream task.

\myparagraph{VOC11~\cite{voc11} \& VOC12~\cite{voc12}} Both VOC11 and VOC12 contains classification, detection, and segmentation tasks in the main competition, thus leading to the common use of downstream task for the segmentation.

\myparagraph{ADE20K~\cite{zhou2017scene, zhou2019semantic}} It collects $27,574$ images from both the SUN and Places205 databases, in which $25,574$ for training and $2,000$ for testing. All $707,868$ objects from $3,688$ categories existing in images are annotated. Especially, this dataset contains $193,238$ annotated object parts and parts of parts, and additional attributes, annotation time, depth ordering for the benefit of the research community.

\myparagraph{NYU-Depth V2~\cite{nyu_depth_v2}} This is a dataset consisting of images and video sequences from 464 indoor scenes that are recorded by both the RGB and Depth cameras from 3 cities. It contains $1,449$ images with the ground truth of depth, and the original RGB values are also provided. In addition, there are $407,024$ new unlabeled frames and additional class labels for the objects in images.

\myparagraph{Cityscapes~\cite{Cordts2015Cvprw, Cordts2016Cityscapes}} It is a dataset of urban street scenes from 50 cities with the ground truth of semantic segmentation. The main instances are vehicles, people, and construction. The high-quality dense pixel annotations contain a volume of $5,000$ images. In addition to the fine annotations, coarser polygonal annotations are provided for a set of $20,000$ images. Moreover, the videos consist of not consistent images with high-quality annotations, and these annotated images with consistently changing views are provided for researchers.

\myparagraph{LVIS~\cite{gupta2019lvis}} It is a dataset for large vocabulary instance segmentation. It features that 1) a category or word in one image is related to the only segmentation object; 2) more than $1,200$ categories are extracted from roughly $160,000$ images; 3) long tails phenomenon exist in these categories; and 4) more than $2,000,000$ high-quality instance segmentation masks.

\myparagraph{Densely Annotated VIdeo Segmentation (DAVIS)~\cite{davis}} It is a video dataset designed for the in-depth analysis of the SOTA in video object segmentation, in which DAVIS 2017~\cite{davis2017} contains both semi-supervised (human-guided at the testing time) and unsupervised (human non-guided at test time) video sequences with multiple annotated instances.

\paragraph{Others} There are many datasets designed for special visual tasks such as inpainting. In addition, this part covers the data collection in the wild.
%

\myparagraph{Paris StreetView~\cite{doerschdata}} The dataset is designed for image inpainting task, which contains $14,900$ training images and 100 testing images collected from street views of Paris. This dataset is collected from Google Street View and mainly focuses on the buildings in the city. 

\myparagraph{Moving-MNIST~\cite{moving-mnist}} Based on MNIST, it is a video dataset designed for evaluating sequence prediction or reconstruction, which contains $10,000$ sequences. Each video is long of 20 frames and consisted of two digits (possibly overlapped) moving inside a $64\times64$ patch. The first benchmark is reported on~\cite{srivastava2015unsupervised} by the method of LSTMs.

\myparagraph{Yahoo Flickr Creative Commons 100 Million (YFCC100M)~\cite{thomee2016yfcc100m, yfcc100m}} The dataset is the largest public multimedia collection that is allowed to search by users for their own targets; this dataset can browse both images and videos. It is free and for researchers to explore and investigate subsets of the YFCC100M in real time. Subsets of the complete dataset can be retrieved by any keyword search and reviewed directly. In addition, the text information attached to any image or video is abundant, such as containing location information and user tags. Briefly, it is more a multimedia library than a domain dataset.

\myparagraph{Data in the Wild} More generalized dataset concept in the self-supervised learning era is composed of multimedia websites, APP, or search engines such as Instagram, Flickr, Google Images, etc. I think pictures in the wild will play a major role in the future study of CV because of the quantity of data, the computation source, and the learning power of PFM.

\subsection{Downstream Tasks and Datasets on Graph}
The purpose of the pretraining graph model is to improve the performance of downstream tasks.
According to the different analysis objects of the downstream tasks, they can be divided into nodes, edges, and graphs.
Meanwhile, the PFMs of GL have been widely used in a mass of fields. 
In this section, we combine the downstream tasks to conduct statistics on the pretraining datasets and the downstream task datasets.

\paragraph{Node-Level Tasks}
Nodes are the most basic element of the graph, so lots of downstream tasks mainly focus on the analysis of nodes.

\myparagraph{Node Classification}
Node ClassiFication (NCF) is one of the most prevalent graph-based tasks, which has important analytical value in most of the different types of graph data.
Different from the pseudo-labels assigned to nodes in the graph in self-supervised methods, the labels in NCF often come from external information such as manual annotation.
Based on Definition~\ref{def: ind_learning} and~\ref{def: trans_learning}, NCF can be divided into two types: transductive and inductive according to the visibility during training, verification, and testing.
In addition, the result of NCF can be single-label or multi-label according to the mutual exclusion of labels.
The statistical results of common NFC datasets are shown in Table~\ref{tab:dataset of nodeL}.

\begin{table*}[htbp]
	\tiny
	\centering
	\caption{The statistics of the datasets for node-level tasks. Homogeneous:Hom, Heterogeneous:Het.}
	\label{tab:dataset of nodeL}
	\resizebox{\textwidth}{!}{
	\begin{tabular}{cccccccccl}  
		\hline 
	\textbf{Task}	& \textbf{Name} & \textbf{Usage} & \textbf{Source} & \textbf{Type} & \textbf{Nodes} & \textbf{Edges} & \textbf{Class} & \textbf{Features} & \textbf{Related Paper}  \\
		\hline
	NCF	&Academia & pretrain & Citation & Hom & 138K & 739K & - & - & \cite{gcc2020kdd} \\\cline{2-10} 
		&DBLP (SNAP) & pretrain & Citation & Hom & 317K & 2M & - & - & \cite{gcc2020kdd} \\\cline{2-10} 
	&	DBLP (NetRep) & pretrain & Citation & Hom & 540K & 30M & - & - & \cite{gcc2020kdd} \\\cline{2-10}
	&	IMDB & pretrain & Movie & Hom & 896K & 8M & - & - & \cite{gcc2020kdd} \\\cline{2-10}
	&	Facebook & pretrain & Social & Hom & 3M & 47M & - & - & \cite{gcc2020kdd} \\\cline{2-10}
	&	LiveJournal & pretrain & Social & Hom & 4M & 86M & - & - & \cite{gcc2020kdd} \\\cline{2-10}
	&	\multirow{2}{*}{Cora} & \multirow{2}{*}{Downstream} & \multirow{2}{*}{Citation} & \multirow{2}{*}{Hom} & \multirow{2}{*}{2,708} & \multirow{2}{*}{5,429} & \multirow{2}{*}{7} & \multirow{2}{*}{1,433} &  \cite{M3S2020aaai,you2020icml,graphsage2017nips,vgae2016nips,supergat2021iclr}  \\
    &	 &  &  &  &  &  &  &  & \cite{dgi2019iclr,cmvrl2020icml,subcom2020icdm,gic2021pakdd,cg3aaai2021,gmi2020www} \\\cline{2-10}
	&	\multirow{2}{*}{CiteSeer} & \multirow{2}{*}{Downstream} & \multirow{2}{*}{Citation} & \multirow{2}{*}{Hom} & \multirow{2}{*}{3,327} & \multirow{2}{*}{4,732} & \multirow{2}{*}{6} & \multirow{2}{*}{3,703} &  \cite{M3S2020aaai,you2020icml,vgae2016nips,supergat2021iclr}  \\
	&	&  &  &  &  &  &  &  & \cite{cmvrl2020icml,subcom2020icdm,gic2021pakdd,cg3aaai2021,gmi2020www} \\\cline{2-10}
	&	\multirow{2}{*}{PubMed} & \multirow{2}{*}{Downstream} & \multirow{2}{*}{Citation} & \multirow{2}{*}{Hom} & \multirow{2}{*}{19K} & \multirow{2}{*}{44K} & \multirow{2}{*}{3} & \multirow{2}{*}{500} &  \cite{M3S2020aaai,cmvrl2020icml,subcom2020icdm,gic2021pakdd, you2020icml}  \\
	&	&  &  &  &  &  &  &  & \cite{cg3aaai2021,gmi2020www,vgae2016nips,supergat2021iclr,dgi2019iclr} \\\cline{2-10}
	&	ACM & Downstream & Citation & Hom &  8,994 & 26K & 4 & 1,902 & \cite{selar2020nips} \\\cline{2-10}
	&	Cora-Full & Downstream & Citation & Hom & 20K & 63K & 70 & 500 & \cite{supergat2021iclr,autossl2021arxiv} \\\cline{2-10}
	&	Cora-ML & Downstream & Citation & Hom & 2,995 & 8,158 & 7 & 2879 & \cite{supergat2021iclr} \\\cline{2-10}
	&	Reddit-233K & Downstream & Social & Hom & 233K & 57M & 210 & 5,414 & \cite{jin2020arxiv,graphsage2017nips,cmvrl2020icml,subcom2020icdm} \\\cline{2-10}
	&	BlogCatalog & Downstream & Social & Hom & 10K & 334K & 39 & - & \cite{deepwalk2014kdd,node2vec2016kdd} \\\cline{2-10}
	&	YouTube & Downstream & Social & Hom & 1M & 3M & 47 & - & \cite{deepwalk2014kdd} \\\cline{2-10}
	&	Reddit-231K & Downstream & Social & Hom & 231K & 11M & 41 & 602 & \cite{s2grl2020arxiv,grace2020arxiv,dgi2019iclr,gmi2020www} \\\cline{2-10}
	&	Amazon & Downstream & Social & Het & 130M & - & - & - &  \cite{gptgnn2020kdd} \\\cline{2-10}
	&	PPI-30K & Downstream & Protein & Het & 3,890 & 77K & 50 & - & \cite{node2vec2016kdd,dgi2019iclr} \\\cline{2-10}
	&	PPI-57K & Downstream & Protein & Het & 57K & 819K & 121 & 50 & \cite{s2grl2020arxiv,supergat2021iclr,grace2020arxiv,subcom2020icdm,gmi2020www} \\\cline{2-10}
	&	IMDB & Downstream & Movie & Hom & 12K & 37K & 4 & 1,256 & \cite{selar2020nips} \\\cline{2-10}
	&	Four-Univ & Downstream & Movie & Hom & 4,518 & 3,426 & 6 & 2,000 & \cite{supergat2021iclr} \\\cline{2-10}
	&	Chameleon & Downstream & Web & Hom & 2,277 & 36K & 6 & 500 & \cite{supergat2021iclr} \\\cline{2-10}
	&	Crocodile & Downstream & Web & Hom & 12K & 180K & 6 & 500 & \cite{supergat2021iclr} \\\cline{2-10}
	&	Flickr-89K & Downstream & Web & Hom & 89K & 450K & 7 & 500 & \cite{supergat2021iclr,subcom2020icdm} \\\cline{2-10}
	&	ogbn-arxiv & Downstream & Web & Hom & 169K & 117K & 40 & 128 & \cite{supergat2021iclr} \\\cline{2-10}
	&	Wiki-CS & Downstream & Web & Hom & 12K & 277K & 10 & 300 & \cite{supergat2021iclr,autossl2021arxiv} \\\cline{2-10}
	&	DBLP & Downstream & Web & Hom & 17K & 53K & 4 & 1639 & \cite{supergat2021iclr,grace2020arxiv} \\\cline{2-10}
	&	Computers & Downstream & Co-purchase & Hom & 14K & 246K & 10 & 767 & \cite{supergat2021iclr,gic2021pakdd,cg3aaai2021,autossl2021arxiv} \\\cline{2-10}
	&	Photo & Downstream & Co-purchase & Hom & 7,650 & 119K & 8 & 745 & \cite{supergat2021iclr,gic2021pakdd,cg3aaai2021,autossl2021arxiv,merit2021arxiv} \\\cline{2-10}
	&	CS & Downstream & Co-author & Hom & 18K & 82K & 15 & 500 & \cite{supergat2021iclr,gic2021pakdd,cg3aaai2021,autossl2021arxiv,merit2021arxiv} \\\cline{2-10}
	&	Physics & Downstream & Co-author & Hom & 35K & 248K & 5 & 500 & \cite{supergat2021iclr,gic2021pakdd,autossl2021arxiv} \\\cline{2-10}
	&	H-index & Downstream & Co-author & Hom & 5,000 & 44K & - & - & \cite{gcc2020kdd} \\\cline{2-10}
	&	Flickr-81K & Downstream & Photo & Hom & 81K & 6M & 195 & - & \cite{deepwalk2014kdd} \\\cline{2-10}
	&	Wikipedia & Downstream & Word & Hom & 4,777 & 185K & 40 & - & \cite{node2vec2016kdd} \\\cline{2-10}
	&	US-Airport & Downstream & Airline & Hom & 1,190 & 13K & - & - & \cite{gcc2020kdd} \\\cline{2-10}
	&	OAG & Downstream & Academic & Het & 178M & 2B & - & - & \cite{gptgnn2020kdd} \\
		\hline

NTKS	&	KDD-ICDM & Downstream & Co-author & Hom  & 2,867/2,607 & 7,637/4,774 & 697 & - & \cite{gcc2020kdd} \\\cline{2-10}
&		SIGIR-CIKM & Downstream & Co-author & Hom & 2,851/3,548 & 6,354/7,076 & 874 & - & \cite{gcc2020kdd} \\\cline{2-10}
	&	SIGMOD-ICDE & Downstream & Co-author & Hom & 2,626/2,559 & 8,304/6,668 & 898 & - & \cite{gcc2020kdd} \\
		\hline
	\end{tabular}
	}
\end{table*}

\myparagraph{Node Clustering}
The goal of Node ClusterIng (NCI) is to divide a graph into different classes or clusters according to a certain standard so that the correlation of nodes in the same cluster is as large as possible, and the irrelevance of nodes that are not in the same cluster is also minimized.
Although in the above-mentioned pretraining tasks, NCI is used as a pretext task has appeared, NCI can still test pretraining graph models based on other pretext tasks.

\myparagraph{Top-K Search}
The goal of task Top-K Search (TKS) is to search the K nodes with the highest predefined associations for a given node in the graph.
Usually, TKS is used for search tasks such as recommendation and alignment.
The detailed statistical results of the datasets are shown in Table~\ref{tab:dataset of nodeL}.


\paragraph{Link-Level Tasks}
The edge is also an important part of the graph structure, which associates independent nodes and is the key to distinguishing graph data from non-relational data.
Especially in some specific fields (e.g., molecules, proteins), edges contain real information, so there are various tasks related to edges.

\myparagraph{Link Classification}
Similar to the NCF, the Link Classification (LC) also assigns one or more labels to a given edge.
In fact, in LC, the nodes at both ends of the edge are still taken into consideration.

\myparagraph{Link Prediction}
Link Prediction (LP) is a common graph task (e.g., knowledge graph). 
The goal of LP is to predict edges that are removed or may exist in the graph.
Similar to NCI, LP is also one of the pretext tasks in self-supervised learning, and its statistic results as shown in Table~\ref{tab:dataset of lc}.

\begin{table*}[htbp]
	\tiny
	\centering
	\caption{The statistics of the datasets for LC. Homogeneous:Hom, Heterogeneous:Het.}
	\label{tab:dataset of lc}
	\resizebox{\textwidth}{!}{
	\begin{tabular}{ccccccccl}  
		\hline 
		\textbf{Name} & \textbf{Usage} & \textbf{Source} & \textbf{Type} & \textbf{Nodes} & \textbf{Edges} & \textbf{Class} & \textbf{Features} & \textbf{Related Paper}  \\
		\hline
		\multirow{2}{*}{Cora} & \multirow{2}{*}{Downstream} & \multirow{2}{*}{Citation} & \multirow{2}{*}{Hom} & \multirow{2}{*}{2,708} & \multirow{2}{*}{5,429} & \multirow{2}{*}{7} & \multirow{2}{*}{1,433} &  \cite{M3S2020aaai,you2020icml,jin2020arxiv,hu2019arxiv,CAGNN2020arxiv,graphsage2017nips,vgae2016nips,supergat2021iclr,grace2020arxiv,s2grl2020arxiv}  \\
		&  &  &  &  &  &  &  & \cite{dgi2019iclr,cmvrl2020icml,subcom2020icdm,gic2021pakdd,cg3aaai2021,graphbert2020arxiv,gmi2020www,merit2021arxiv} \\\hline
		\multirow{2}{*}{CiteSeer} & \multirow{2}{*}{Downstream} & \multirow{2}{*}{Citation} & \multirow{2}{*}{Hom} & \multirow{2}{*}{3,327} & \multirow{2}{*}{4,732} & \multirow{2}{*}{6} & \multirow{2}{*}{3,703} &  \cite{M3S2020aaai,you2020icml,jin2020arxiv,CAGNN2020arxiv,s2grl2020arxiv,vgae2016nips,supergat2021iclr,dgi2019iclr,grace2020arxiv}  \\
		&  &  &  &  &  &  &  & \cite{cmvrl2020icml,subcom2020icdm,gic2021pakdd,cg3aaai2021,graphbert2020arxiv,gmi2020www,autossl2021arxiv,merit2021arxiv} \\\hline
		\multirow{2}{*}{PubMed} & \multirow{2}{*}{Downstream} & \multirow{2}{*}{Citation} & \multirow{2}{*}{Hom} & \multirow{2}{*}{19K} & \multirow{2}{*}{44K} & \multirow{2}{*}{3} & \multirow{2}{*}{500} &  \cite{M3S2020aaai,you2020icml,jin2020arxiv,hu2019arxiv,CAGNN2020arxiv,s2grl2020arxiv,vgae2016nips,supergat2021iclr,grace2020arxiv,dgi2019iclr}  \\
		&  &  &  &  &  &  &  & \cite{cmvrl2020icml,subcom2020icdm,gic2021pakdd,cg3aaai2021,graphbert2020arxiv,gmi2020www,merit2021arxiv} \\\hline
		ML-100K & Downstream & Movie & Hom & 2,625 & 100K & 5 & - & \cite{hu2019arxiv} \\\hline
		ML-1M & Downstream & Movie & Hom  & 9,940 & 1M & 5 & - & \cite{hu2019arxiv} \\\hline
		BlogCatalog-5K & Downstream & Social & Hom & 5,196 & 172K & 6 & 8,189 & \cite{s2grl2020arxiv,gmi2020www} \\\hline
		Amazon & Downstream & Social & Het  & 130M & - & - & - &  \cite{gptgnn2020kdd} \\\hline
		PPI-57K & Downstream & Protein & Het & 57K & 819K & 121 & 50 & \cite{s2grl2020arxiv,supergat2021iclr,grace2020arxiv,subcom2020icdm,gmi2020www} \\\hline
		Flickr-7K & Downstream & Photo & Hom & 7,575 & 240M & 9 & 12,047 & \cite{s2grl2020arxiv,gmi2020www} \\\hline
		Last-FM & Downstream & Music & Hom & 15K & 73K & 122 & - & \cite{selar2020nips} \\\hline
		Book-Crossing & Downstream & Book & Hom & 111K & 443K & 52 & - & \cite{selar2020nips} \\\hline
		OAG & Downstream & Academic & Het  & 178M & 2B & - & - & \cite{gptgnn2020kdd} \\
		\hline
	\end{tabular}
	}
\end{table*}

\myparagraph{Top-K Recommendation}
Top-K Recommendation (TKR) is exactly the same as the definition of TKS, the difference lies in the sorting goal.

\paragraph{Graph-Level Tasks}
The graph-level task generally focuses on the distribution of nodes, edges, and attributes in a given graph, in order to infer the possible properties of the entire graph.

\myparagraph{Graph Classification}
Graph Classification (GC) is commonly used in social, molecular, and protein graph data, which aims to predict the property of the given community, chemical compound, and protein.
The statistic results as shown in Table~\ref{tab:dataset of gc}.

\begin{table*}[htbp]
	\tiny
	\centering
	\caption{The statistics of the datasets for GC. Homogeneous:Hom, Heterogeneous:Het.}
	\label{tab:dataset of gc}
	\resizebox{\textwidth}{!}{
	\begin{tabular}{ccccccccl}  
		\hline 
		\textbf{Name} & \textbf{Usage} & \textbf{Source} & \textbf{Type} & \textbf{Graphs} & \textbf{ Nodes} & \textbf{Edges} & \textbf{Class} & \textbf{Related Paper}  \\
		\hline
		ZINC15 & Pretraining & Molecule & Hom & 2M & - & - & - & \cite{hu2020iclr,grover2020nips} \\\hline
		ChEMBL & Pretraining & Molecule & Hom & 456K & - & - & - & \cite{hu2020iclr,grover2020nips} \\\hline
		PPI-pre & Pretraining & Protein & Het & 395K & - & - & - & \cite{hu2020iclr} \\\hline
		MUTAG & Downstream & Molecule & Hom & 188 & - & - & 2 & \cite{hu2020iclr,igsd2020arxiv,cmvrl2020icml,infog2020iclr,sugar2021www,joao2021icml,mocl2021kdd,lin2021prototypical} \\\hline
		PTC & Downstream & Molecule & Hom & 344 & - & - & 2 & \cite{hu2020iclr,igsd2020arxiv,cmvrl2020icml,infog2020iclr,sugar2021www,lin2021prototypical} \\\hline
		BBBP & Downstream & Molecule & Hom & 2,039 & - & - & 2 & \cite{hu2020iclr,grover2020nips,micro2020arxiv,joao2021icml,graphlog2021arxiv,mocl2021kdd} \\\hline
		Tox21 & Downstream & Molecule & Hom & 7,831 & - & - & 24 & \cite{hu2020iclr,grover2020nips,micro2020arxiv,joao2021icml,graphlog2021arxiv,mocl2021kdd} \\\hline
		ToxCast & Downstream & Molecule & Hom & 8,575 & - & - & 1,234 & \cite{hu2020iclr,grover2020nips,micro2020arxiv,joao2021icml,graphlog2021arxiv,mocl2021kdd} \\\hline
		SIDER & Downstream & Molecule & Hom & 1,427 & - & - & 54 & \cite{hu2020iclr,grover2020nips,micro2020arxiv,joao2021icml,graphlog2021arxiv,mocl2021kdd} \\\hline
		ClinTox & Downstream & Molecule & Hom & 1,478 & - & - & 4 & \cite{hu2020iclr,grover2020nips,micro2020arxiv,joao2021icml,graphlog2021arxiv,mocl2021kdd} \\\hline
		MUV & Downstream & Molecule & Hom & 93K & - & - & 34 & \cite{hu2020iclr,joao2021icml,graphlog2021arxiv} \\\hline
		HIV & Downstream & Molecule & Hom & 41K & - & - & 2 & \cite{hu2020iclr,micro2020arxiv,joao2021icml,graphlog2021arxiv} \\\hline
		BACE & Downstream & Molecule & Hom & 1,513 & - & - & 2 & \cite{hu2020iclr,micro2020arxiv,joao2021icml,graphlog2021arxiv,mocl2021kdd} \\\hline
		PPI-88K & Downstream & Protein & Het & 88K & - & - & 80 & \cite{hu2020iclr} \\\hline
		IMDB-M & Downstream & Movie & Hom & 1,500 & 19K & 99K & 3 & \cite{hu2019arxiv,gcc2020kdd,igsd2020arxiv,cmvrl2020icml,infog2020iclr} \\\hline
		IMDB-B & Downstream & Movie & Hom & 1,000 & 19K & 97K & 2 & \cite{hu2019arxiv,gcc2020kdd,igsd2020arxiv,cmvrl2020icml,infog2020iclr,joao2021icml} \\\hline
		FreeSolv & Downstream & Molecule & Hom & 642 & - & - & - & \cite{grover2020nips} \\\hline
		ESOL & Downstream & Molecule & Hom & 1,128 & - & - & - & \cite{grover2020nips} \\\hline
		Lipophilicity & Downstream & Molecule & Hom & 4,200 & - & - & - & \cite{grover2020nips} \\\hline
		QM7 & Downstream & Molecule & Hom & 6,830 & - & - & - & \cite{grover2020nips} \\\hline
		QM8 & Downstream & Molecule & Hom & 22K & - & - & - & \cite{grover2020nips} \\\hline
		COLLAB & Downstream & Co-author & Hom & 5,000 & 373K & - & 3 & \cite{gcc2020kdd,igsd2020arxiv,joao2021icml,lin2021prototypical} \\\hline
		RDT-B & Downstream & Co-author & Hom & 2,000 & 859K & - & 2 & \cite{gcc2020kdd,infog2020iclr,joao2021icml,lin2021prototypical} \\\hline
		RDT-M & Downstream & Co-author & Hom & 5,000 & 3M & - & 5 & \cite{gcc2020kdd,infog2020iclr,joao2021icml,lin2021prototypical} \\\hline
		NCI1 & Downstream & Molecule & Hom & 4,110 & 123K & 132K & 2 & \cite{graphcl2020nips,cssl2021aaai,igsd2020arxiv,sugar2021www,joao2021icml,lin2021prototypical} \\\hline
		NCI109 & Downstream & Molecule & Hom & 4,127 & 123K & 133K & 2 & \cite{sugar2021www} \\\hline
		PROTEINS & Downstream & Molecule & Hom & 1,113 & 44K & 81K & 2 & \cite{graphcl2020nips,cssl2021aaai,sugar2021www,joao2021icml,lin2021prototypical} \\\hline
		D$\&$D & Downstream & Molecule & Hom & 1,178 & 335K & 843K & 2 & \cite{sugar2021www,joao2021icml} \\\hline
		Mutagenicity & Downstream & Molecule & Hom & 4,337 & 131K & 134K & 2 & \cite{cssl2021aaai} \\\hline
		METR-LA & Downstream & Traffic & Hom & 1 & 207 & - & - & \cite{stdgiarxiv2019} \\	
		\hline
	\end{tabular}
	}
\end{table*}

\paragraph{Data Source}
The PFMs of GL have been widely used in a mass of fields. 
We will descript the details of the pretraining datasets and the downstream task datasets.

\myparagraph{Citation and Co-author network}
A citation is a basic local representation, whose structure reflects the citation relationships of papers in a research direction or field.
Specifically, a citation network is a kind of relational data composed of research papers as nodes and citation relations as edges.
Among them, the citation network used in the GL model usually comes from local samples of common citation databases, e.g., Cora, Citeseer, and PubMed, and serves as downstream tasks.
Similarly, 
the co-author network is a dataset of scientific collaboration that corresponds to a researcher's ego network, in which the researcher and their collaborators are nodes and an edge indicates collaboration between two researchers.
According to different requirements of downstream tasks, such co-author networks can be used for various tasks, e.g., node classification and graph classification.

\myparagraph{Molecular and protein network}
A molecular network usually refers to a compound composed of atoms and atomic bonds, and predicting the properties of the compound is usually regarded as a graph classification task.
For example, MUTAG is a collection of nitroaromatic compounds whose goal is to predict their mutagenicity to Salmonella typhimurium.
PTC uses a graph to show the structure of multiple compounds and aims to predict the carcinogenicity of different compounds in rats.
The protein network is a collection of proteins classified as either enzymes or non-enzymes.
The amino acids are represented by nodes, and two nodes are connected by an edge if they are less than 6 Angstroms apart.

\myparagraph{Social and Movie network}
The social network is the social-relational data in the real network environment, which usually represents the relationship between users or posts.
For instance, 
Reddit is a graph dataset comprised of Reddit posts made in September 2014.
BlogCatalog is a graph dataset that represents a network of social relationships between bloggers who are listed on the BlogCatalog website.
The movie network is usually composed of actors and their co-occurrence participation in the movie.
For example, IMDB-B is a movie collaboration dataset that contains a large number of self-networks of actors who play movie roles in IMDB. 
Nodes in each graph represent actors/actresses, and if they appear in the same film, an edge connects them.
These graphs are based on action and romance genres.
The difference between IMDB-M and IMDB-B is that a node in the graph represents one or more actors.

\myparagraph{Others}
Some of the rarer graph data are used to test the universality of the PFM, such as word networks (Wikipedia), book networks (Book-crossing), and airline networks (US-Airport).
In addition, there are also some special graph structures adapted to specific models, such as spatiotemporal graphs (METR-LA).

\bibliographystyle{ieeetr}
\bibliography{PFM}

\end{document}